\documentclass[11pt]{article}

\usepackage[a4paper,margin=1.1in]{geometry}
\usepackage[T1]{fontenc}
\usepackage[expansion=false]{microtype}
\usepackage{amsmath,amssymb,amsthm}
\usepackage[numbers,sort&compress]{natbib}
\usepackage{graphicx}
\usepackage{booktabs}
\usepackage{bm}
\usepackage{mathtools}
\usepackage{xcolor}
\usepackage{etoolbox}
\usepackage{url}
\usepackage{tikz}
\usetikzlibrary{arrows.meta, positioning, calc}
\usepackage[colorlinks=true,linkcolor=black!60!blue,citecolor=black!60!blue,urlcolor=black!60!blue]{hyperref}
\setcitestyle{authoryear,round}

\theoremstyle{plain}
\newtheorem{theorem}{Theorem}
\newtheorem{lemma}[theorem]{Lemma}
\newtheorem{proposition}[theorem]{Proposition}
\newtheorem{corollary}[theorem]{Corollary}
\theoremstyle{definition}
\newtheorem{definition}[theorem]{Definition}
\newtheorem{assumption}[theorem]{Assumption}

\input{refshim.tex}        % \Cref without cleveref

\makeatletter
\newcounter{algbox}
\renewcommand{\thealgbox}{\arabic{algbox}}

\providecommand{\toclevel@algbox}{0}
\def\fps@algbox{tbp}
\def\ftype@algbox{4}
\def\ext@algbox{loa}
\def\fnum@algbox{Algorithm~\thealgbox}
\newenvironment{algbox}[1][tbp]{\par\addvspace{8pt}\noindent\begin{minipage}{\linewidth}\def\@captype{algbox}}{\end{minipage}\par\addvspace{8pt}}
\makeatother
\graphicspath{{figs/}}

\newenvironment{keywords}{\par\smallskip\noindent\textbf{Keywords:} }{\par}

\renewcommand{\vec}[1]{\bm{#1}}
\newcommand{\mat}[1]{\bm{\mathrm{#1}}}
\newcommand{\trans}{^{\mkern-1.5mu\mathsf{T}}}
\newcommand{\Acl}{\mat{A}_{\mathrm{cl}}}
\newcommand{\Lpsi}{L_{\Psi}}

\newcommand{\R}{\mathbb{R}}

\newcommand{\tr}{\mathrm{tr}}

\newcommand{\Jtheta}{\mat{J}_{\theta}}

\newcommand{\changliu}[1]{}
\newcommand{\tikreply}[1]{}

\IfFileExists{figs/paper_numbers.tex}{% Auto-generated by paper_export. Do not edit.

\newcommand{\pnAETotFp}{18}
\newcommand{\pnAETotN}{40}

\newcommand{\pnAIMujocoDoubleCartPoleZeroConv}{0}

\newcommand{\pnAIMujocoDoubleCartPoleZeroNAnch}{9}

\newcommand{\pnArmEpisodes}{30}

\newcommand{\pnArmFiveIbcMsMed}{2.1}

\newcommand{\pnArmFiveIbcSucc}{100}

\newcommand{\pnArmFiveLinSucc}{100}

\newcommand{\pnArmFiveStdSucc}{100}

\newcommand{\pnArmFiveTunedSucc}{1}

\newcommand{\pnArmGoals}{10}
\newcommand{\pnArmModels}{10}

\newcommand{\pnArmPFiveSucc}{3.1\times10^{-5}}

\newcommand{\pnArmPStdCost}{1.6\times10^{-4}}

\newcommand{\pnArmTwentyIbcCost}{543}

\newcommand{\pnArmTwentyIbcSucc}{100}

\newcommand{\pnArmTwentyLinSucc}{100}

\newcommand{\pnArmTwentyStdCost}{566}

\newcommand{\pnArmTwentyStdSucc}{100}

\newcommand{\pnArmTwentyTunedSucc}{91}

\newcommand{\pnCatchCells}{420}

\newcommand{\pnCatchMatch}{420}
\newcommand{\pnCatchModels}{40}
\newcommand{\pnCatchStdLin}{0}

\newcommand{\pnCenUncN}{40}

\newcommand{\pnClAeClUnstPct}{3}

\newcommand{\pnClAeLinUnstPct}{61}

\newcommand{\pnClAeNlUnstPct}{5}

\newcommand{\pnClCenUnstPct}{100}

\newcommand{\pnClDkucUnstPct}{28}

\newcommand{\pnClKmUnstPct}{61}

\newcommand{\pnClNoIdUnstPct}{70}

\newcommand{\pnClPolyThreeUnstPct}{100}

\newcommand{\pnClPolyTwoUnstPct}{100}

\newcommand{\pnClRbfUnstPct}{47}

\newcommand{\pnClRecUnstPct}{100}

\newcommand{\pnClSkelUnstPct}{100}

\newcommand{\pnClSocUnstPct}{100}

\newcommand{\pnClTikUnstPct}{0}

\newcommand{\pnDCLocTotConc}{43}

\newcommand{\pnDCLocTotN}{105}

\newcommand{\pnDagCpBase}{0.60}
\newcommand{\pnDagCpBetter}{10}
\newcommand{\pnDagCpCtrlCells}{0}
\newcommand{\pnDagCpN}{10}
\newcommand{\pnDagCpNew}{0.70}
\newcommand{\pnDagCpOracle}{0.70}

\newcommand{\pnDagCtrlN}{30}
\newcommand{\pnDagCtrlStable}{2}
\newcommand{\pnDagDcpBase}{0.06}
\newcommand{\pnDagDcpBetter}{9}
\newcommand{\pnDagDcpCtrlCells}{0}
\newcommand{\pnDagDcpN}{9}
\newcommand{\pnDagDcpNew}{0.15}
\newcommand{\pnDagDcpOracle}{0.15}

\newcommand{\pnDagDpuBase}{0.40}
\newcommand{\pnDagDpuBetter}{10}
\newcommand{\pnDagDpuN}{10}
\newcommand{\pnDagDpuNew}{1.50}

\newcommand{\pnDagMcpBase}{0.75}
\newcommand{\pnDagMcpBetter}{6}
\newcommand{\pnDagMcpCtrlCells}{6}
\newcommand{\pnDagMcpN}{10}
\newcommand{\pnDagMcpNew}{0.75}
\newcommand{\pnDagMcpOracle}{0.90}

\newcommand{\pnDagTotN}{39}
\newcommand{\pnDagTotStable}{39}

\newcommand{\pnDmCatch}{90}
\newcommand{\pnDmCtrl}{16}
\newcommand{\pnDmCtrlN}{10}
\newcommand{\pnDmCtrlZero}{8}
\newcommand{\pnDmEnt}{84}
\newcommand{\pnDmEps}{20}
\newcommand{\pnDmLin}{90}
\newcommand{\pnDmModels}{20}
\newcommand{\pnDmOld}{4}
\newcommand{\pnDmTikAll}{83}
\newcommand{\pnDmTikLat}{83}
\newcommand{\pnDmTikUnc}{82}

\newcommand{\pnEDMDTotFp}{15}
\newcommand{\pnEDMDTotN}{40}

\newcommand{\pnExDkucFp}{40}

\newcommand{\pnExDkucN}{40}

\newcommand{\pnExKmFp}{23}

\newcommand{\pnExKmN}{40}

\newcommand{\pnExSkelFp}{40}

\newcommand{\pnExSkelN}{40}

\newcommand{\pnExSocFp}{40}

\newcommand{\pnExSocN}{40}

\newcommand{\pnFpOLUpctFp}{66}
\newcommand{\pnFpOLUtrained}{115}
\newcommand{\pnFpOLUunstK}{73}

\newcommand{\pnFpReluMin}{1.1\times10^{-5}}

\newcommand{\pnFpUncOLUN}{39}

\newcommand{\pnFpUncOLUTrained}{40}

\newcommand{\pnGenAeClCertFp}{36}

\newcommand{\pnGenAeClN}{40}

\newcommand{\pnGenAeNlCertFp}{37}

\newcommand{\pnGenAeNlN}{40}

\newcommand{\pnGenLinKMax}{6.3\times10^{3}}

\newcommand{\pnGenPolyThreeCertFp}{10}

\newcommand{\pnGenPolyThreeN}{40}

\newcommand{\pnGenPolyTwoCertFp}{29}

\newcommand{\pnGenPolyTwoN}{40}

\newcommand{\pnGenRffCertFp}{0}

\newcommand{\pnGenRffN}{40}

\newcommand{\pnGoAnchors}{7}

\newcommand{\pnGoCertN}{150}
\newcommand{\pnGoDAMax}{21}
\newcommand{\pnGoDAMin}{5}

\newcommand{\pnGoEpisodes}{150}

\newcommand{\pnGoRhoTrueMed}{1.0000}
\newcommand{\pnGoRrMax}{4.4\times10^{-4}}

\newcommand{\pnGrFreeSucc}{77}
\newcommand{\pnGrLinCompMed}{0.68}
\newcommand{\pnGrLinFall}{50}
\newcommand{\pnGrLinSucc}{3}

\newcommand{\pnGrOldCompMed}{0.01}
\newcommand{\pnGrOldFall}{10}
\newcommand{\pnGrOldSucc}{26}
\newcommand{\pnGrOursCompMed}{0.83}
\newcommand{\pnGrOursFall}{21}
\newcommand{\pnGrOursFric}{73}
\newcommand{\pnGrOursMassHi}{74}
\newcommand{\pnGrOursMassLo}{57}
\newcommand{\pnGrOursNoise}{67}
\newcommand{\pnGrOursSucc}{67}

\newcommand{\pnGrPSl}{9.5\times10^{-3}}

\newcommand{\pnGrPSmppi}{2.3\times10^{-3}}
\newcommand{\pnGrSlCompMed}{5.35}
\newcommand{\pnGrSlFall}{4}

\newcommand{\pnGrSlSucc}{51}
\newcommand{\pnGrSmppiCompMed}{6.74}
\newcommand{\pnGrSmppiFall}{15}
\newcommand{\pnGrSmppiSucc}{47}

\newcommand{\pnGsRMax}{10^{2}}
\newcommand{\pnGsRMin}{10^{-2}}
\newcommand{\pnGsUncAll}{39}

\newcommand{\pnGsUncEstate}{182}
\newcommand{\pnGsUncN}{40}

\newcommand{\pnICDefZBilStab}{20}

\newcommand{\pnICDefZCov}{100}
\newcommand{\pnICDefZCovN}{100}

\newcommand{\pnICDefZLCov}{80}

\newcommand{\pnICDefZTBilStab}{20}

\newcommand{\pnICDefZTCov}{100}
\newcommand{\pnICDefZTCovN}{100}

\newcommand{\pnICDefZTLCov}{100}

\newcommand{\pnICDsgFiveTCov}{100}
\newcommand{\pnICDsgFiveTCovN}{100}

\newcommand{\pnICDsgFiveTLCov}{100}

\newcommand{\pnICDsgSevenTVBilStab}{20}

\newcommand{\pnICDsgSevenTVCov}{100}
\newcommand{\pnICDsgSevenTVCovN}{100}

\newcommand{\pnICDsgSevenTVLCov}{100}

\newcommand{\pnICDsgThEightTJBilStab}{20}

\newcommand{\pnICDsgThEightTJCov}{100}
\newcommand{\pnICDsgThEightTJCovN}{100}

\newcommand{\pnICDsgThEightTJLCov}{100}

\newcommand{\pnIDDefZCov}{40}
\newcommand{\pnIDDefZCovN}{40}

\newcommand{\pnIDDefZLCov}{40}

\newcommand{\pnIDDefZStab}{20}

\newcommand{\pnIDDsgNineVCov}{40}
\newcommand{\pnIDDsgNineVCovN}{40}

\newcommand{\pnIDDsgNineVLCov}{40}

\newcommand{\pnIDDsgSixCov}{40}
\newcommand{\pnIDDsgSixCovN}{40}

\newcommand{\pnIDDsgSixLCov}{40}

\newcommand{\pnIDDsgSixStab}{0}

\newcommand{\pnIDDsgTenJCov}{40}
\newcommand{\pnIDDsgTenJCovN}{40}

\newcommand{\pnIDDsgTenJLCov}{40}

\newcommand{\pnIDDsgTenJStab}{20}

\newcommand{\pnIbcRhoMax}{0.9949}

\newcommand{\pnIbcSrcRhoMax}{113}
\newcommand{\pnIbcSrcRhoMin}{1.002}

\newcommand{\pnInvCartPoleOneRoll}{0.012}

\newcommand{\pnInvCartPoleZeroRoll}{0.013}

\newcommand{\pnInvMujocoDoubleCartPoleOneN}{6}

\newcommand{\pnInvMujocoDoubleCartPoleOneUnst}{5}

\newcommand{\pnItFp}{46}

\newcommand{\pnItUnst}{45}

\newcommand{\pnKcCpDkucN}{10}
\newcommand{\pnKcCpDkucS}{10}

\newcommand{\pnKcCpKmN}{7}
\newcommand{\pnKcCpKmS}{0}

\newcommand{\pnKcCpNnN}{10}
\newcommand{\pnKcCpNnS}{10}

\newcommand{\pnKcCpPenN}{10}
\newcommand{\pnKcCpPenS}{10}

\newcommand{\pnKcCpRefN}{1}
\newcommand{\pnKcCpRefS}{1}
\newcommand{\pnKcCpSkelN}{10}
\newcommand{\pnKcCpSkelS}{0}

\newcommand{\pnKcCpSocN}{10}
\newcommand{\pnKcCpSocS}{0}

\newcommand{\pnKcCpStdN}{10}
\newcommand{\pnKcCpStdS}{0}

\newcommand{\pnKcCpTikN}{10}
\newcommand{\pnKcCpTikS}{10}

\newcommand{\pnKcDcpDkucN}{10}
\newcommand{\pnKcDcpDkucS}{0}

\newcommand{\pnKcDcpNnN}{10}
\newcommand{\pnKcDcpNnS}{1}

\newcommand{\pnKcDcpPenN}{10}
\newcommand{\pnKcDcpPenS}{0}

\newcommand{\pnKcDcpRefN}{1}
\newcommand{\pnKcDcpRefS}{1}
\newcommand{\pnKcDcpSkelN}{10}
\newcommand{\pnKcDcpSkelS}{0}

\newcommand{\pnKcDcpSocN}{10}
\newcommand{\pnKcDcpSocS}{0}

\newcommand{\pnKcDcpStdN}{10}
\newcommand{\pnKcDcpStdS}{0}

\newcommand{\pnKcDcpTikN}{10}
\newcommand{\pnKcDcpTikS}{10}

\newcommand{\pnKcDpuDkucN}{10}
\newcommand{\pnKcDpuDkucS}{9}

\newcommand{\pnKcDpuKmN}{10}
\newcommand{\pnKcDpuKmS}{9}

\newcommand{\pnKcDpuNnN}{10}
\newcommand{\pnKcDpuNnS}{0}

\newcommand{\pnKcDpuPenN}{10}
\newcommand{\pnKcDpuPenS}{7}

\newcommand{\pnKcDpuRefN}{1}
\newcommand{\pnKcDpuRefS}{1}
\newcommand{\pnKcDpuSkelN}{10}
\newcommand{\pnKcDpuSkelS}{0}

\newcommand{\pnKcDpuSocN}{10}
\newcommand{\pnKcDpuSocS}{0}

\newcommand{\pnKcDpuStdN}{10}
\newcommand{\pnKcDpuStdS}{0}

\newcommand{\pnKcDpuTikN}{10}
\newcommand{\pnKcDpuTikS}{10}

\newcommand{\pnKcMcpDkucN}{10}
\newcommand{\pnKcMcpDkucS}{10}

\newcommand{\pnKcMcpKmN}{6}
\newcommand{\pnKcMcpKmS}{0}

\newcommand{\pnKcMcpNnN}{10}
\newcommand{\pnKcMcpNnS}{10}

\newcommand{\pnKcMcpPenN}{10}
\newcommand{\pnKcMcpPenS}{10}

\newcommand{\pnKcMcpRefN}{1}
\newcommand{\pnKcMcpRefS}{1}
\newcommand{\pnKcMcpSkelN}{10}
\newcommand{\pnKcMcpSkelS}{0}

\newcommand{\pnKcMcpSocN}{10}
\newcommand{\pnKcMcpSocS}{0}

\newcommand{\pnKcMcpStdN}{10}
\newcommand{\pnKcMcpStdS}{0}

\newcommand{\pnKcMcpTikN}{10}
\newcommand{\pnKcMcpTikS}{10}

\newcommand{\pnKmKErr}{10^{-12}}

\newcommand{\pnKmSolveMed}{7.3}

\newcommand{\pnLatKoopMs}{10.3}
\newcommand{\pnLatMlpMs}{17.4}

\newcommand{\pnMpkASpread}{0.085}

\newcommand{\pnMpkCertStd}{0}
\newcommand{\pnMpkCertStdN}{30}
\newcommand{\pnMpkEp}{36}

\newcommand{\pnMpkGoals}{6}

\newcommand{\pnMpkResid}{10^{-12}}

\newcommand{\pnMpkSeeds}{5}

\newcommand{\pnMpkStableOurs}{30}
\newcommand{\pnMpkStableStd}{2}

\newcommand{\pnMrCert}{29}
\newcommand{\pnMrCertN}{30}
\newcommand{\pnMrLinFiveSucc}{100}
\newcommand{\pnMrLinOneSucc}{100}

\newcommand{\pnMrLinTenSucc}{100}
\newcommand{\pnMrLinTwentySucc}{100}
\newcommand{\pnMrLinTwoSucc}{100}

\newcommand{\pnMrMppiFiveSucc}{98}
\newcommand{\pnMrMppiOneSucc}{0}

\newcommand{\pnMrMppiTenSucc}{98}
\newcommand{\pnMrMppiTwentySucc}{98}
\newcommand{\pnMrMppiTwoSucc}{1}

\newcommand{\pnMrOursFiveSucc}{100}

\newcommand{\pnMrOursOneSucc}{100}

\newcommand{\pnMrOursTenRep}{100}
\newcommand{\pnMrOursTenSucc}{89}
\newcommand{\pnMrOursTwentySucc}{97}
\newcommand{\pnMrOursTwoSucc}{100}

\newcommand{\pnMrPSlOne}{0.029}

\newcommand{\pnMrSlFiveSucc}{100}
\newcommand{\pnMrSlOneSucc}{92}

\newcommand{\pnMrSlTenSucc}{97}
\newcommand{\pnMrSlTwentySucc}{100}
\newcommand{\pnMrSlTwoSucc}{100}

\newcommand{\pnMrStdFiveSucc}{0}
\newcommand{\pnMrStdOneSucc}{0}
\newcommand{\pnMrStdTenSucc}{0}
\newcommand{\pnMrStdTwentySucc}{0}
\newcommand{\pnMrStdTwoSucc}{0}

\newcommand{\pnNcNoFp}{40}
\newcommand{\pnNcNoN}{40}

\newcommand{\pnNcNoUnst}{28}
\newcommand{\pnNcP}{3.4\times10^{-3}}

\newcommand{\pnPrIbcAgree}{100}

\newcommand{\pnPrKmax}{4}
\newcommand{\pnPrKmin}{0.05}
\newcommand{\pnPrNk}{80}
\newcommand{\pnPrUncAgree}{14}

\newcommand{\pnRepDcpN}{10}

\newcommand{\pnRepDcpOffN}{6}

\newcommand{\pnRepDcpOffStab}{6}

\newcommand{\pnRepDcpStab}{10}

\newcommand{\pnRepUpN}{12}

\newcommand{\pnRepUpStab}{12}

\newcommand{\pnRhoRouteAgree}{0.0201}
\newcommand{\pnRhoRouteMed}{0.0002}
\newcommand{\pnRhoRouteN}{129}

\newcommand{\pnRouteOtherErrMax}{0.01}

\newcommand{\pnRtInvRecOwn}{92}

\newcommand{\pnRtLatOneMlp}{58}

\newcommand{\pnRtMlpRec}{100}

\newcommand{\pnSMpPsfData}{0.020}

\newcommand{\pnSMviolPsfData}{0}
\newcommand{\pnSMviolPsfKoop}{41}

\newcommand{\pnScgCpIbcSucc}{100}
\newcommand{\pnScgCpIbcViol}{4}
\newcommand{\pnScgCpLqrEstSucc}{100}
\newcommand{\pnScgCpLqrEstViol}{2}
\newcommand{\pnScgCpLqrSucc}{100}
\newcommand{\pnScgCpLqrViol}{2}
\newcommand{\pnScgCpMpcSucc}{100}
\newcommand{\pnScgCpMpcViol}{32}
\newcommand{\pnScgCpN}{10}

\newcommand{\pnScgCpPsfEstSucc}{0}
\newcommand{\pnScgCpPsfEstViol}{72}
\newcommand{\pnScgCpPsfLatSucc}{0}
\newcommand{\pnScgCpPsfLatViol}{86}

\newcommand{\pnScgCpStdSucc}{0}

\newcommand{\pnScgCpStdViol}{96}
\newcommand{\pnScgEpisodes}{25}
\newcommand{\pnScgNnMppiSucc}{0}
\newcommand{\pnScgNnMppiViol}{7}
\newcommand{\pnScgNnSucc}{91}
\newcommand{\pnScgNnViol}{28}

\newcommand{\pnSrFreeHalfRec}{3}
\newcommand{\pnSrFreeHalfSw}{7}

\newcommand{\pnSrFreeOneRec}{15}
\newcommand{\pnSrFreeOneSw}{15}

\newcommand{\pnSrFreeQuarterRec}{4}
\newcommand{\pnSrFreeQuarterSw}{6}

\newcommand{\pnSrFreeTwoRec}{18}
\newcommand{\pnSrFreeTwoSw}{17}

\newcommand{\pnSrLinHalfRec}{37}
\newcommand{\pnSrLinHalfSw}{25}

\newcommand{\pnSrLinOneRec}{47}
\newcommand{\pnSrLinOneSw}{33}

\newcommand{\pnSrLinQuarterRec}{13}
\newcommand{\pnSrLinQuarterSw}{8}

\newcommand{\pnSrLinTwoRec}{42}
\newcommand{\pnSrLinTwoSw}{50}

\newcommand{\pnSrMlpHalfRec}{97}
\newcommand{\pnSrMlpHalfSw}{91}

\newcommand{\pnSrMlpOneRec}{97}
\newcommand{\pnSrMlpOneSw}{91}

\newcommand{\pnSrMlpTwoRec}{97}
\newcommand{\pnSrMlpTwoSw}{91}

\newcommand{\pnSrPostN}{6{,}982}

\newcommand{\pnSrRecN}{60}

\newcommand{\pnSrSwN}{12}

\newcommand{\pnSrTikHalfRec}{99}
\newcommand{\pnSrTikHalfSw}{88}

\newcommand{\pnSrTikOneRec}{99}
\newcommand{\pnSrTikOneSw}{88}

\newcommand{\pnSrTikQuarterRec}{99}
\newcommand{\pnSrTikQuarterSw}{88}

\newcommand{\pnSrTikTwoRec}{100}

\newcommand{\pnSrTikTwoSw}{100}

\newcommand{\pnSwIBilData}{99}

\newcommand{\pnSwIBilFreeData}{99}

\newcommand{\pnSwIBilFreeLifted}{0}

\newcommand{\pnSwIBilFreePlan}{57}

\newcommand{\pnSwIBilFreePlanReach}{100}

\newcommand{\pnSwIBilFreeUpFp}{20}

\newcommand{\pnSwIBilFreeUpUnst}{20}
\newcommand{\pnSwIBilLifted}{88}

\newcommand{\pnSwIBilPlan}{95}

\newcommand{\pnSwIBilPlanReach}{100}

\newcommand{\pnSwIBilUpFp}{20}

\newcommand{\pnSwIBilUpUnst}{0}

\newcommand{\pnSwIFtData}{76}

\newcommand{\pnSwIFtLifted}{0}

\newcommand{\pnSwIFtPlan}{0}

\newcommand{\pnSwIFtPlanReach}{74}

\newcommand{\pnSwIFtUpFp}{20}

\newcommand{\pnSwIFtUpUnst}{20}

\newcommand{\pnSwILinData}{16}

\newcommand{\pnSwILinLifted}{11}

\newcommand{\pnSwILinPlan}{6}

\newcommand{\pnSwILinPlanReach}{41}

\newcommand{\pnSwILinUpFp}{20}

\newcommand{\pnSwILinUpUnst}{0}

\newcommand{\pnSwISrcData}{91}

\newcommand{\pnSwISrcLifted}{0}

\newcommand{\pnSwISrcPlan}{0}

\newcommand{\pnSwISrcPlanReach}{95}

\newcommand{\pnSwISrcUpFp}{20}

\newcommand{\pnSwISrcUpUnst}{20}

\newcommand{\pnSysCartPoleDt}{0.02}
\newcommand{\pnSysCartPoleTr}{2.1\,M}
\newcommand{\pnSysCartPoleU}{10}
\newcommand{\pnSysDampingPendulumDt}{0.02}
\newcommand{\pnSysDampingPendulumTr}{2.1\,M}
\newcommand{\pnSysDampingPendulumU}{8}
\newcommand{\pnSysDoublePendulumDt}{0.01}
\newcommand{\pnSysDoublePendulumTr}{2.1\,M}
\newcommand{\pnSysDoublePendulumU}{6}
\newcommand{\pnSysDoublePendulumUpDt}{0.01}
\newcommand{\pnSysDoublePendulumUpTr}{2.1\,M}
\newcommand{\pnSysDoublePendulumUpU}{20}
\newcommand{\pnSysFrankaCleanDt}{0.02}
\newcommand{\pnSysFrankaCleanTr}{2.1\,M}
\newcommand{\pnSysFrankaCleanU}{10}

\newcommand{\pnSysGoTwoTr}{1.1\,M}

\newcommand{\pnSysMujocoCartPoleDt}{0.02}
\newcommand{\pnSysMujocoCartPoleSwingDt}{0.02}
\newcommand{\pnSysMujocoCartPoleSwingTr}{5.6\,M}
\newcommand{\pnSysMujocoCartPoleSwingU}{10}
\newcommand{\pnSysMujocoCartPoleTr}{2.1\,M}
\newcommand{\pnSysMujocoCartPoleU}{10}
\newcommand{\pnSysMujocoDoubleCartPoleDt}{0.01}
\newcommand{\pnSysMujocoDoubleCartPoleTr}{2.1\,M}
\newcommand{\pnSysMujocoDoubleCartPoleU}{20}

\newcommand{\pnTSNarrowN}{50000}
\newcommand{\pnTSWideN}{200000}

\newcommand{\pnTikPairTested}{40}
\newcommand{\pnTikPairTrained}{40}

}{}

\title{\bfseries Making Koopman Control Guarantees Hold on Robots}
\author{%
  Abulikemu Abuduweili \qquad Changliu Liu\\[3pt]
  Robotics Institute, Carnegie Mellon University \\[2pt]
    }
\date{}

\begin{document}
\maketitle

\begin{abstract}
Koopman operator theory lets linear controllers handle nonlinear robots: an encoder lifts the state into a latent space with approximately linear dynamics, on which LQR or MPC is designed. A controller that stabilizes the latent model, however, may fail to stabilize the real nonlinear system. Stability is analyzed on latent rollouts, where the linear model propagates its own predictions, but on the robot the nonlinear dynamics produce each next state, which the controller re-encodes at every step. We prove that latent stability transfers to the robot when the latent model reproduces the true closed-loop dynamics to first order on the embedding's tangent space at the target. We also derive a margin for approximate matching and show that an open-loop stable latent model cannot match the plant's linearization at an unstable equilibrium. We propose Tangent-Invariant Koopman (TIK), which estimates a local plant linearization from offline transitions and builds first-order matching to it into the Koopman matrices. At the target, TIK's feedback gain equals the LQR gain of this linearization, and a bilinear input term models state-dependent actuation during large maneuvers. On four simulated open-loop unstable benchmarks, all 40 Standard Koopman controllers stabilize their latent models but are locally unstable on the real system, whereas all 40 TIK controllers stabilize it. An offline stability test with empirical curvature estimates identifies the stable TIK controllers on two cart-pole systems but is inconclusive under high noise. Given explicit model-error and curvature bounds, an invariant terminal set guarantees constraint satisfaction under local feedback after entry. TIK reaches 88\% success in constrained cart-pole swing-up with a 0.25 ms planning budget and controls simulated seven-joint arms within 1–5 ms per step.

\medskip\noindent\textbf{Code and project page:} \url{https://walleclipse.github.io/tik-koopman}
\end{abstract}

\begin{keywords}
Koopman operator, model predictive control, stability certificates
\end{keywords}

\section{Introduction}\label{sec:intro}\label{sec:intro_gap}

% A Koopman controller can stabilize its learned model yet destabilize the system it controls. The reason is a mismatch between the two closed loops. In the learned model, the controller acts on propagated latent predictions. On the robot, it acts on freshly encoded measurements of the nonlinear plant (Figure 1a,b). Stability of the first loop need not imply stability of the second.

% A Koopman controller is designed on a learned model but deployed on a nonlinear plant. During design, we propagate the model
% in lifted coordinates and check its closed-loop stability. On the robot, the controller acts on a new encoding of the measured state after every
% step of the nonlinear plant (\Cref{fig:pipeline}a,b). These two loops can give opposite stability verdicts. In the double cart-pole
% example in \Cref{fig:pipeline}c, Standard Koopman predicts recovery from a $0.6^\circ$ tilt while the simulated plant falls.

\begin{figure}[tbp]
\centering
% Figure 1 of the paper: the latent loop (what LQR/MPC certifies) and the deployed loop (what the robot runs).
% Drawn in a Nature/Science schematic style: two framed panels side by side, grayscale encoding distinguishes the loop components.
\definecolor{modelStroke}{HTML}{555555}
\definecolor{modelFill}{HTML}{EEEEEE}
\definecolor{plantStroke}{HTML}{353535}
\definecolor{plantFill}{HTML}{DEDEDE}
\definecolor{tikInk}{HTML}{1A1A1A}
\definecolor{tikgray}{HTML}{555555}
\definecolor{tikRule}{HTML}{D6D6D6}
\definecolor{tikPanelBg}{HTML}{FAFAFA}
\resizebox{\linewidth}{!}{%
\begin{tikzpicture}[
  font=\sffamily\fontsize{8}{9.5}\selectfont, text=tikInk,
  panel/.style={draw=tikRule, line width=0.55pt, rounded corners=4pt, fill=tikPanelBg},
  blk/.style={draw=#1, line width=0.65pt, rounded corners=3pt, minimum height=8.6mm,
              inner xsep=4.5pt, inner ysep=2.5pt, align=center},
  learned/.style={blk=modelStroke, fill=modelFill},
  ctrl/.style={blk=tikInk!75, fill=white},
  plant/.style={blk=plantStroke, fill=plantFill},
  arr/.style={-{Stealth[length=4.2pt,width=3.5pt]}, line width=0.75pt, draw=tikInk!82},
  sig/.style={font=\sffamily\fontsize{7}{8.5}\selectfont, text=tikgray},
  statModel/.style={draw=modelStroke!65, fill=modelFill, rounded corners=2.5pt, inner xsep=6pt, inner ysep=2.2pt,
                    font=\sffamily\fontsize{7}{8.5}\selectfont, text=tikInk, align=center},
  statPlant/.style={draw=plantStroke!70, fill=plantFill, rounded corners=2.5pt, inner xsep=6pt, inner ysep=2.2pt,
                      font=\sffamily\fontsize{7}{8.5}\selectfont, text=tikInk, align=center},
  letter/.style={font=\sffamily\bfseries\fontsize{9}{10}\selectfont, anchor=north west},
  ttl/.style={font=\sffamily\bfseries\fontsize{7.8}{9.2}\selectfont, anchor=north west}
]
% ---------------- panel a: the latent loop ----------------
\draw[panel] (-0.30, -1.72) rectangle (6.80, 1.05);
\node[letter] at (-0.16, 0.95) {a};
\node[ttl] at (0.16, 0.94) {Latent loop};
\node[ctrl, minimum width=20mm] (Ka) at (0.92, 0.02) {Local feedback\\[-1pt] $\vec{u}_t=-\mat{K}\vec{z}_t$};
\node[learned, minimum width=30mm] (Ma) at (4.85, 0.02) {Learned linear model\\[-1pt] $\hat{\mathbf{z}}_{t+1}=\mat{A}\vec{z}_t+\mat{B}\vec{u}_t$};
\draw[arr] (Ka.east) -- node[sig, midway, above=1pt] {$\vec{u}_t$} (Ma.west);
\coordinate (MaBot) at (4.85, -0.82);
\coordinate (KaBot) at (0.92, -0.82);
\draw[arr] (Ma.south) -- (MaBot) -- node[sig, midway, above=1pt] {predicted state $\hat{\vec{z}}_{t+1}$} (KaBot) -- (Ka.south);
\node[statModel] at (3.25, -1.32) {Nominal certificate: $\Acl=\mat{A}-\mat{B}\mat{K}\;\;(\rho(\Acl)<1)$};

% ---------------- panel b: the deployed loop ----------------
\begin{scope}[xshift=7.10cm]
\draw[panel] (-0.30, -1.72) rectangle (7.95, 1.05);
\node[letter] at (-0.16, 0.95) {b};
\node[ttl] at (0.16, 0.94) {Deployed loop};
\node[learned, minimum width=21mm] (Eb) at (0.95, 0.02) {Re-encoding\\[-1pt] $\vec{z}_t=\Phi_\theta(\vec{x}_t)$};
\node[ctrl, minimum width=20mm] (Kb) at (3.65, 0.02) {Local feedback\\[-1pt] $\vec{u}_t=-\mat{K}\vec{z}_t$};
\node[plant, minimum width=24mm] (Pb) at (6.35, 0.02) {True system\\[-1pt] $\vec{x}_{t+1}=f(\vec{x}_t,\vec{u}_t)$};
\draw[arr] (Eb.east) -- node[sig, midway, above=1pt] {$\vec{z}_t$} (Kb.west);
\draw[arr] (Kb.east) -- node[sig, midway, above=1pt] {$\vec{u}_t$} (Pb.west);
\coordinate (PbBot) at (6.35, -0.82);
\coordinate (EbBot) at (0.95, -0.82);
\draw[arr] (Pb.south) -- (PbBot) -- node[sig, midway, above=1pt] {measured state $\vec{x}_{t+1}$} (EbBot) -- (Eb.south);
\node[statPlant] at (3.82, -1.32) {Deployed Jacobian: $\mat{F}=\mat{A}_f-\mat{B}_f\mat{K}\Jtheta$};
\end{scope}
\end{tikzpicture}%
}%
\par\vspace{3pt}
\noindent\begin{minipage}[t]{0.49\linewidth}
  \vspace{0pt}
  \noindent\textsf{\textbf{c}\enspace \footnotesize{\textbf{Double cart-pole}}}\par\vspace{2pt}
  \includegraphics[width=\linewidth]{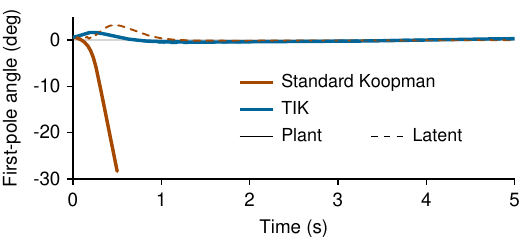}
\end{minipage}\hfill
\begin{minipage}[t]{0.49\linewidth}
  \vspace{0pt}
  \noindent\textsf{\textbf{d}\enspace \footnotesize{\textbf{Tangent action at the target}}}\par\vspace{5pt}
  \centering% Schematic first-order state action at the target.
\resizebox{0.85\linewidth}{!}{%
\begin{tikzpicture}[
  font=\sffamily\footnotesize,
  arrow/.style={-{Stealth[length=3.5pt,width=3pt]},line width=.8pt},
  label text/.style={font=\sffamily\footnotesize,inner sep=1pt}
]
\draw[black!45,line width=1pt]
  (0.1,0.45) .. controls (1.45,0.60) and (2.12,1.02) .. (3.15,1.02)
  .. controls (4.2,1.02) and (4.87,0.60) .. (6.2,0.45);
\node[label text,anchor=north west,text=black!65] at (0.12,0.43) {Encoded manifold $\mathcal{M}$};
\draw[black!65,densely dashed] (0.1,1.02) -- (6.20,1.02);
\node[label text,anchor=south east] at (6.2,1.76) {Tangent space $\mathrm{range}(\Jtheta)$};
\draw[black!45,line width=.4pt] (5.72,1.74) -- (5.72,1.03);
\draw[arrow] (3.15,1.02) -- (5.28,1.02);
\node[label text,anchor=south] at (4.52,1.10) {$\Jtheta\vec{v}$};
\draw[arrow,draw=black!75,line width=1.1pt] (3.15,1.02) -- (0.45,1.02);
\node[label text,anchor=south west] at (0.11,1.12) {TIK: $\Jtheta\hat{\mat{A}}_f\vec{v}$};
\draw[arrow,draw=black!65] (3.15,1.02) -- (1.55,2.11);
\node[label text,anchor=south west] at (0.11,2.17) {Off-tangent prediction $\mat{A}\Jtheta\vec{v}$};
\fill (3.15,1.02) circle (1.5pt);
\node[label text,anchor=north] at (3.15,0.90) {Target};
\end{tikzpicture}%
}
\end{minipage}
\caption{\textbf{The nominal and deployed loops can disagree.}
(a) The learned model propagates its own predictions. (b) Deployment feeds back the re-encoded plant state.
(c) First-pole angle for Standard Koopman and TIK on shared axes (seed $42$, initial tilt $0.6^\circ$).
Each controller has a latent prediction (dashed) and a simulated plant rollout (solid). The Standard Koopman plant trace
ends when the pole falls.
(d) The true linearized image of a tangent vector lies in $\mathrm{range}(\Jtheta)$. A learned update can leave this space.
TIK sets $\mat{A}\Jtheta\vec{v}=\Jtheta\hat{\mat{A}}_f\vec{v}$, which equals the true tangent action when the fit is exact.}
\label{fig:pipeline}
\end{figure}

Koopman operator theory lets linear controllers handle nonlinear robots~\citep{williams2015edmd,lusch2018deep,shi2022deep}. An encoder lifts the physical state into a latent space in which the dynamics are approximately linear, and a linear controller is designed on this latent model. However, a controller that stabilizes the latent model may fail to stabilize the real nonlinear system. The latent model propagates its own predictions (\Cref{fig:pipeline}a), whereas on the robot the nonlinear dynamics produce each next state and the controller re-encodes it at every step (\Cref{fig:pipeline}b). Stability of the first loop therefore does not guarantee stability of the second. In the double cart-pole example in \Cref{fig:pipeline}c, Standard Koopman predicts recovery from a $0.6^\circ$ tilt, while the simulated plant falls.

\changliu{in (a) and (b), try not using blue and orange as it confuses with the colors in (c).}%
\tikreply{Panels (a) and (b) now use grayscale. Orange and blue identify the controllers in panel (c).}%
\changliu{we only need to keep on example in (c), and we should illustrate the tangent space using figure in appendix B here, making it (d).}%
\tikreply{Panel (c) shows one double cart-pole example, with both controllers on the same axes. Panel (d) shows the tangent-space geometry from the original supplement.}%

With linear latent dynamics, quadratic costs, and linear
constraints, model predictive control (MPC) is a convex quadratic program (QP)~\citep{korda2018linear}. Near a target inside the constraints, MPC with a Riccati terminal cost applies the same feedback as linear-quadratic regulation (LQR) on the latent model.
The usual \emph{nominal stability test} checks that the latent closed-loop matrix $\Acl$ is Schur, that is, all its eigenvalues lie inside the unit circle.
The real system, however, is locally governed by the physical closed-loop Jacobian $\mat{F}$, which also depends on the derivatives of the encoder. A model with low prediction error can
therefore still give the wrong stability verdict.

% We ask when the nominal test certifies the plant. At the target, the embedding Jacobian $\Jtheta$ maps physical perturbations
% to the tangent space of the encoded states (\Cref{fig:pipeline}d). If $\Jtheta\mat{F}=\Acl\Jtheta$, every eigenvalue of $\mat{F}$
% is an eigenvalue of $\Acl$ (\Cref{thm:transfer}).

When does latent stability still hold on the robot after re-encoding? The answer lies in the tangent space of the learned embedding at the target. If the latent closed-loop dynamics reproduce the physical closed-loop Jacobian on this space, every eigenvalue of the physical closed loop also appears in the latent model, so latent stability implies local stability of the robot (\Cref{thm:transfer}). We also give a margin for approximate matching, and we show that a latent model
that is open-loop stable cannot match the plant's open-loop linearization at an open-loop unstable equilibrium (\Cref{cor:stablelift}).

We then build this matching into the model. \emph{Tangent-Invariant Koopman} (TIK) enforces two first-order equations
$\mat{A}\Jtheta=\Jtheta\hat{\mat{A}}_f$ and $\mat{B}=\Jtheta\hat{\mat{B}}_f$, where $(\hat{\mat{A}}_f,\hat{\mat{B}}_f)$ is a local plant linearization fitted from
offline transitions (\Cref{prop:ibc}). For a fixed encoder, these equations are linear in $\mat{A}$ and $\mat{B}$. Near the target, the fit then determines
\changliu{what is the representation of tangent invariance? why is that linear?}%
\tikreply{For fixed encoder weights, the lifting Jacobian is fixed. The two displayed matching equations are linear in the entries of the model matrices.}%
the feedback: TIK's gain equals the LQR gain of the fitted linearization. Away from the target, the learned latent model predicts the dynamics,
including state-dependent actuation during large maneuvers. We also give a stability test that accounts for estimation error (\Cref{thm:knownfp}) and an invariant
terminal set under explicit curvature and noise assumptions (\Cref{thm:region_main}).

On four simulated open-loop unstable benchmarks, all $40$ Standard Koopman controllers pass the nominal test but are locally unstable on the real system, whereas all $40$ TIK controllers stabilize it. Further experiments evaluate offline stability verification, constrained maneuvers, and control under short planning budgets.

\section{Related Work}\label{sec:related}

\paragraph{Learning Koopman models for control.}
\citet{williams2015edmd} fit a linear operator on a fixed dictionary. Neural methods learn the observables, often with a
latent-linearity loss that penalizes errors in the lifted coordinates~\citep{lusch2018deep,shi2022deep}.
Linear predictors with control inputs lead to QP-based MPC~\citep{korda2018linear,proctor2018generalizing}, and bilinear predictors
allow the input effect to depend on the state~\citep{bruder2021bilinear}. Applications include manipulation and legged
locomotion~\citep{chen2024korol,li2025continual,jung2026wholebody}. Exact finite-dimensional embeddings have structural
limits~\citep{brunton2016invariant,liu2025immersions}. Van Goor et al.\ study reprojection onto the encoded
manifold~\citep{vangoor2023reprojection}, and \citet{ge2026koopman} study how the action operating point affects learned
world-model accuracy for humanoid tracking. We study how re-encoding the measurement affects closed-loop stability at a target.

\paragraph{Local nonlinear stability and Koopman guarantees.}
Lyapunov's indirect method linearizes the closed loop in the physical state space~\citep{khalil2002nonlinear,elaydi2005difference}.
Koopman eigenfunction theory relates physical Jacobians to stability under exact embeddings~\citep{mauroy2016global}.
For learned predictors, \citet{shang2025expstab} establish local exponential stability under a
sufficiently small proportional one-step prediction error and identify
vanishing first-order prediction error as a sufficient condition. SafEDMD uses proportional surrogate-error bounds for robust feedback
synthesis~\citep{nuske2023finite,strasser2023feedback,strasser2024safedmd}; \citet{strasser2025overview} survey how model-error bounds support closed-loop guarantees. \citet{jain2026verification} estimate the Jacobian of a candidate closed loop from
near-equilibrium perturbations and use it to construct a region of
attraction. We derive an explicit singular-value margin for the
closed-loop error Jacobian. TIK matches derivatives to a fitted
state-input model and tests the remaining fit error in the direction
induced by the deployed feedback.
Open-loop stable parameterizations~\citep{mamakoukas2023learning,fan2024stablekoopman} encode useful prior knowledge
for stable plants. Our obstruction concerns using them at unstable equilibria.
\changliu{Here we should review existing nonlinear stability analysis methods that utilize linearization, and say the main difference in this work is that prior work are focusing on linearization on the original state space, but this work focuses on linearization on the latent space.}%
\tikreply{The related work now starts from the physical Jacobian in Lyapunov analysis and explains how the tangent-space relation connects it to the latent model.}%

\section{Background and Problem Setup}\label{sec:setup}\label{sec:formulation}\label{sec:results}

We consider a discrete-time nonlinear system, the \emph{plant} or true system, $\vec{x}_{t+1}=f(\vec{x}_t,\vec{u}_t)$ with physical state $\vec{x}_t\in\R^{n_x}$,
input $\vec{u}_t\in\R^{n_u}$, and unknown continuously differentiable dynamics $f$. We have logged transitions
$(\vec{x}_t,\vec{u}_t,\vec{x}_{t+1})$ and a specified equilibrium $(\vec{x}^\ast,\vec{u}^\ast)$ with $f(\vec{x}^\ast,\vec{u}^\ast)=\vec{x}^\ast$.
The goal is to stabilize this equilibrium under state and input limits. We
shift coordinates so that $(\vec{x}^\ast,\vec{u}^\ast)=(\vec{0},\vec{0})$ (\Cref{app:mainproofs} keeps
$(\vec{x}^\ast,\vec{u}^\ast)$ explicit) and write $\mat{A}_f=\partial f/\partial\vec{x}|_{(\vec{0},\vec{0})}$ and
$\mat{B}_f=\partial f/\partial\vec{u}|_{(\vec{0},\vec{0})}$ for the true Jacobians at the target. Let $\rho(\cdot)$ denote the
spectral radius (largest eigenvalue magnitude). The equilibrium is \emph{open-loop unstable} if $\rho(\mat{A}_f)>1$, as in an
upright pendulum or cart-pole; stabilizing such systems requires active feedback and is our focus.

\textbf{Neural Koopman model.} The Koopman operator advances observables by composition with the dynamics. For a fixed input $\vec{u}$, this action is
$\psi\mapsto\psi(f(\cdot,\vec{u}))$~\citep{koopman1931hamiltonian,proctor2018generalizing,korda2018linear}. It is linear even for nonlinear $f$, but acts
on an infinite-dimensional function space. In practice, it is approximated on $N$ observables stacked into an embedding
$\Phi_\theta:\R^{n_x}\to\R^{N}$ that concatenates $\vec{x}$ with a multilayer perceptron (MLP) encoder $\Psi_\theta:\R^{n_x}\to\R^{N-n_x}$ (writing
$[\vec{a};\,\vec{b}]$ for column stacking)~\citep{shi2022deep,chen2024korol,li2025continual,jung2026wholebody}:
\begin{align}\label{eq:latent_linear_model}
  \vec{z}_t=\Phi_\theta(\vec{x}_t)=\bigl[\vec{x}_t;\,\Psi_\theta(\vec{x}_t)-\Psi_\theta(\vec{0})\bigr],\qquad
  \hat{\vec{z}}_{t+1}=\mat{A}\vec{z}_t+\mat{B}\vec{u}_t,\qquad
  \vec{x}_t=\mat{C}\vec{z}_t,\ \ \mat{C}=[\mat{I}_{n_x}\ \ \mat{0}].
\end{align}
Thus $\Phi_\theta(\vec{0})=\vec{0}$ and $\mat{C}$ reads out the state linearly, keeping state costs and bounds convex in
$\vec{z}$. Because finite observable sets rarely span an exact invariant subspace~\citep{brunton2016invariant},
$\hat{\vec{z}}_{t+1}$ generally differs from $\Phi_\theta(\vec{x}_{t+1})$. \emph{Standard Koopman} trains $(\Phi_\theta,\mat{A},\mat{B})$ on
offline trajectories with a multi-step state-prediction loss and an inverse control loss~\citep{abuduweili2025scaling};
\emph{Koopman + invariance penalty} adds a latent-linearity loss following~\citet{lusch2018deep} (\Cref{app:training}).
This loss penalizes feature prediction errors on training trajectories. Tangent invariance (\Cref{def:ti}) constrains derivatives at the target.
\changliu{is this the invariance mentioned in the intro?}%
\tikreply{The introduction refers to first-order tangent invariance. The cited latent-linearity loss penalizes feature prediction errors on training trajectories.}%

\textbf{Koopman MPC and LQR.} Given $\vec{z}_0=\Phi_\theta(\vec{x}_t)$, Koopman MPC solves the convex QP
\begin{align}\label{eq:kmpc_qp}
  \min_{\vec{u}_{0:N_h-1}}\ &\textstyle\sum_{k=0}^{N_h-1}\bigl(\vec{z}_k\trans\mat{C}\trans\mat{Q}_x\mat{C}\vec{z}_k+\vec{u}_k\trans\mat{R}\vec{u}_k\bigr)
  +\vec{z}_{N_h}\trans\mat{P}\vec{z}_{N_h}
  \nonumber\\ \text{s.t.}\ \ &\vec{z}_{k+1}=\mat{A}\vec{z}_k+\mat{B}\vec{u}_k,\ \ \vec{x}_{\min}\le\mat{C}\vec{z}_k\le\vec{x}_{\max},\ \ |\vec{u}_k|\le\vec{u}_{\max},
\end{align}
and applies $\vec{u}_t=\vec{u}_0$~\citep{korda2018linear}, where $\mat{Q}_x\succeq0$, $\mat{R}\succ0$, and $\mat{P}\succeq0$ solves the Riccati equation of
$(\mat{A},\mat{B},\mat{C}\trans\mat{Q}_x\mat{C},\mat{R})$. When constraints are inactive, \eqref{eq:kmpc_qp} gives the Koopman LQR policy,
$\vec{u}_t=-\mat{K}\vec{z}_t$ with $\mat{K}=(\mat{R}+\mat{B}\trans\mat{P}\mat{B})^{-1}\mat{B}\trans\mat{P}\mat{A}$
(\Cref{prop:kmpc}). Both policies apply the same input in a neighborhood of the target and thus induce the same local
closed-loop Jacobian.
\changliu{I do not see the causality here with "so"}%
\tikreply{With the Riccati terminal cost and inactive constraints, both policies apply the same action locally, so they have the same deployed Jacobian.}%

\subsection{The Latent Loop versus the Deployed Loop}\label{sec:lifted_cert}\label{sec:deployed}

\textbf{Latent loop.} The model feeds back its own predictions, giving $\vec{z}_{t+1}=\Acl\vec{z}_t$ with
$\Acl=\mat{A}-\mat{B}\mat{K}$. The nominal stability certificate is $\rho(\Acl)<1$ ($\Acl$ is Schur).
With $\mat{R}\succ0$, stabilizability of $(\mat{A},\mat{B})$ and detectability of
$(\mat{A},\mat{Q}_x^{1/2}\mat{C})$ ensure a stabilizing LQR solution.

\textbf{Deployed loop.} The robot measures $\vec{x}_t$, encodes $\vec{z}_t=\Phi_\theta(\vec{x}_t)$, and applies
$\vec{u}_t=-\mat{K}\Phi_\theta(\vec{x}_t)$ to the true plant (\Cref{fig:pipeline}b). With target embedding Jacobian
$\Jtheta:=\partial\Phi_\theta/\partial\vec{x}|_{\vec{0}}=[\mat{I}_{n_x};\,\partial\Psi_\theta/\partial\vec{x}|_{\vec{0}}]\in\R^{N\times n_x}$
($\mat{C}\Jtheta=\mat{I}_{n_x}$, full column rank), the deployed loop and its target Jacobian are
\begin{align}\label{eq:deployed_loop}
  \vec{x}_{t+1}=g(\vec{x}_t):=f\bigl(\vec{x}_t,-\mat{K}\Phi_\theta(\vec{x}_t)\bigr),\qquad
  \mat{F}:=\frac{\partial g}{\partial\vec{x}}\Big|_{\vec{0}}=\mat{A}_f-\mat{B}_f\mat{K}\Jtheta.
\end{align}
By Lyapunov's indirect method~\citep{khalil2002nonlinear}, the target is locally exponentially stable on the true system if
$\rho(\mat{F})<1$ and unstable if $\rho(\mat{F})>1$. The nominal certificate checks the $N\times N$ matrix $\Acl$, whereas the
robot is governed by the $n_x\times n_x$ matrix $\mat{F}$. For a ReLU encoder, the Jacobian is constant in a neighborhood of a target where the activation pattern is fixed (\Cref{app:protocol}).
\changliu{what is "exact off
activation boundaries"?}%
\tikreply{The encoder is affine in a neighborhood where the ReLU activation pattern is fixed, so its Jacobian is constant there. The protocol describes the boundary check.}%

\section{When Does the Nominal Certificate Hold on the Robot?}\label{sec:rhogap}

We now compare $\Acl$ with $\mat{F}$ and identify when the nominal certificate transfers to the robot.

\textbf{One-step latent error.} Given $(\vec{x}_t,\vec{u}_t)$, the true next encoded state is $\Phi_\theta(f(\vec{x}_t,\vec{u}_t))$
and the linear model predicts $\mat{A}\Phi_\theta(\vec{x}_t)+\mat{B}\vec{u}_t$, giving the one-step latent prediction error
$E(\vec{x},\vec{u}):=\Phi_\theta(f(\vec{x},\vec{u}))-\mat{A}\Phi_\theta(\vec{x})-\mat{B}\vec{u}$.
The physical prediction loss directly supervises the first $n_x$ coordinates $\mat{C}E$. Multi-step training can influence
the other coordinates through later state predictions, but it imposes no exact derivative constraints at the target.
LQR uses the full latent model, including the learned feature coordinates.

\begin{definition}[Tangent invariance]\label{def:ti}
The model is \emph{tangent-invariant} at the target $(\vec{0},\vec{0})$ if the first derivatives of $E(\vec{x},\vec{u})$ vanish there:
\begin{align}\label{eq:ti}
  \frac{\partial E}{\partial\vec{x}}\Big|_{(\vec{0},\vec{0})}=\Jtheta\mat{A}_f-\mat{A}\Jtheta=\mat{0},\qquad
  \frac{\partial E}{\partial\vec{u}}\Big|_{(\vec{0},\vec{0})}=\Jtheta\mat{B}_f-\mat{B}=\mat{0}.
\end{align}
\end{definition}

Near a smooth target, the encoded states form a manifold $\mathcal{M}$ with tangent space $\mathrm{range}(\Jtheta)$
(\Cref{fig:pipeline}d). Tangent invariance here means exact first-order matching, not just preserving this subspace:
$\mat{A}\Jtheta\vec{v}=\Jtheta\mat{A}_f\vec{v}$ and $\mat{B}\vec{w}=\Jtheta\mat{B}_f\vec{w}$ for state and input
perturbations $\vec{v},\vec{w}$. For fixed $\Jtheta$, these are linear constraints on $(\mat{A},\mat{B})$.
Along the closed loop $\vec{u}=-\mat{K}\Phi_\theta(\vec{x})$, the one-step latent error is $\Phi_\theta(g(\vec{x}))-\Acl\Phi_\theta(\vec{x})$,
with target Jacobian
\begin{align}\label{eq:ecal}
  \mathcal{E} \;:=\; \Jtheta\mat{F}-\Acl\Jtheta
  \;=\; (\Jtheta\mat{A}_f-\mat{A}\Jtheta)-(\Jtheta\mat{B}_f-\mat{B})\mat{K}\Jtheta.
\end{align}
Exact open-loop matching makes $\mathcal{E}=\mat{0}$ for every gain. For a fixed gain, the two errors in \eqref{eq:ecal} can cancel even when each is nonzero.
For a Schur matrix $\mat{M}$, let
$d_1(\mat{M})=\min_{|s|=1}\sigma_{\min}(s\mat{I}-\mat{M})$ be its distance to the nearest complex matrix with a unit-circle eigenvalue.

\begin{theorem}[Transfer of the nominal stability certificate]\label{thm:transfer}
For the state-retaining lifting in \eqref{eq:latent_linear_model}, assume differentiability at the target and inactive input
limits nearby. Since $\mat{C}\Jtheta=\mat{I}$ and $\|\mat{C}\|_2=1$, each eigenvalue $\lambda$ of $\mat{F}$ satisfies
$\sigma_{\min}(\lambda\mat{I}-\Acl)\le\|\mathcal{E}\|_2$.
Consequently: (a)~if $\mathcal{E}=\mat{0}$ for the deployed gain, every eigenvalue of $\mat{F}$ is an eigenvalue of $\Acl$, so $\rho(\mat{F})\le\rho(\Acl)$ and
$\rho(\Acl)<1$ guarantees local exponential stability on the true system; (b)~if $\Acl$ is Schur and
$\|\mathcal{E}\|_2<d_1(\Acl)$, the target equilibrium is locally exponentially stable.
\end{theorem}
Exact matching embeds every physical closed-loop eigenvalue in the latent model. Approximate matching preserves stability
when the derivative error is smaller than the complex stability radius $d_1(\Acl)$. This margin accounts for sensitivity
to perturbations, including nonnormal dynamics. It is sufficient and may be conservative; failing the test does not imply
instability. Proofs are in \Cref{app:mainproofs}.

\begin{corollary}[Obstruction at unstable equilibria]\label{cor:stablelift}
At an equilibrium with $\rho(\mat{A}_f)>1$, tangent invariance implies $\rho(\mat{A})>1$.
\end{corollary}
Exact tangent matching embeds every eigenvalue of $\mat{A}_f$ in $\mat{A}$. A Schur $\mat{A}$ therefore cannot satisfy
these conditions at an unstable target. This does not rule out its use as an approximate predictor.
\Cref{app:example} illustrates how exact one-step physical prediction can coexist with a misleading latent certificate.
\changliu{To save space, we can skip the example in the main text. Just say \Cref{app:example} gives a numerical example.}%
\tikreply{The worked scalar example is in \Cref{app:example}. The main text points to it.}%

\section{Tangent-Invariant Koopman (TIK)}\label{sec:method}

TIK uses near-target transitions to estimate the local plant dynamics and trains a lifted model for prediction beyond that
neighborhood. \Cref{alg:tik} summarizes the method.
\changliu{need an overview of the algorithm first, then talk about the guarantees. We should move algorithm 1 to the main text.}%
\tikreply{Moved \Cref{alg:tik} to the beginning of the method, before the guarantees.}%

\begin{algbox}[tbp]
\hrule
\caption{\textbf{Tangent-Invariant Koopman (TIK)}}
\label{alg:tik}
\small
\textbf{Inputs.} Offline trajectories, target equilibrium, costs $\mat{Q}_x,\mat{R}$, and a Standard Koopman checkpoint.
\begin{enumerate}\setlength{\itemsep}{0pt}\setlength{\topsep}{2pt}
\item Fit $(\hat{\mat{A}}_f,\hat{\mat{B}}_f)$ using \eqref{eq:localfit}; require a full-rank weighted design.
\item Initialize the encoder and $\mat{W}$ from the checkpoint and any bilinear terms at zero. Fine-tune with
\eqref{eq:ibc} rebuilt at every update. Require the LQR conditions in \Cref{prop:ibc}(b); stop if they fail.
\item Solve the latent Riccati equation, set $\hat{\mat{K}}=\mat{K}\Jtheta$, and form
$\hat{\mat{F}}=\hat{\mat{A}}_f-\hat{\mat{B}}_f\hat{\mat{K}}$.
\item With curvature and noise bounds, test the stability margin (\Cref{thm:knownfp}). For dual-mode control, compute an
invariant $\Omega_c$ if \Cref{thm:region_main} holds. Otherwise report the corresponding guarantee as inconclusive.
\item For local regulation, solve the linear-model QP. For large maneuvers, plan with the bilinear predictor. If a terminal
set was obtained, switch to $\vec{u}=-\hat{\mat{K}}\vec{x}$ on entry to $\Omega_c$.
\end{enumerate}
\hrule
\end{algbox}

\subsection{Fitting and Parameterizing the Lifted Dynamics}

\textbf{Local plant fit.}\label{sec:datacert} The first-order Taylor model of $f$ describes the dynamics near the target.
The plant derivatives are unknown, so we estimate them from $M$ offline transitions near the target. For
$\vec{p}_i=[\vec{x}_i;\,\vec{u}_i]$, solve
\begin{align}\label{eq:localfit}
  (\hat{\mat{D}},\hat{\vec{c}})=\arg\min_{\mat{D},\vec{c}}
  \textstyle\sum_{i=1}^{M}w_i\|\vec{x}_{i+1}-\mat{D}\vec{p}_i-\vec{c}\|_2^2,\qquad
  \hat{\mat{D}}=[\hat{\mat{A}}_f\ \ \hat{\mat{B}}_f].
\end{align}
Weights $w_i>0$ decrease with $\|\vec{p}_i\|_2$. The intercept $\hat{\vec{c}}$ absorbs a small offset; the centered lifted
model keeps the target at zero.
\changliu{why not using direct Taylor expansion?}%
\tikreply{Direct Taylor linearization requires the plant equations or their derivatives. The local regression estimates the coefficients from logged transitions.}%

\textbf{Tangent-invariant parameterization.} Let $\mat{P}_{\perp}=\mat{I}-\Jtheta\Jtheta^+$ be the orthogonal projector onto
\changliu{I think it makes sense to let $\mat{J}$ carry the subscript $\theta$ throughout}%
\tikreply{The target lifting Jacobian is denoted by $\Jtheta$. In the multi-goal appendix, goal indices distinguish Jacobians evaluated at different equilibria.}%
the complement of the tangent space, where $\Jtheta^+$ is the Moore--Penrose pseudoinverse. TIK sets
\begin{align}\label{eq:ibc}
  \mat{A}(\theta,\mat{W})=\Jtheta\hat{\mat{A}}_f\Jtheta^{+}+\mat{W}\mat{P}_{\perp},
  \qquad
  \mat{B}(\theta)=\Jtheta\hat{\mat{B}}_f .
\end{align}
The fit determines $\mat{A}$ on the tangent space and $\mat{B}$ at the target. The trainable $\mat{W}$ determines the action
of $\mat{A}$ on the complement. In our experiments, we initialize $\theta$ and $\mat{W}$ from the encoder and $\mat{A}$ of
the same-seed Standard Koopman checkpoint. At each update, we recompute $\Jtheta$ and rebuild $\mat{A},\mat{B}$.
Automatic differentiation propagates prediction-loss gradients through $\Jtheta$ and its pseudoinverse to $\theta$.
The parameterization enforces matching to the fit at every update. A hinge penalty encourages the dynamics on the
orthogonal complement to be Schur, a condition we check for the LQR guarantee (\Cref{app:ibc}).
\changliu{isn't \eqref{eq:ibc} a parameterization? how is the differentiation work?}%
\tikreply{Equation \eqref{eq:ibc} defines the parameterization. Rebuilding the matrices enforces the identities. Automatic differentiation carries the loss gradient through the encoder Jacobian and pseudoinverse.}%

\textbf{State-dependent actuation.} The linear predictor has the same input matrix at every state. In swing-up, the effect
of a cart force on pole acceleration can change sign between the hanging and upright configurations. An optional bilinear
term~\citep{bruder2021bilinear} gives the predictor the input matrix
$\mat{B}+[\mat{N}_1\vec{z}\ \cdots\ \mat{N}_{n_u}\vec{z}]$ at state $\vec{z}$. This adds
$\sum_{j=1}^{n_u}u_j\mat{N}_j\vec{z}$ to the predicted update. The added term and its first derivatives vanish at the
centered target. \Cref{sec:exp_swingup} evaluates its effect on swing-up performance.
\changliu{I do not understand why we need the bilinear model.}%
\tikreply{The bilinear term gives state-dependent actuation away from the target. The swing-up ablation reports upright entry and full task success. Local LQR uses the centered linearization.}%

\begin{proposition}[Structural properties of TIK]\label{prop:ibc}
For any differentiable lifting at the target, \eqref{eq:ibc} satisfies:
(a)~$\mat{A}\Jtheta=\Jtheta\hat{\mat{A}}_f$ and $\mat{B}=\Jtheta\hat{\mat{B}}_f$, with
$\mathcal{E}=\Jtheta(\mat{A}_f-\hat{\mat{A}}_f)-\Jtheta(\mat{B}_f-\hat{\mat{B}}_f)\mat{K}\Jtheta$.
(b)~If $\mat{P}_{\perp}\mat{W}\mat{P}_{\perp}$ is Schur, $\mat{R}\succ0$,
$(\hat{\mat{A}}_f,\hat{\mat{B}}_f)$ is stabilizable, and $(\hat{\mat{A}}_f,\mat{Q}_x^{1/2})$ is detectable, then
$\mat{K}\Jtheta=\hat{\mat{K}}$, the physical LQR gain for $(\hat{\mat{A}}_f,\hat{\mat{B}}_f,\mat{Q}_x,\mat{R})$.
The deployed loop has Jacobian $\mat{F}=\mat{A}_f-\mat{B}_f\hat{\mat{K}}$.
(c)~The bilinear term preserves (a) and (b) for the local linearization and its LQR.
\end{proposition}

% Let $U_\perp$ have orthonormal columns spanning
% $\operatorname{range}(J_\theta)^\perp$ and set
% $A_\perp=U_\perp^\top A U_\perp$. For a scalar feedback multiplier $k$,
% \[
% \rho(A-kBK)=\max\left\{
%   \rho(\widehat A_f-k\widehat B_f\widehat K),\,
%   \rho(A_\perp)
% \right\}.
% \]
% The deployed spectral radius is $\rho(A_f-kB_f\widehat K)$.
% When $A_\perp$ is Schur, the latent and fitted physical matrices have
% the same Schur verdict at each $k$.

\Cref{prop:ibc} separates local stabilization from nonlinear prediction. TIK recovers the fitted physical LQR gain at the target.
The learned observables and bilinear terms model dynamics away from it. Any improvement over fitted LQR must therefore occur
beyond the local linear regime. The proof is in \Cref{app:mainproofs}.
\changliu{I think instead of giving a proof sketch here, it worths more high level explanation why this proposition is important.}%
\tikreply{The main text explains why the local gain comes from the physical fit and how the learned lift can help away from the target. The full proof is in the supplement.}%

\subsection{A Stability Certificate from Offline Data}\label{sec:dc}

Stability of the fitted closed loop does not establish plant stability. We bound the error of the fitted Jacobian along the deployed feedback
and compare it with the \emph{real stability radius}
$r_{\mathbb{R}}(\hat{\mat{F}})$, the smallest spectral norm of a real perturbation that places an eigenvalue on the unit
circle~\citep{qiu1995formula}. Below this radius, the fitted and true Jacobians have the same stability classification.

\begin{assumption}[Local curvature and noise bounds]\label{assum:curv}
The target and regressors $\vec{p}_i$ lie in a convex neighborhood $\mathcal{N}$ on which $f$ is twice continuously
differentiable. For $\vec{p}\in\mathcal{N}$, $\|D^2f(\vec{p})[\vec{a},\vec{b}]\|_2\le L\|\vec{a}\|_2\|\vec{b}\|_2$.
The regressors are exact and the measured successor obeys $\|\vec{x}_{i+1}-f(\vec{p}_i)\|_2\le\epsilon$.
\changliu{second-order continuous?}%
\tikreply{Yes. The assumption specifies twice continuously differentiable dynamics and bounds the Hessian on the fit neighborhood.}%
\end{assumption}

Write $\mat{D}_f=[\mat{A}_f\ \ \mat{B}_f]$. Taylor's theorem bounds the residual
$\vec{\ell}_i=\vec{x}_{i+1}-\mat{D}_f\vec{p}_i$ by $\varrho_i=\frac{L}{2}\|\vec{p}_i\|_2^2+\epsilon$.
Let $\vec{q}_i=[\vec{p}_i;1]$. The weighted fit requires $\mat{\Sigma}=\sum_iw_i\vec{q}_i\vec{q}_i\trans\succ0$.
Let $\vec{g}_{i,D}$ be the first $n_x+n_u$ entries of $\mat{\Sigma}^{-1}\vec{q}_i$. The normal equations and the triangle inequality give
\begin{align}\label{eq:fitbound}
  \|(\hat{\mat{D}}-\mat{D}_f)\mat{\Gamma}\|_2
  \le \eta(\mat{\Gamma})
  := \textstyle\sum_{i=1}^{M} w_i\varrho_i\|\mat{\Gamma}\trans\vec{g}_{i,D}\|_2 .
\end{align}
The bound holds for each dimensionally compatible $\mat{\Gamma}$. We use $\mat{\Gamma}=[\mat{I}_{n_x};-\mat{K}_o]$ to bound
error along the state-input perturbations induced by the deployed policy, rather than over all input directions.
\changliu{does it need to be a unit matrix? and for $\eta$ definition, does it need to be the uppoer bound over all $\mat{\Gamma}$?}%
\tikreply{The inequality holds for each conformable multiplier, including non-unit matrices. The certificate evaluates it only for the multiplier induced by the deployed gain.}%

\begin{theorem}[Stability certificate from offline data]\label{thm:knownfp}
Suppose the deployed policy preserves the target equilibrium, is unsaturated near it, and has Jacobian $-\mat{K}_o$ there. For lifted LQR,
$\mat{K}_o=\mat{K}\Jtheta$, which equals $\hat{\mat{K}}$ for TIK under \Cref{prop:ibc}(b). Let
$\hat{\mat{F}}=\hat{\mat{D}}\mat{\Gamma}$ have no eigenvalue on the unit circle. Under \Cref{assum:curv} and the full-rank
fit condition, if $\eta(\mat{\Gamma})<r_{\mathbb{R}}(\hat{\mat{F}})$, the target is locally exponentially stable on the
plant when $\rho(\hat{\mat{F}})<1$ and unstable when $\rho(\hat{\mat{F}})>1$.
\end{theorem}

The test is inconclusive if the strict inequality cannot be established. A verified lower bound on the radius is sufficient.
Formal certification requires valid curvature and noise bounds; our experimental curvature estimates are empirical. \Cref{app:mainproofs} gives the proof, a conservative numerical
test, and extensions to noisy regressors and cubic remainders.
\subsection{Koopman MPC and a Certified Terminal Set}\label{sec:kmpc}\label{sec:rmpc_main}

\begin{proposition}[Koopman MPC near the target]\label{prop:kmpc}
Consider \eqref{eq:kmpc_qp} with $\mat{R}\succ0$, stabilizing Riccati terminal weight $\mat{P}$, and the target in the strict
interior of the constraints. Near the target, where no constraint is active, the MPC law equals the Koopman LQR law
$\vec{u}=-\mat{K}\Phi_\theta(\vec{x})$ for all $N_h\ge1$. Under \Cref{prop:ibc}(b), TIK's deployed MPC loop has Jacobian
$\mat{F}=\mat{A}_f-\mat{B}_f\hat{\mat{K}}$, and \Cref{thm:knownfp} applies.
\end{proposition}

Local regulation uses the convex linear-model QP in \eqref{eq:kmpc_qp} (\Cref{app:mainproofs}).
For large maneuvers, we use the bilinear predictor with model predictive path integral control
(MPPI)~\citep{williams2017information} or the QP-MPC variant in \Cref{app:dcpdual}.
On entry to $\Omega_c$, the controller switches to $\vec{u}=-\hat{\mat{K}}\vec{x}$.

\begin{theorem}[Certified invariant terminal set]\label{thm:region_main}
Let $\hat{\mat{F}}$ be Schur, $\mat{S}\succ0$ solve $\hat{\mat{F}}\trans\mat{S}\hat{\mat{F}}-\mat{S}=-\mat{Q}$ for $\mat{Q}\succ0$,
and set $\hat\vartheta=\|\mat{S}^{1/2}\hat{\mat{F}}\mat{S}^{-1/2}\|_2<1$ and $\kappa=\lambda_{\max}(\mat{S})/\lambda_{\min}(\mat{S})$.
Suppose $g_{\mathrm{loc}}(\vec{x})=f(\vec{x},-\hat{\mat{K}}\vec{x})$ satisfies $\|\mat{F}-\hat{\mat{F}}\|_2\le\eta$ and
$\|g_{\mathrm{loc}}(\vec{x})-\mat{F}\vec{x}\|_2\le\frac12L_H\|\vec{x}\|_2^2+\frac16L_T\|\vec{x}\|_2^3$ on a convex set
$\mathcal{N}$. Choose $r>0$ such that
$\hat\vartheta+\sqrt\kappa(\eta+\tfrac12L_Hr+\tfrac16L_Tr^2)\le1$.
Every ellipsoid $\Omega_c=\{\vec{x}:\vec{x}\trans\mat{S}\vec{x}\le c\}$ with $c\le\lambda_{\min}(\mat{S})r^2$ contained in
$\mathcal{N}$ and satisfying the state and local feedback input limits is forward invariant under $g_{\mathrm{loc}}$.
If the bound is strict, trajectories in $\Omega_c$ converge exponentially to $\vec{0}$.
\end{theorem}

After entry into $\Omega_c$, the local controller keeps subsequent sampled states and inputs within their limits under
the stated bounds. The theorem guarantees neither safe entry nor safety between sampling instants.
\Cref{app:rmpc} gives the proof and the Lyapunov-metric formulation.

\section{Experiments}\label{sec:experiments}\label{sec:exp_setup}

All experiments use simulated plants (\Cref{app:setup}), and all feedback is tested on the nonlinear simulator (\Cref{app:protocol}). \emph{Oracle} LQR and linear MPC use
the simulator's target Jacobians, obtained by finite differences of its one-step map, as known-dynamics references. \emph{Fitted} linear LQR/MPC and TIK instead estimate
$(\hat{\mat{A}}_f,\hat{\mat{B}}_f)$ from logged transitions \eqref{eq:localfit}; neural NMPC linearizes its learned predictor at each step (\Cref{app:training}).
Simulator derivatives provide the stability reference and the oracle models; \Cref{app:training} describes controller tuning. The upright Koopman MPC comparison uses $N_h=20$ and identity LQR weights.
\changliu{what is oracle linearization?}%
\tikreply{Oracle means the controller receives derivatives of the simulator plant map. Fitted linear MPC and TIK estimate their dynamics from logged transitions.}%

\subsection{Does the Nominal Certificate Hold on the True System? (Q1)}\label{sec:exp_fixedpoint}\label{sec:exp_ibc}\label{sec:exp_kmpc}

\begin{table}[tbp]
\caption{\textbf{Local plant stability at the upright target.} Plant-stable models among those with a Schur latent closed loop (Koopman rows)
or among evaluated models (oracle MPC, neural NMPC); $10$ seeds per learned method, one oracle per system. A dash indicates that no run yielded a
stabilizing LQR gain.}

\label{tab:classes}\label{tab:kmpc}
\centering\small\setlength{\tabcolsep}{2.2pt}
\renewcommand{\arraystretch}{0.92}
\begin{tabular}{@{}lcccc@{}}
\toprule
 & \multicolumn{4}{c}{Plant-stable models (counts)} \\
\cmidrule(l){2-5}
Controller & \shortstack{Cart-pole\\(ODE)} & \shortstack{Cart-pole\\(MuJoCo)} & \shortstack{Double\\cart-pole} & \shortstack{Double\\pendulum} \\
\midrule
Oracle linear MPC & $\pnKcCpRefS/\pnKcCpRefN$ & $\pnKcMcpRefS/\pnKcMcpRefN$ & $\pnKcDcpRefS/\pnKcDcpRefN$ & $\pnKcDpuRefS/\pnKcDpuRefN$ \\
Neural model + NMPC & $\pnKcCpNnS/\pnKcCpNnN$ & $\pnKcMcpNnS/\pnKcMcpNnN$ & $\pnKcDcpNnS/\pnKcDcpNnN$ & $\pnKcDpuNnS/\pnKcDpuNnN$ \\
\midrule
Standard Koopman & $\pnKcCpStdS/\pnKcCpStdN$ & $\pnKcMcpStdS/\pnKcMcpStdN$ & $\pnKcDcpStdS/\pnKcDcpStdN$ & $\pnKcDpuStdS/\pnKcDpuStdN$ \\
Koopman + invariance penalty & $\pnKcCpPenS/\pnKcCpPenN$ & $\pnKcMcpPenS/\pnKcMcpPenN$ & $\pnKcDcpPenS/\pnKcDcpPenN$ & $\pnKcDpuPenS/\pnKcDpuPenN$ \\
EDMD~\citep{korda2018linear} & $\pnKcCpKmS/\pnKcCpKmN$ & $\pnKcMcpKmS/\pnKcMcpKmN$ & -- & $\pnKcDpuKmS/\pnKcDpuKmN$ \\
Deep Koopman with control~\citep{shi2022deep} & $\pnKcCpDkucS/\pnKcCpDkucN$ & $\pnKcMcpDkucS/\pnKcMcpDkucN$ & $\pnKcDcpDkucS/\pnKcDcpDkucN$ & $\pnKcDpuDkucS/\pnKcDpuDkucN$ \\
Stable operator~\citep{mamakoukas2023learning} & $\pnKcCpSocS/\pnKcCpSocN$ & $\pnKcMcpSocS/\pnKcMcpSocN$ & $\pnKcDcpSocS/\pnKcDcpSocN$ & $\pnKcDpuSocS/\pnKcDpuSocN$ \\
Stable embedding~\citep{fan2024stablekoopman} & $\pnKcCpSkelS/\pnKcCpSkelN$ & $\pnKcMcpSkelS/\pnKcMcpSkelN$ & $\pnKcDcpSkelS/\pnKcDcpSkelN$ & $\pnKcDpuSkelS/\pnKcDpuSkelN$ \\
\midrule
\textbf{TIK (ours)} & $\pnKcCpTikS/\pnKcCpTikN$ & $\pnKcMcpTikS/\pnKcMcpTikN$ & $\pnKcDcpTikS/\pnKcDcpTikN$ & $\pnKcDpuTikS/\pnKcDpuTikN$ \\
\bottomrule
\end{tabular}
\end{table}
\changliu{does it imply the model is not controllable?}%
\tikreply{A failed Riccati solve alone cannot establish lack of controllability. Stabilizability, detectability, and numerical conditioning also matter.}%

All $40$ Standard Koopman models have Schur latent closed loops but unstable physical Jacobians at the target.
All $40$ TIK models have Schur physical Jacobians (\Cref{tab:classes}). The soft penalty stabilizes all single cart-pole models and
seven double-pendulum models, but no double cart-pole model. Neither stable-by-construction variant stabilizes any of its $40$ models.
EDMD stabilizes nine double-pendulum models; on the double cart-pole, its Riccati solver found no stabilizing gain (this does not prove it uncontrollable).
\changliu{all?}%
\tikreply{Yes. All ten double cart-pole models with the soft penalty fail in the centered experiment. The supplementary penalty sweep evaluates models at their own equilibria.}%
\changliu{this notation is hard to interpret}%
\tikreply{Each stable-by-construction variant stabilizes zero of its 40 models.}%

\textbf{Prediction and sensitivity to feedback.} Standard Koopman and the soft-penalty model have nearly identical $14$-step
state errors on the ODE cart-pole ($\pnInvCartPoleZeroRoll$ and $\pnInvCartPoleOneRoll$), but differ in plant stability.
The neural model yields lower prediction error on the double cart-pole and double pendulum, yet only one of its $20$ NMPC
controllers is locally stable. To test whether nominal stability remains informative as feedback strength changes, we scale
the nominal gain by $\pnPrNk$ values $k\in[\pnPrKmin,\pnPrKmax]$. Nominal and deployed verdicts agree for
$\pnPrUncAgree\%$ of Standard Koopman model--gain pairs and $\pnPrIbcAgree\%$ for TIK (\Cref{fig:predict} in \Cref{app:fixedpoint_more}).
\changliu{why do we want to scale $\mat{K}$ by $k\in[\pnPrKmin,\pnPrKmax]$?}%
\tikreply{Scaling the gain tests whether the latent stability verdict remains predictive over different feedback strengths.}%

\textbf{Recovery from initial tilts.} TIK uses the fitted physical LQR gain (\Cref{prop:ibc}) and matches oracle LQR's recovery outcome in $\pnCatchMatch$ of $\pnCatchCells$ model--tilt tests across
$\pnCatchModels$ models, versus $\pnCatchStdLin$ for linearized Standard Koopman (\Cref{fig:roa}; protocol in \Cref{app:protocol}). This comparison uses local linear feedback. For larger tilts, we start from the offline TIK checkpoint
\changliu{it is still unclear how oracle LQR is implemented}%
\tikreply{Oracle LQR uses the simulator target Jacobians and applies the resulting gain to the nonlinear simulator. Fitted LQR uses the logged-data estimate.}%
and add rollouts of its lifted policy from tilted starts to the training data. The median recovery range increased from $0.06$ to $0.15$ rad on
double cart-pole and from $0.40$ to $1.50$ rad on double pendulum. All $\pnDagTotStable$ completed
\changliu{is it fine-tuned on standard Koopman but using TIK structure?}%
\tikreply{Initial TIK training starts from the Standard Koopman checkpoint with the TIK parameterization. Later recovery training starts from that TIK checkpoint and adds closed-loop rollouts while retaining tangent matching.}%
TIK runs yield stable controllers. One of the $40$ seeds diverged during training. The same training without tangent matching yields $\pnDagCtrlStable$ stable controllers out of $\pnDagCtrlN$ (\Cref{app:dagger}).

\subsection{Offline Stability Verification (Q2)}\label{sec:exp_datacert}\label{sec:datacert_exp}

\begin{table}[tbp]
\caption{\textbf{Empirical offline stability-margin tests.} Controllers passing the estimated margin test in \Cref{thm:knownfp}
($20$ per method per row). Simulator Jacobians classify zero Standard Koopman and all $20$ TIK controllers as locally stable.}

\label{tab:ibccert}\label{tab:ibccert_full}
\centering\small\setlength{\tabcolsep}{2.6pt}
\renewcommand{\arraystretch}{0.90}
\begin{tabular}{@{}lllcc@{}}
\toprule
 & & & \multicolumn{2}{c}{Passed estimated margin test} \\
\cmidrule(lr){4-5}
System & Offline data regime & Noise $\epsilon$ & Standard Koopman & \textbf{TIK} \\
\midrule
Cart-pole & Near target ($400$ pts) & $0$ & $0/20$ & $\pnICDefZBilStab/20$ \\
(upright) & Two scales & $0$ & $0/20$ & $\pnICDefZTBilStab/20$ \\
 & Two scales, noisy states & $10^{-7}$ & $0/20$ & $\pnICDsgSevenTVBilStab/20$ \\
 & Closed-loop trajectories & $3\times10^{-8}$ & $0/20$ & $\pnICDsgThEightTJBilStab/20$ \\
\midrule
Double cart-pole & Two scales & $0$ & $0/20$ & $\pnIDDefZStab/20$ \\
(upright) & Closed-loop trajectories & $10^{-10}$ & $0/20$ & $\pnIDDsgTenJStab/20$ \\
 & Two scales (high noise) & $10^{-6}$ & $0/20$ & $\pnIDDsgSixStab/20$ \\
\bottomrule
\end{tabular}
\end{table}

All $20$ TIK controllers pass the estimated margin test in six regimes; none passes in the high-noise double cart-pole
regime ($\epsilon=10^{-6}$), although all have Schur physical Jacobians. No Standard Koopman controller passes.
These are empirical checks, not formal certificates: the curvature estimate upper-bounds the simulator curvature in only $80$ of
$100$ near-target cart-pole fits (\Cref{tab:ibccert_checks}).

\changliu{all 20?}%
\tikreply{The 20 refers to controllers per method per main-table row. The supplementary coverage table reports 100 fits per cart-pole regime and 40 per double cart-pole regime.}%
\changliu{should this refer to table 2?}%
\tikreply{Yes. The main stability table reports the decisions and identifies the high-noise regime. The supplementary table reports coverage against the simulator.}%

\subsection{Constrained Control and Large-Amplitude Maneuvers (Q3)}\label{sec:exp_bench}\label{sec:exp_rmpc}\label{sec:exp_swingup}\label{sec:exp_recover}

\begin{table}[tbp]
\caption{\textbf{Dual-mode MPC for constrained cart-pole swing-up and recovery across compute budgets.} Share of rollouts
($\pnSrSwN$ swing-up and $\pnSrRecN$ recovery starts per model; $20$ TIK and Standard Koopman models, $10$ neural, $1$ fitted linear) entering their terminal set and stabilizing within force and rail
limits. Sets $\Omega_c$ are estimated using \Cref{thm:region}; formal guarantees require valid bounds. A dash indicates an exceeded budget.}
\label{tab:swrmpc}
\centering\small\setlength{\tabcolsep}{2.8pt}
\renewcommand{\arraystretch}{0.90}
\begin{tabular}{@{}lrrrrrrrr@{}}
\toprule
 & \multicolumn{4}{c}{Swing-up success (\%)} & \multicolumn{4}{c}{Recovery success (\%)} \\
\cmidrule(lr){2-5}\cmidrule(l){6-9}
Predictor and terminal set & $2$\,ms & $1$\,ms & $0.5$\,ms & $0.25$\,ms & $2$\,ms & $1$\,ms & $0.5$\,ms & $0.25$\,ms \\
\midrule
Fitted linear, estimated $\Omega_c$ & $\pnSrLinTwoSw$ & $\pnSrLinOneSw$ & $\pnSrLinHalfSw$ & $\pnSrLinQuarterSw$ & $\pnSrLinTwoRec$ & $\pnSrLinOneRec$ & $\pnSrLinHalfRec$ & $\pnSrLinQuarterRec$ \\
Neural, estimated $\Omega_c$ & $\pnSrMlpTwoSw$ & $\pnSrMlpOneSw$ & $\pnSrMlpHalfSw$ & -- & $\pnSrMlpTwoRec$ & $\pnSrMlpOneRec$ & $\pnSrMlpHalfRec$ & -- \\
Standard Koopman, latent terminal set & $\pnSrFreeTwoSw$ & $\pnSrFreeOneSw$ & $\pnSrFreeHalfSw$ & $\pnSrFreeQuarterSw$ & $\pnSrFreeTwoRec$ & $\pnSrFreeOneRec$ & $\pnSrFreeHalfRec$ & $\pnSrFreeQuarterRec$ \\
\midrule
\textbf{TIK}, estimated $\Omega_c$ & $\pnSrTikTwoSw$ & $\pnSrTikOneSw$ & $\pnSrTikHalfSw$ & $\pnSrTikQuarterSw$ & $\pnSrTikTwoRec$ & $\pnSrTikOneRec$ & $\pnSrTikHalfRec$ & $\pnSrTikQuarterRec$ \\
\bottomrule
\end{tabular}
\end{table}

\textbf{Cart-pole swing-up and recovery.} \Cref{tab:swrmpc} tests tilts up to $135^\circ$ and kicks up to $6$\,rad/s.
Terminal sets use the Lyapunov-metric formulation in \Cref{thm:region}. All $\pnSrPostN$ rollouts entering $\Omega_c$
converged without subsequent violations. TIK achieves $\pnSrTikTwoSw\%$ swing-up success at $2$\,ms and
$\pnSrTikQuarterSw\%$ at $0.25$\,ms, where neural MPPI exceeds the budget. At $1$ and $0.5$\,ms, neural MPPI achieves
$\pnSrMlpOneSw\%$ and TIK achieves $\pnSrTikOneSw\%$. Tuning details are in \Cref{app:training}.

\textbf{Double cart-pole QP-MPC.} Trained on the original data only, TIK succeeds in $\pnDmOld\%$ of rollouts. After adding near-target and closed-loop data it
reaches $\pnDmTikAll\%$, compared with $0\%$ for Standard Koopman and $\pnDmCtrl\%$ for the same-data control without tangent matching
(\Cref{app:dcpdual}). TIK remains below fitted linear MPC and fitted LQR ($\pnDmLin\%$ and $\pnDmCatch\%$). Most tested starts can already be recovered by the fitted local controller.

\textbf{Linear versus bilinear TIK.} In the cart-pole ablation, the MPPI-only swing-up success rate is $\pnSwILinPlan\%$ with linear input and $\pnSwIBilPlan\%$ with
bilinear input (upright entry: $\pnSwILinPlanReach\%$ and $\pnSwIBilPlanReach\%$). With separate local balancing controllers, the bilinear model reaches
$\pnSwIBilLifted\%$ overall success with lifted LQR and $\pnSwIBilData\%$ with fitted physical LQR. Their gains agree at the target, but their actions can differ away from it (\Cref{app:recover}). Without tangent matching, its lifted LQR policy is
unstable on the plant for all $20$ seeds (\Cref{tab:ibc}).
\changliu{check the table being referenced, should it be table 3?}%
\tikreply{The compute-budget comparison is in \Cref{tab:swrmpc}. The linear-versus-bilinear ablation uses a separate protocol and points to \Cref{tab:ibc}.}%

\subsection{Real-Time Control on Simulated Manipulators (Q4)}\label{sec:exp_robots}\label{app:robots}

The simulated seven-joint arms test control under short planning budgets (\Cref{tab:robots}). On Kinova Gen3 obstacle avoidance~\citep{sun2025spark}, TIK, fitted linear MPC,
and Standard Koopman each reach $100\%$ success at $5$\,ms; the median TIK solve takes $\pnArmFiveIbcMsMed$\,ms. On the six-goal torque-controlled arm, which uses one
predictor across goals with distinct linearizations, TIK reaches $\pnMrOursOneSucc\%$ at $1$\,ms, matching workspace-center linear MPC ($\pnMrLinOneSucc\%$) and exceeding neural NMPC ($\pnMrSlOneSucc\%$) and Standard Koopman ($\pnMrStdOneSucc\%$). This benchmark shows no success-rate advantage over linear MPC (\Cref{tab:multipoint}).

\section{Conclusion and Limitations}\label{sec:conclusion}

A Koopman controller that is stable on its latent model is also locally stable on the robot when the latent closed loop
reproduces the physical closed-loop Jacobian on the tangent space of the embedding. TIK builds this matching into the model
with a local linearization fitted from offline data; the stability test bounds the remaining error between this fit and the
plant. Near the target, TIK applies the LQR gain of this fit; away from it, the bilinear predictor models state-dependent actuation.

The formal certificates require valid curvature and noise bounds and a differentiable local model; our experiments use empirical curvature
estimates in simulation. The terminal-set result covers the local controller after entry. The effects of hardware noise and contact transitions require further study.

\subsection*{Generative AI Use}
The authors used generative AI tools to polish the writing and to help with the proofs of the theorems. The authors reviewed and verified all AI-assisted content, and they are responsible for the accuracy and integrity of the submitted work.

\bibliography{main}

\begin{thebibliography}{35}
\providecommand{\natexlab}[1]{#1}
\providecommand{\url}[1]{\texttt{#1}}
\expandafter\ifx\csname urlstyle\endcsname\relax
  \providecommand{\doi}[1]{doi: #1}\else
  \providecommand{\doi}{doi: \begingroup \urlstyle{rm}\Url}\fi

\bibitem[Abuduweili and Liu(2024)]{abuduweili2024lipschitz}
Abulikemu Abuduweili and Changliu Liu.
\newblock Estimating neural network robustness via {Lipschitz} constant and
  architecture sensitivity.
\newblock In \emph{CoRL Workshop on Safe and Robust Robot Learning for
  Operation in the Real World}, 2024.

\bibitem[Abuduweili et~al.(2026)Abuduweili, Pang, Li, and
  Liu]{abuduweili2025scaling}
Abulikemu Abuduweili, Yuyang Pang, Feihan Li, and Changliu Liu.
\newblock Scaling law of neural {K}oopman operators.
\newblock \emph{arXiv preprint arXiv:2602.19943}, 2026.

\bibitem[Bruder et~al.(2021)Bruder, Fu, and Vasudevan]{bruder2021bilinear}
Daniel Bruder, Xun Fu, and Ram Vasudevan.
\newblock Advantages of bilinear {Koopman} realizations for the modeling and
  control of systems with unknown dynamics.
\newblock \emph{IEEE Robotics and Automation Letters}, 6\penalty0 (3):\penalty0
  4369--4376, 2021.

\bibitem[Brunton et~al.(2016)Brunton, Brunton, Proctor, and
  Kutz]{brunton2016invariant}
Steven~L. Brunton, Bingni~W. Brunton, Joshua~L. Proctor, and J.~Nathan Kutz.
\newblock {K}oopman invariant subspaces and finite linear representations of
  nonlinear dynamical systems for control.
\newblock \emph{PLoS ONE}, 11\penalty0 (2):\penalty0 e0150171, 2016.

\bibitem[Chen et~al.(2024)Chen, Abuduweili, Agrawal, Han, Ravichandar, Liu, and
  Ichnowski]{chen2024korol}
Hongyi Chen, Abulikemu Abuduweili, Aviral Agrawal, Yunhai Han, Harish
  Ravichandar, Changliu Liu, and Jeffrey Ichnowski.
\newblock {KOROL}: Learning visualizable object feature with {Koopman} operator
  rollout for manipulation.
\newblock In \emph{Conference on Robot Learning (CoRL)}, 2024.

\bibitem[Elaydi(2005)]{elaydi2005difference}
Saber Elaydi.
\newblock \emph{An Introduction to Difference Equations}.
\newblock Springer, 3 edition, 2005.
\newblock \doi{10.1007/0-387-27602-5}.

\bibitem[Fan et~al.(2026)Fan, Yi, Rye, Shi, and
  Manchester]{fan2024stablekoopman}
Fletcher Fan, Bowen Yi, David Rye, Guodong Shi, and Ian~R. Manchester.
\newblock Learning stable {K}oopman embeddings for identification and control.
\newblock \emph{Automatica}, 189:\penalty0 113003, 2026.
\newblock \doi{10.1016/j.automatica.2026.113003}.

\bibitem[Ge et~al.(2026)Ge, Wang, Jia, Han, Shi, Zhou, and
  Huang]{ge2026koopman}
Haizhou Ge, Mengshuo Wang, Yufei Jia, Lei Han, Lu~Shi, Guyue Zhou, and Ruqi
  Huang.
\newblock Why {K}oopman structure does not help differentiable world models for
  humanoid whole-body tracking.
\newblock \emph{arXiv preprint arXiv:2611.00951}, 2026.

\bibitem[Jain(2026)]{jain2026verification}
Tushar Jain.
\newblock Data-driven stability verification and region of attraction
  estimation for {K}oopman-lifted nonlinear systems.
\newblock \emph{IFAC Journal of Systems and Control}, 36:\penalty0 100424,
  2026.
\newblock \doi{10.1016/j.ifacsc.2026.100424}.

\bibitem[Jung et~al.(2026)Jung, Abuduweili, Li, and Liu]{jung2026wholebody}
Sebin Jung, Abulikemu Abuduweili, Jiaxing Li, and Changliu Liu.
\newblock Whole-body safe control of robotic systems with {K}oopman neural
  dynamics.
\newblock In \emph{IEEE International Conference on Robotics and Automation
  (ICRA)}, pages 23949--23956, 2026.

\bibitem[Khalil(2002)]{khalil2002nonlinear}
Hassan~K. Khalil.
\newblock \emph{Nonlinear Systems}.
\newblock Prentice Hall, 3rd edition, 2002.

\bibitem[Koopman(1931)]{koopman1931hamiltonian}
Bernard~O Koopman.
\newblock Hamiltonian systems and transformation in {H}ilbert space.
\newblock \emph{Proceedings of the National Academy of Sciences}, 17\penalty0
  (5):\penalty0 315--318, 1931.

\bibitem[Korda and Mezi{\'c}(2018)]{korda2018linear}
Milan Korda and Igor Mezi{\'c}.
\newblock Linear predictors for nonlinear dynamical systems: {K}oopman operator
  meets model predictive control.
\newblock \emph{Automatica}, 93:\penalty0 149--160, 2018.

\bibitem[Li et~al.(2025)Li, Abuduweili, Sun, Chen, Zhao, and
  Liu]{li2025continual}
Feihan Li, Abulikemu Abuduweili, Yifan Sun, Rui Chen, Weiye Zhao, and Changliu
  Liu.
\newblock Continual learning and lifting of {Koopman} dynamics for linear
  control of legged robots.
\newblock In \emph{Learning for Dynamics and Control Conference (L4DC)}, volume
  283 of \emph{PMLR}, 2025.

\bibitem[Liu et~al.(2025)Liu, Ozay, and Sontag]{liu2025immersions}
Zexiang Liu, Necmiye Ozay, and Eduardo~D. Sontag.
\newblock Properties of immersions for systems with multiple limit sets with
  implications to learning {K}oopman embeddings.
\newblock \emph{Automatica}, 2025.

\bibitem[Lusch et~al.(2018)Lusch, Kutz, and Brunton]{lusch2018deep}
Bethany Lusch, J~Nathan Kutz, and Steven~L Brunton.
\newblock Deep learning for universal linear embeddings of nonlinear dynamics.
\newblock \emph{Nature Communications}, 9\penalty0 (1):\penalty0 4950, 2018.

\bibitem[Mamakoukas et~al.(2023)Mamakoukas, Abraham, and
  Murphey]{mamakoukas2023learning}
Giorgos Mamakoukas, Ian Abraham, and Todd~D. Murphey.
\newblock Learning stable models for prediction and control.
\newblock \emph{IEEE Transactions on Robotics}, 39\penalty0 (3):\penalty0
  2255--2275, 2023.

\bibitem[Mauroy and Mezi{\'c}(2016)]{mauroy2016global}
Alexandre Mauroy and Igor Mezi{\'c}.
\newblock Global stability analysis using the eigenfunctions of the {Koopman}
  operator.
\newblock \emph{IEEE Transactions on Automatic Control}, 61\penalty0
  (11):\penalty0 3356--3369, 2016.
\newblock \doi{10.1109/TAC.2016.2518918}.

\bibitem[N{\"u}ske et~al.(2023)N{\"u}ske, Peitz, Philipp, Schaller, and
  Worthmann]{nuske2023finite}
Feliks N{\"u}ske, Sebastian Peitz, Friedrich Philipp, Manuel Schaller, and Karl
  Worthmann.
\newblock Finite-data error bounds for {K}oopman-based prediction and control.
\newblock \emph{Journal of Nonlinear Science}, 33\penalty0 (1):\penalty0 14,
  2023.

\bibitem[Ortega and Rheinboldt(1970)]{ortega1970iterative}
James~M. Ortega and Werner~C. Rheinboldt.
\newblock \emph{Iterative Solution of Nonlinear Equations in Several
  Variables}.
\newblock Academic Press, 1970.

\bibitem[Proctor et~al.(2018)Proctor, Brunton, and
  Kutz]{proctor2018generalizing}
Joshua~L Proctor, Steven~L Brunton, and J~Nathan Kutz.
\newblock Generalizing {K}oopman theory to allow for inputs and control.
\newblock \emph{SIAM Journal on Applied Dynamical Systems}, 17\penalty0
  (1):\penalty0 909--930, 2018.

\bibitem[Qiu et~al.(1995)Qiu, Bernhardsson, Rantzer, Davison, Young, and
  Doyle]{qiu1995formula}
Li~Qiu, Bo~Bernhardsson, Anders Rantzer, Edward~J. Davison, Peter~M. Young, and
  John~C. Doyle.
\newblock A formula for computation of the real stability radius.
\newblock \emph{Automatica}, 31\penalty0 (6):\penalty0 879--890, 1995.

\bibitem[Shang et~al.(2026{\natexlab{a}})Shang, Cort{\'e}s, and
  Zheng]{shang2025expstab}
Xu~Shang, Jorge Cort{\'e}s, and Yang Zheng.
\newblock On the exponential stability of {K}oopman model predictive control.
\newblock In \emph{Proceedings of the 8th Annual Learning for Dynamics and
  Control Conference}, volume 331 of \emph{Proceedings of Machine Learning
  Research}, pages 1694--1712, 2026{\natexlab{a}}.
\newblock URL \url{https://proceedings.mlr.press/v331/shang26a.html}.

\bibitem[Shang et~al.(2026{\natexlab{b}})Shang, Haseli, Cort{\'e}s, and
  Zheng]{shang2026existence}
Xu~Shang, Masih Haseli, Jorge Cort{\'e}s, and Yang Zheng.
\newblock On the existence of {Koopman} linear embeddings for controlled
  nonlinear systems.
\newblock \emph{arXiv preprint arXiv:2602.14537}, 2026{\natexlab{b}}.

\bibitem[Shi and Meng(2022)]{shi2022deep}
Haojie Shi and Max Q-H Meng.
\newblock Deep {K}oopman operator with control for nonlinear systems.
\newblock \emph{IEEE Robotics and Automation Letters}, 7\penalty0 (3):\penalty0
  7700--7707, 2022.

\bibitem[Str{\"a}sser et~al.(2025)Str{\"a}sser, Schaller, Worthmann, Berberich,
  and Allg{\"o}wer]{strasser2023feedback}
Robin Str{\"a}sser, Manuel Schaller, Karl Worthmann, Julian Berberich, and
  Frank Allg{\"o}wer.
\newblock {K}oopman-based feedback design with stability guarantees.
\newblock \emph{IEEE Transactions on Automatic Control}, 70\penalty0
  (1):\penalty0 355--370, 2025.

\bibitem[Str{\"a}sser et~al.(2026{\natexlab{a}})Str{\"a}sser, Schaller,
  Worthmann, Berberich, and Allg{\"o}wer]{strasser2024safedmd}
Robin Str{\"a}sser, Manuel Schaller, Karl Worthmann, Julian Berberich, and
  Frank Allg{\"o}wer.
\newblock {SafEDMD}: A {K}oopman-based data-driven controller design framework
  for nonlinear dynamical systems.
\newblock \emph{Automatica}, 185:\penalty0 112732, 2026{\natexlab{a}}.
\newblock \doi{10.1016/j.automatica.2025.112732}.

\bibitem[Str{\"a}sser et~al.(2026{\natexlab{b}})Str{\"a}sser, Worthmann,
  Mezi{\'c}, Berberich, Schaller, and Allg{\"o}wer]{strasser2025overview}
Robin Str{\"a}sser, Karl Worthmann, Igor Mezi{\'c}, Julian Berberich, Manuel
  Schaller, and Frank Allg{\"o}wer.
\newblock An overview of {K}oopman-based control: {F}rom error bounds to
  closed-loop guarantees.
\newblock \emph{Annual Reviews in Control}, 61:\penalty0 101035,
  2026{\natexlab{b}}.
\newblock \doi{10.1016/j.arcontrol.2025.101035}.

\bibitem[Sun et~al.(2025)Sun, Chen, Yun, Fang, Jung, Li, Li, Zhao, and
  Liu]{sun2025spark}
Yifan Sun, Rui Chen, Kai~S. Yun, Yikuan Fang, Sebin Jung, Feihan Li, Bowei Li,
  Weiye Zhao, and Changliu Liu.
\newblock {SPARK}: {S}afe {P}rotective and {A}ssistive {R}obot {K}it.
\newblock In \emph{IFAC Symposium on Robotics}, 2025.

\bibitem[van Goor et~al.(2023)van Goor, Mahony, Schaller, and
  Worthmann]{vangoor2023reprojection}
Pieter van Goor, Robert Mahony, Manuel Schaller, and Karl Worthmann.
\newblock Reprojection methods for {K}oopman-based modelling and prediction.
\newblock In \emph{IEEE Conference on Decision and Control (CDC)}, pages
  315--321, 2023.

\bibitem[Wabersich and Zeilinger(2018)]{wabersich2018linear}
Kim~P. Wabersich and Melanie~N. Zeilinger.
\newblock Linear model predictive safety certification for learning-based
  control.
\newblock In \emph{IEEE Conference on Decision and Control}, pages 7130--7135,
  2018.

\bibitem[Wang and Manchester(2023)]{wang2023direct}
Ruigang Wang and Ian~R. Manchester.
\newblock Direct parameterization of {L}ipschitz-bounded deep networks.
\newblock In \emph{International Conference on Machine Learning (ICML)}, 2023.

\bibitem[Williams et~al.(2017)Williams, Wagener, Goldfain, Drews, Rehg, Boots,
  and Theodorou]{williams2017information}
Grady Williams, Nolan Wagener, Brian Goldfain, Paul Drews, James~M. Rehg, Byron
  Boots, and Evangelos~A. Theodorou.
\newblock Information theoretic {MPC} for model-based reinforcement learning.
\newblock In \emph{IEEE International Conference on Robotics and Automation
  (ICRA)}, pages 1714--1721, 2017.

\bibitem[Williams et~al.(2015)Williams, Kevrekidis, and
  Rowley]{williams2015edmd}
Matthew~O Williams, Ioannis~G Kevrekidis, and Clarence~W Rowley.
\newblock A data-driven approximation of the {K}oopman operator: Extending
  dynamic mode decomposition.
\newblock \emph{Journal of Nonlinear Science}, 25\penalty0 (6):\penalty0
  1307--1346, 2015.

\bibitem[Yuan et~al.(2022)Yuan, Hall, Zhou, Brunke, Greeff, Panerati, and
  Schoellig]{yuan2022safecontrolgym}
Zhaocong Yuan, Adam~W. Hall, Siqi Zhou, Lukas Brunke, Melissa Greeff, Jacopo
  Panerati, and Angela~P. Schoellig.
\newblock safe-control-gym: A unified benchmark suite for safe learning-based
  control and reinforcement learning in robotics.
\newblock \emph{IEEE Robotics and Automation Letters}, 7\penalty0 (4):\penalty0
  11142--11149, 2022.

\end{thebibliography}

\clearpage
\appendix
% Supplementary material: proofs, protocols, and additional experiments (submitted in the same PDF after the references).
% ══════════════════════════════════════════════════════════════
%  Notation
% ══════════════════════════════════════════════════════════════
\section{Notation}\label{app:notation}

\begin{table}[!ht]
\caption{\textbf{Summary of notation.} Symbols are grouped by the section in which they are introduced. Symbols used only
locally within a single proof are defined where they appear.}
\label{tab:notation}
\centering\small\setlength{\tabcolsep}{4pt}
\renewcommand{\arraystretch}{0.98}
\begin{tabular}{@{}p{0.29\linewidth}p{0.67\linewidth}@{}}
\toprule
Symbol & Meaning \\
\midrule
\multicolumn{2}{@{}l}{\emph{System, Koopman model, and controller (\Cref{sec:setup})}}\\
$\vec{x},\vec{u},f$; $n_x,n_u$ & physical state, control input, true dynamics $\vec{x}_{t+1}=f(\vec{x}_t,\vec{u}_t)$; state and input dimensions \\
$(\vec{x}^\ast,\vec{u}^\ast)$ & target equilibrium, shifted to the origin $(\vec{0},\vec{0})$ in the main text \\
$\mat{A}_f,\mat{B}_f$ & Jacobians of the true dynamics $f$ at the target equilibrium \\
$\mathcal{K}$ & infinite-dimensional Koopman operator (\Cref{sec:setup}) \\
$\Psi_\theta$, $\Phi_\theta(\vec{x})=[\vec{x};\,\Psi_\theta(\vec{x})-\Psi_\theta(\vec{0})]$, $N$ & neural network encoder, embedding function, latent dimension \\
$\vec{z}$, $\hat{\vec{z}}$; $\mat{C}=[\mat{I}_{n_x}\ \mat{0}]$ & encoded latent state, linear prediction; state projection $\vec{x}=\mat{C}\vec{z}$ \\
$\mat{A},\mat{B}$ & learned Koopman matrices \\
$\Jtheta,\Jtheta^{+}$ & Jacobian of $\Phi_\theta$ at the target ($\mat{C}\Jtheta=\mat{I}$) and its Moore--Penrose pseudoinverse $(\Jtheta\trans\Jtheta)^{-1}\Jtheta\trans$ \\
$\mat{Q}_x,\mat{R}$; $\mat{P}$; $\mat{K}$; $N_h$ & LQR cost weights; Riccati terminal matrix; Koopman LQR gain (\Cref{sec:setup}); MPC horizon \\
$\Acl=\mat{A}-\mat{B}\mat{K}$ & latent closed-loop matrix; the nominal stability certificate is $\rho(\Acl)<1$ \\
$g$; $\mat{F}=\mat{A}_f-\mat{B}_f\mat{K}\Jtheta$ & deployed nonlinear closed loop and its Jacobian at the target~\eqref{eq:deployed_loop} \\
\multicolumn{2}{@{}l}{\emph{Tangent invariance and certificate transfer (\Cref{sec:rhogap}, \Cref{app:further})}}\\
$E(\vec{x},\vec{u})$; $\mathcal{E}$ & one-step latent prediction error (\Cref{sec:rhogap}); closed-loop error Jacobian~\eqref{eq:ecal} \\
$d_1(\mat{M})$; $r_{\mathbb{R}}(\mat{M})$ & complex stability distance to the unit circle; real stability radius~\eqref{eq:real_radius} \\
$\mat{\Pi}=\Jtheta\mat{C}$; $\mat{F}_0$, $\mat{\Delta}$ & re-encoding projection~\eqref{eq:projector}; model part and physical linearization error of $\mat{F}$ (\Cref{prop:decomp}) \\
$\mat{M}$ & reference matrix (\Cref{thm:deployed_cert}) \\
\multicolumn{2}{@{}l}{\emph{TIK and data-driven certification (\Cref{sec:method})}}\\
$\vec{p}_i=[\vec{x}_i;\,\vec{u}_i]$; $w_i$; $\vec{c},\mat{D}$ & state-input pair of offline transition $i$; regression weight; offset and matrix of local fit~\eqref{eq:localfit} \\
$\hat{\mat{A}}_f,\hat{\mat{B}}_f$ & offline least-squares estimate of $(\mat{A}_f,\mat{B}_f)$ from $\hat{\mat{D}}=[\hat{\mat{A}}_f\ \ \hat{\mat{B}}_f]$ \\
$\mat{W}$; $\mat{N}_j$ & unconstrained trainable matrix on the orthogonal complement~\eqref{eq:ibc}; bilinear input matrices (\Cref{prop:ibc}(c)) \\
$\hat{\mat{K}}$; $\mat{K}_o$; $\hat{\mat{F}}$ & LQR gain of $(\hat{\mat{A}}_f,\hat{\mat{B}}_f,\mat{Q}_x,\mat{R})$; negative of the deployed policy Jacobian at the target; $\hat{\mat{A}}_f-\hat{\mat{B}}_f\mat{K}_o$ \\
$L$, $\epsilon$, $\mathcal{N}$ & Hessian curvature constant, measurement noise bound, neighborhood in \Cref{assum:curv} \\
$\varrho_i$; $\mat{\Sigma}$; $\vec{g}_{i,D}$; $\mat{\Gamma}$; $\eta(\mat{\Gamma})$ & residual bound of transition $i$; weighted Gram matrix; linear estimator vector; right multiplier; error bound~\eqref{eq:fitbound} \\
$g_{\mathrm{loc}}$; $L_H,L_T$ & deployed loop under local controller $\vec{u}=-\hat{\mat{K}}\vec{x}$; second- and third-order curvature constants (\Cref{thm:region_main}) \\
$\mat{S}$, $\mat{Q}$, $\hat\vartheta$, $\kappa$; $r$, $\Omega_c$ & Lyapunov matrix of $\hat{\mat{F}}$ and weight, contraction factor, condition number; certified radius and terminal set \\
\bottomrule
\end{tabular}
\end{table}

\section{A Scalar Example of a Nominally Certified but Unstable Controller}\label{app:example}

Consider the scalar system $x_{t+1}=f(x_t,u_t)=1.1x_t+0.2x_t^2+u_t$, which is open-loop unstable at $(x^\ast,u^\ast)=(0,0)$
with $A_f=1.1$ and $B_f=1$. We use the quadratic embedding $\vec{z}_t=\Phi(x_t)=[x_t,\,x_t^2]\trans$ with Jacobian
$\Jtheta=\partial\Phi/\partial x\big|_{0}=[1,\,0]\trans$. The true next encoded state is
$\vec{z}_{t+1}=\Phi(f(x_t,u_t))=[1.1x_t+0.2x_t^2+u_t;\,(1.1x_t+0.2x_t^2+u_t)^2]$: because the second coordinate
$z_{t+1,2}=(1.1x_t+0.2x_t^2+u_t)^2$ contains only terms of degree two or higher in $(x_t,u_t)$, its first derivatives at the origin are zero.
The full encoded update therefore has derivatives $\Jtheta A_f=[1.1,\,0]\trans$ and $\Jtheta B_f=[1,\,0]\trans$.

Consider the following approximate linear model $\hat{\vec{z}}_{t+1}=\mat{A}\vec{z}_t+\mat{B}u_t$:
\begin{align*}
  \mat{A}=\begin{bmatrix}1.1&0.2\\ -0.1&0.9\end{bmatrix},\qquad
  \mat{B}=\begin{bmatrix}1\\ -0.75\end{bmatrix},\qquad
  \hat{\vec{z}}_{t+1}=\begin{bmatrix}1.1x_t+0.2x_t^2+u_t\\ -0.1x_t+0.9x_t^2-0.75u_t\end{bmatrix}.
\end{align*}
Its first row $\hat{x}_{t+1}=\mat{C}\hat{\vec{z}}_{t+1}=1.1x_t+0.2x_t^2+u_t$ reproduces $f(x_t,u_t)$ with \emph{zero error
everywhere}. The second row approximates $x_{t+1}^2$ using the linear terms $-0.1x_t$
and $-0.75u_t$. These terms violate tangent invariance~\eqref{eq:ti} at the origin because
$\Jtheta A_f-\mat{A}\Jtheta=[0,\,0.1]\trans$ and $\Jtheta B_f-\mat{B}=[0,\,0.75]\trans$.

LQR with $Q_x=R=1$ on $(\mat{A},\mat{B})$
gives $\mat{K}\approx[0.02247,\,-0.86339]$. In the \emph{latent loop}, $\Acl=\mat{A}-\mat{B}\mat{K}$ has eigenvalues $0.95$ and $0.38$,
so the controller is nominally certified stable with $\rho(\Acl)=0.95<1$: because $\mat{B}=[1,\,-0.75]\trans$, LQR stabilizes
$\Acl$ via the gain $-0.863$ on $z_2$ while placing only $0.022$ on $z_1=x_t$.

In the \emph{deployed loop} on the true system,
re-encoding $\Phi(x_t)=[x_t,\,x_t^2]\trans$ gives $x_{t+1}=(1.1-K_1)x_t+(0.2-K_2)x_t^2$.
The quadratic term has zero slope at the origin, leaving $F=1.1-K_1\approx1.078>1$.
From $x_0=0.01$, the latent prediction decays to $0.00274$ at step $30$, while the plant state grows to $1.29$.

Enforcing tangent invariance sets $A_{21}=0$ and $B_2=0$, giving
\[
A_{\mathrm{TIK}}=\begin{bmatrix}1.1&0.2\\0&0.9\end{bmatrix},
\qquad
B_{\mathrm{TIK}}=\begin{bmatrix}1\\0\end{bmatrix}.
\]
% Enforcing tangent invariance sets the $(2,1)$
% entries of $\mat{A}$ and $\mat{B}$ to zero, $\mat{A}_{\mathrm{TIK}}=\left[\begin{smallmatrix}1.1&0.2\\0&0.9\end{smallmatrix}\right]$
% and $\mat{B}_{\mathrm{TIK}}=\left[\begin{smallmatrix}1\\0\end{smallmatrix}\right]$. 
LQR then gives
$\mat{K}_{\mathrm{TIK}}\approx[0.70343,\,0.19888]$. The deployed Jacobian is $F\approx0.397$, so the target is
locally exponentially stable.

% ══════════════════════════════════════════════════════════════
%  B  Proofs, remarks and further related work
% ══════════════════════════════════════════════════════════════
\section{Proofs of the Main Results}\label{app:mainproofs}

Where needed, this appendix keeps the target equilibrium $(\vec{x}^\ast,\vec{u}^\ast)$ explicit, whereas the main text shifts coordinates
so that $(\vec{x}^\ast,\vec{u}^\ast)=(\vec{0},\vec{0})$. We prove the results in \Cref{sec:rhogap} and
\Cref{sec:method}. The proofs of \Cref{prop:multipoint} and \Cref{cor:noexact} appear in \Cref{app:multipoint}. For any
real square matrix $\mat{M}$ with no eigenvalue on the unit circle, the real stability radius~\citep{qiu1995formula} is given by
\begin{align}\label{eq:real_radius}
  r_{\mathbb{R}}(\mat{M}) \;=\; \Bigl[ \sup_{|s|=1} \mu_{\mathbb{R}}\bigl((s\mat{I}-\mat{M})^{-1}\bigr) \Bigr]^{-1},
  \quad
  \mu_{\mathbb{R}}(\mat{X}) \;=\; \inf_{\gamma\in(0,1]} \sigma_2\!\left(\begin{bmatrix}
    \mathrm{Re}\,\mat{X} & -\gamma\,\mathrm{Im}\,\mat{X}\\
    \gamma^{-1}\mathrm{Im}\,\mat{X} & \mathrm{Re}\,\mat{X}
  \end{bmatrix}\right),
\end{align}
where $\sigma_2$ denotes the second-largest singular value. A conservative numerical test uses a lower bound on the
complex distance $d_1(\mat{M})$, which is no larger than $r_{\mathbb{R}}(\mat{M})$. For $m$ equally spaced points $s_j$ on the
unit circle, the singular-value perturbation inequality gives
\[
  d_1(\mat{M})\ge \min_j\sigma_{\min}(s_j\mat{I}-\mat{M})-2\sin\!\bigl(\pi/(2m)\bigr).
\]
For a verified lower bound, subtract any numerical error bound on the sampled singular values as well. The uncorrected
sampled minimum is an upper bound on $d_1$.

\subsection{Further Results on the Deployed Loop}\label{app:further}

We first state two structural results referenced in \Cref{sec:rhogap} and \Cref{sec:method}. Let
$\vec{r}(\vec{x},\vec{u})=\mat{C}E(\vec{x},\vec{u})$ denote the physical state block of the one-step prediction
error (\Cref{sec:rhogap}), and let $\mat{D}_x=\partial\vec{r}/\partial\vec{x}$ and $\mat{D}_u=\partial\vec{r}/\partial\vec{u}$
at the target equilibrium. Define the matrix
\begin{align}\label{eq:projector}
  \mat{\Pi}=\Jtheta\mat{C}=\begin{bmatrix}\mat{I}&\mat{0}\\ \partial\Psi_\theta/\partial\vec{x}&\mat{0}\end{bmatrix}.
\end{align}
Geometrically, $\mat{\Pi}$ represents state re-encoding to first order: it retains the physical state coordinates of a latent
vector and replaces the encoder coordinates with the linearized re-encoding of that physical
state~\citep{vangoor2023reprojection}. Because $\mat{C}\Jtheta=\mat{I}$, $\mat{\Pi}$ is a projection ($\mat{\Pi}^2=\mat{\Pi}$)
whose range is the tangent space $\mathrm{range}(\Jtheta)$.

\begin{proposition}[Decomposition of the deployed Jacobian]\label{prop:decomp}
Suppose $\Phi_\theta$ is differentiable at the target equilibrium. Then:
\begin{enumerate}
  \item[(a)] The deployed Jacobian decomposes as $\mat{F}=\mat{F}_0+\mat{\Delta}$, where $\mat{F}_0=\mat{C}\Acl\Jtheta$ is the
  model-predicted transition and $\mat{\Delta}=\mat{D}_x-\mat{D}_u\mat{K}\Jtheta$ is the physical linearization error.
  \item[(b)] The matrices $\mat{F}_0$ and $\mat{\Pi}\Acl$ share the same nonzero eigenvalues with identical multiplicities; in
  particular, $\rho(\mat{F}_0)=\rho(\mat{\Pi}\Acl)$.
  \item[(c)] If $(\mat{I}-\mat{\Pi})\Acl\Jtheta=\mat{0}$, meaning that $\Acl$ maps the tangent space $\mathrm{range}(\Jtheta)$
  into itself, then the characteristic polynomial of $\mat{F}_0$ divides that of $\Acl$, so $\rho(\mat{F}_0)\le\rho(\Acl)$.
\end{enumerate}
\end{proposition}

\Cref{prop:decomp} shows that the discrepancy between $\rho(\Acl)$ and $\rho(\mat{F})$ arises from two distinct sources: the
physical linearization error $\mat{\Delta}$, and the tangent-space invariance defect $(\mat{I}-\mat{\Pi})\Acl\Jtheta$ when the
latent loop leaves the tangent space. Even a model with zero physical linearization error $\mat{\Delta}=\mat{0}$ can fail on
the true system if $(\mat{I}-\mat{\Pi})\Acl\Jtheta\ne\mat{0}$, which can make $\rho(\mat{\Pi}\Acl)>1$ while $\rho(\Acl)<1$.

\begin{theorem}[Stability from a reference matrix]\label{thm:deployed_cert}
Assume a differentiable lifting and inactive input limits near the target. Let $\mat{M}$ be a real $n_x\times n_x$
matrix with no eigenvalue on the unit circle. If
$\|\mat{F}-\mat{M}\|_2<r_{\mathbb{R}}(\mat{M})$, then $\mat{F}$ and $\mat{M}$ have the same number of eigenvalues outside
the closed unit disc and none on the unit circle. In particular:
\begin{enumerate}
  \item[(a)] Taking $\mat{M}=\mat{F}_0$, if $\|\mat{\Delta}\|_2<r_{\mathbb{R}}(\mat{F}_0)$, local exponential stability of the
  deployed loop is decided by $\rho(\mat{F}_0)=\rho(\mat{\Pi}\Acl)<1$.
  \item[(b)] Taking $\mat{M}=\hat{\mat{F}}=\hat{\mat{A}}_f-\hat{\mat{B}}_f\mat{K}_o$ constructed from estimated Jacobians for any
  feedback policy preserving the target equilibrium and differentiable there with Jacobian $-\mat{K}_o$, where $\mat{K}_o=\mat{K}\Jtheta$ for lifted LQR and
  $\mat{K}_o=\hat{\mat{K}}$ for TIK, the condition holds whenever
  $\|(\hat{\mat{D}}-[\mat{A}_f\ \ \mat{B}_f])[\mat{I};-\mat{K}_o]\|_2<r_{\mathbb{R}}(\hat{\mat{F}})$.
\end{enumerate}
\end{theorem}

\subsection{Proofs of Transfer, Parameterization, and Stability}

\begin{proof}[Proof of \Cref{prop:decomp}]
(a)~Differentiating $f(\vec{x},\vec{u})=\mat{C}(\mat{A}\Phi_\theta(\vec{x})+\mat{B}\vec{u})+\vec{r}(\vec{x},\vec{u})$ at the
target yields $\mat{A}_f=\mat{C}\mat{A}\Jtheta+\mat{D}_x$ and $\mat{B}_f=\mat{C}\mat{B}+\mat{D}_u$. Substituting these into
$\mat{F}=\mat{A}_f-\mat{B}_f\mat{K}\Jtheta$ gives
$\mat{F}=\mat{C}(\mat{A}-\mat{B}\mat{K})\Jtheta+(\mat{D}_x-\mat{D}_u\mat{K}\Jtheta)=\mat{F}_0+\mat{\Delta}$.
(b)~Because $\mat{C}\Jtheta=\mat{I}$, we have $\mat{\Pi}^2=\Jtheta(\mat{C}\Jtheta)\mat{C}=\mat{\Pi}$ and
$\mathrm{range}(\mat{\Pi})=\mathrm{range}(\Jtheta)$. For any matrices $\mat{X}\in\R^{N\times n_x}$ and
$\mat{Y}\in\R^{n_x\times N}$, the products $\mat{X}\mat{Y}$ and $\mat{Y}\mat{X}$ share the same nonzero eigenvalues with
identical algebraic multiplicities; setting $\mat{X}=\Jtheta$ and $\mat{Y}=\mat{C}\Acl$ gives $\mat{Y}\mat{X}=\mat{F}_0$ and
$\mat{X}\mat{Y}=\mat{\Pi}\Acl$.
(c)~The condition $(\mat{I}-\mat{\Pi})\Acl\Jtheta=\mat{0}$ implies $\Acl\Jtheta=\mat{\Pi}\Acl\Jtheta=\Jtheta\mat{F}_0$. Because
$\Jtheta$ has full column rank, $\mathrm{range}(\Jtheta)$ is an $\Acl$-invariant subspace and $\mat{F}_0$ is the matrix
representation of $\Acl$ restricted to $\mathrm{range}(\Jtheta)$ in the basis formed by the columns of $\Jtheta$. Extending
the columns of $\Jtheta$ to a full basis of $\R^N$ puts $\Acl$ into block upper-triangular form with $\mat{F}_0$ as its
top-left diagonal block; hence the characteristic polynomial of $\mat{F}_0$ divides that of $\Acl$, and every eigenvalue of
$\mat{F}_0$ is an eigenvalue of $\Acl$.
\end{proof}

\begin{proof}[Proof of \Cref{thm:transfer}]
For $\mat{F}\vec{v}=\lambda\vec{v}$ with $\|\vec{v}\|_2=1$, \eqref{eq:ecal} gives
$(\lambda\mat{I}-\Acl)\Jtheta\vec{v}=\mathcal{E}\vec{v}$. Since $\|\Jtheta\vec{v}\|_2\ge\|\mat{C}\Jtheta\vec{v}\|_2=1$,
$\sigma_{\min}(\lambda\mat{I}-\Acl)\le\|\mathcal{E}\vec{v}\|_2/\|\Jtheta\vec{v}\|_2\le\|\mathcal{E}\|_2$.
If $\mathcal{E}=\mat{0}$, $\Jtheta\vec{v}$ is a nonzero eigenvector of $\Acl$ with eigenvalue $\lambda$.
For part~(b), when $\Acl$ is Schur, the resolvent norm $s\mapsto\|(s\mat{I}-\Acl)^{-1}\|_2$ is subharmonic on the exterior of the open unit
disc $\{s\in\mathbb{C}:|s|\ge1\}$ and vanishes as $|s|\to\infty$. By the maximum modulus principle for subharmonic functions,
it attains its maximum on the boundary $|s|=1$. Taking reciprocals gives
$\min_{|s|\ge1}\sigma_{\min}(s\mat{I}-\Acl)=\min_{|s|=1}\sigma_{\min}(s\mat{I}-\Acl)=d_1(\Acl)$. Thus, if
$\|\mathcal{E}\|_2<d_1(\Acl)$, no eigenvalue of $\mat{F}$ can satisfy $|\lambda|\ge1$, so $\rho(\mat{F})<1$ and Lyapunov's
indirect method~\citep{khalil2002nonlinear} ensures local exponential stability of the deployed loop.
\end{proof}

\begin{proof}[Proof of \Cref{cor:stablelift}]
If $\mat{A}_f\vec{v}=\lambda\vec{v}$ with $\vec{v}\ne\vec{0}$, then $\mat{A}\Jtheta\vec{v}=\lambda\Jtheta\vec{v}$ by
\eqref{eq:ti}. Full column rank gives $\Jtheta\vec{v}\ne\vec{0}$, so $\rho(\mat{A})\ge\rho(\mat{A}_f)>1$.
\end{proof}

\begin{proof}[Proof of \Cref{thm:deployed_cert}]
Let $\mat{Z}=\mat{F}-\mat{M}\in\R^{n_x\times n_x}$. Consider the continuous matrix path $\mat{M}(\tau)=\mat{M}+\tau\mat{Z}$ for
$\tau\in[0,1]$. If $\mat{M}(\tau)$ had an eigenvalue $s$ on the unit circle ($|s|=1$) for some $\tau\in(0,1]$, then
$s\mat{I}-\mat{M}-\tau\mat{Z}=(s\mat{I}-\mat{M})(\mat{I}-\tau(s\mat{I}-\mat{M})^{-1}\mat{Z})$ would be singular. By the
definition of the real structured singular value $\mu_{\mathbb{R}}$~\citep{qiu1995formula}, this requires
$\tau\|\mat{Z}\|_2\ge 1/\mu_{\mathbb{R}}((s\mat{I}-\mat{M})^{-1})\ge r_{\mathbb{R}}(\mat{M})$, which contradicts
$\|\mat{Z}\|_2<r_{\mathbb{R}}(\mat{M})$. Because the eigenvalues of $\mat{M}(\tau)$ vary continuously with $\tau$ and never
cross the unit circle, $\mat{F}=\mat{M}(1)$ and $\mat{M}=\mat{M}(0)$ have the same number of eigenvalues outside the unit disc.
Part~(a) follows with $\mat{M}=\mat{F}_0$ and \Cref{prop:decomp}(b); part~(b) follows from
$\mat{F}-\hat{\mat{F}}=([\mat{A}_f\ \ \mat{B}_f]-\hat{\mat{D}})[\mat{I};-\mat{K}_o]$.
\end{proof}

\begin{proof}[Proof of \Cref{prop:ibc}]
We prove the result in the general coordinate form $\mat{G}=\Jtheta\mat{T}$, where $\mat{T}=\mat{I}$ in standard coordinates
and $\mat{T}$ is the chart Jacobian when an angle $\theta$ is represented in the state by $(\sin\theta,\cos\theta)$ (in which
case $\hat{\mat{A}}_f,\hat{\mat{B}}_f$ are expressed in minimal angle coordinates and $\mat{Q}_x$ is replaced by
$\mat{T}\trans\mat{Q}_x\mat{T}$). In this coordinate form, replace $\Jtheta$ by $\mat{G}$ in \eqref{eq:ibc}.

(a)~Because $\mat{G}$ has full column rank, $\mat{G}^{+}\mat{G}=\mat{I}$ and $(\mat{I}-\mat{G}\mat{G}^{+})\mat{G}=\mat{0}$.
Multiplying~\eqref{eq:ibc} on the right by $\mat{G}$ yields $\mat{A}\mat{G}=\mat{G}\hat{\mat{A}}_f$, while
$\mat{B}=\mat{G}\hat{\mat{B}}_f$ holds by definition. Substituting these into~\eqref{eq:ecal} yields
$\mathcal{E}=\mat{G}\mat{F}-(\mat{A}-\mat{B}\mat{K})\mat{G}=\mat{G}(\mat{A}_f-\hat{\mat{A}}_f)-\mat{G}(\mat{B}_f-\hat{\mat{B}}_f)\mat{K}\mat{G}$.
\par (b)~In a basis of $\R^N$ adapted to $\mathrm{range}(\mat{G})\oplus\mathrm{range}(\mat{G})^\perp$, $\mat{A}$ is block
upper-triangular with top-left block $\hat{\mat{A}}_f$ and bottom-right block given by the restriction of
$(\mat{I}-\mat{G}\mat{G}^{+})\mat{W}(\mat{I}-\mat{G}\mat{G}^{+})$ to $\mathrm{range}(\mat{G})^\perp$, while
$\mathrm{range}(\mat{B})\subseteq\mathrm{range}(\mat{G})$ lies entirely in the first block. Because the orthogonal complement
block is Schur by assumption and $(\hat{\mat{A}}_f,\hat{\mat{B}}_f)$ is stabilizable, the full latent pair $(\mat{A},\mat{B})$
is stabilizable. Similarly, any eigenvector of $\mat{A}$ with $|\lambda|\ge1$ must lie in $\mathrm{range}(\mat{G})$, where
$\mat{C}\mat{G}=\mat{T}$ has full column rank, so detectability of $(\hat{\mat{A}}_f,(\mat{T}\trans\mat{Q}_x\mat{T})^{1/2})$
ensures detectability of $(\mat{A},\mat{Q}_x^{1/2}\mat{C})$. Therefore, the latent LQR Riccati equation has a unique
stabilizing solution $\mat{P}\succeq0$. For an initial state on the tangent space, $\vec{z}_0=\mat{G}\vec{x}_0$, any
control sequence $\vec{u}_{0:\infty}$ keeps the latent trajectory inside $\mathrm{range}(\mat{G})$ via
$\vec{z}_t=\mat{G}\vec{x}_t$ with $\vec{x}_{t+1}=\hat{\mat{A}}_f\vec{x}_t+\hat{\mat{B}}_f\vec{u}_t$, and the stage costs match
identically: $\vec{z}_t\trans\mat{C}\trans\mat{Q}_x\mat{C}\vec{z}_t=\vec{x}_t\trans\mat{T}\trans\mat{Q}_x\mat{T}\vec{x}_t$.
Because $\mat{R}\succ0$ makes the optimal control sequence unique, the optimal first control action of the latent LQR problem
from $\vec{z}_0=\mat{G}\vec{x}_0$ must equal that of the physical LQR problem on
$(\hat{\mat{A}}_f,\hat{\mat{B}}_f,\mat{T}\trans\mat{Q}_x\mat{T},\mat{R})$: $-\mat{K}\mat{G}\vec{x}_0=-\hat{\mat{K}}\vec{x}_0$
for all $\vec{x}_0$, proving $\mat{K}\mat{G}=\hat{\mat{K}}$.
\par (c)~With the target equilibrium written explicitly, the bilinear term is
$\sum_{j=1}^{n_u}(u_{t,j}-u^\ast_j)\mat{N}_j(\vec{z}_t-\Phi_\theta(\vec{x}^\ast))$. Both this term and its first partial
derivatives with respect to $\vec{z}_t$ and $\vec{u}_t$ vanish at $(\vec{z}_t,\vec{u}_t)=(\Phi_\theta(\vec{x}^\ast),\vec{u}^\ast)$,
leaving the linearization $(\mat{A},\mat{B})$ and properties (a)--(b) unchanged.
\end{proof}

\begin{proof}[Proof of \Cref{prop:kmpc}]
Since $\mat{R}\succ0$, substituting $\vec{z}_k=\mat{A}^k\vec{z}_0+\sum_{j=0}^{k-1}\mat{A}^{k-1-j}\mat{B}\vec{u}_j$
turns~\eqref{eq:kmpc_qp} into a strictly convex quadratic program in the input sequence $\vec{u}_{0:N_h-1}$. Since the target
equilibrium lies in the strict interior of the state and input constraints, the unconstrained minimizer satisfies all constraints in
an open neighborhood of the target. With the stabilizing Riccati solution $\mat{P}$ as the terminal weight, the terminal cost
$\vec{z}_{N_h}\trans\mat{P}\vec{z}_{N_h}$ equals the optimal infinite-horizon LQR cost-to-go from $\vec{z}_{N_h}$, so by
Bellman's principle of optimality the unconstrained finite-horizon minimizer applies $\vec{u}_0=-\mat{K}\vec{z}_0$ at the first
step for every horizon $N_h\ge1$. Combining this with \Cref{prop:ibc}(b) completes the proof.
\end{proof}

\begin{proof}[Proof of \Cref{thm:knownfp}]
In the centered coordinates of the main text, write $\mat{H}_f=[\mat{D}_f\ \vec{0}]$ and
$\hat{\mat{H}}=[\hat{\mat{D}}\ \hat{\vec{c}}]$. Since
$\vec{x}_{i+1}=\mat{H}_f\vec{q}_i+\vec{\ell}_i$, the weighted normal equations give
\[
 (\hat{\mat{H}}-\mat{H}_f)\mat{\Sigma}
 =\sum_i w_i\vec{\ell}_i\vec{q}_i\trans,\qquad
 (\hat{\mat{D}}-\mat{D}_f)\mat{\Gamma}
 =\sum_i w_i\vec{\ell}_i(\mat{\Gamma}\trans\vec{g}_{i,D})\trans.
\]
The full-rank condition makes it possible to invert $\mat{\Sigma}$. Bounding each rank-one summand by
$w_i\varrho_i\|\mat{\Gamma}\trans\vec{g}_{i,D}\|_2$ proves \eqref{eq:fitbound}, including the fitted intercept.
This deterministic inequality holds simultaneously for all dimensionally compatible $\mat{\Gamma}$, so a gain chosen using the same
data is allowed when the curvature and noise bounds hold.
The deployed map $g(\vec{x})=f(\vec{x},\vec{\pi}(\vec{x}))$ is differentiable at the equilibrium with Jacobian
$\mat{F}=\mat{A}_f-\mat{B}_f\mat{K}_o=[\mat{A}_f\ \ \mat{B}_f]\mat{\Gamma}$, while $\hat{\mat{F}}=\hat{\mat{D}}\mat{\Gamma}$
with $\mat{\Gamma}=[\mat{I};-\mat{K}_o]$. By~\eqref{eq:fitbound}, $\|\mat{F}-\hat{\mat{F}}\|_2\le\eta<r_{\mathbb{R}}(\hat{\mat{F}})$.
Applying \Cref{thm:deployed_cert}(b) and Lyapunov's indirect method~\citep{khalil2002nonlinear} establishes local exponential
stability when $\rho(\hat{\mat{F}})<1$ and instability when $\rho(\hat{\mat{F}})>1$.
\end{proof}

The following two corollaries extend \Cref{thm:knownfp} to stochastic measurement noise (including noisy regression states)
and to third-order Taylor remainders used in our two-scale offline verification protocol (\Cref{sec:exp_datacert}).

\begin{corollary}[Offline certificate with stochastic and regressor noise]\label{cor:knownfp_noise}
In \Cref{thm:knownfp}, condition on the regressors, weights, and $\mat{\Gamma}$. Suppose the additional successor errors
are conditionally independent across transitions, conditionally zero mean, and bounded by $|e_{ij}|\le\epsilon'$. This conditioning must preserve the noise assumptions; a gain fitted to the same noisy successors does not automatically satisfy them. Then with
probability at least $1-\delta$, the conclusion of \Cref{thm:knownfp} holds with $\eta$ replaced by
\begin{align}\label{eq:probbound}
  \eta_\delta \;=\; \eta \;+\; \epsilon'\sqrt{2n_x\log(2n_xm/\delta)}\;\Bigl(\sum_i w_i^2\|\mat{\Gamma}\trans\vec{g}_{i,D}\|_2^2\Bigr)^{1/2},
\end{align}
Here $m$ is the number of columns of $\mat{\Gamma}$. For bounded regressor errors
$\tilde{\vec{p}}_i=\vec{p}_i+\vec{v}_i$, $\|\vec{v}_i\|_2\le b$, use the full-rank measured design to compute the
Gram matrix and estimator vectors. In $\varrho_i$, replace $\epsilon$ by $\epsilon+D_{\max}b$ for a valid bound
$D_{\max}\ge\|Df(\vec{p}^\ast)\|_2$, and upper-bound the distance by $\|\tilde{\vec{p}}_i-\vec{p}^\ast\|_2+b$.
The deterministic regressor correction also applies without the stochastic term.
\end{corollary}

\begin{proof}
The zero-mean noise contribution to $(\hat{\mat{D}}-\mat{D}_f)\mat{\Gamma}$ is
$\mat{\Xi}=\sum_iw_i\vec{e}_i(\mat{\Gamma}\trans\vec{g}_{i,D})\trans$. Each entry $\Xi_{jl}$ is a sum of independent zero-mean variables
bounded by $\epsilon'w_i|(\mat{\Gamma}\trans\vec{g}_{i,D})_l|$. Applying Hoeffding's inequality to each of the $n_x m$ entries
with failure probability $\delta/(n_x m)$ and bounding $\|\mat{\Xi}\|_2\le\|\mat{\Xi}\|_F$ yields~\eqref{eq:probbound}.
The regressor correction follows by expanding at the true target and writing the linear term as
$Df(\vec{p}^\ast)(\tilde{\vec{p}}_i-\vec{p}^\ast)-Df(\vec{p}^\ast)\vec{v}_i$.
For a gain fitted to the same noisy successors, the conditional independence in the corollary needs separate justification.
One can instead bound the estimation error with $\mat{\Gamma}=\mat{I}$ and multiply by $\|\mat{\Gamma}\|_2$ for the fitted gain.
\end{proof}

\begin{corollary}[Offline certificate with a cubic Taylor remainder]\label{cor:cubic}
Let $f$ be three times continuously differentiable ($C^3$) on the neighborhood $\mathcal{N}$ in \Cref{assum:curv}. In the induced Euclidean multilinear norms,
assume $\|D^2 f(\vec{p}^\ast)\|\le L_H$ and $\|D^3 f(\vec{p})\|\le L_T$ for all $\vec{p}\in\mathcal{N}$. Then the estimation error bound~\eqref{eq:fitbound} and \Cref{thm:knownfp} hold with
$\varrho_i = \frac{1}{2}L_H \|\vec{p}_i - \vec{p}^\ast\|_2^2 + \frac{1}{6}L_T \|\vec{p}_i - \vec{p}^\ast\|_2^3 + \epsilon$.
The same argument applies in an input-weighted norm when the derivative bounds and Taylor distances use that norm.
\end{corollary}

\begin{proof}
By Taylor's theorem with integral remainder around $\vec{p}^\ast$, for $\vec{d}=\vec{p}_i-\vec{p}^\ast$ we have
$f(\vec{p}_i)-f(\vec{p}^\ast)-Df(\vec{p}^\ast)\vec{d}=\tfrac12D^2f(\vec{p}^\ast)[\vec{d},\vec{d}]+\int_0^1\tfrac{(1-t)^2}{2}D^3f(\vec{p}^\ast+t\vec{d})[\vec{d},\vec{d},\vec{d}]\,dt$,
whose Euclidean norm is bounded by $\tfrac12L_H\|\vec{d}\|_2^2+\tfrac16L_T\|\vec{d}\|_2^3$. Adding the measurement
noise bound $\epsilon$ gives the stated residual bound $\varrho_i$.
\end{proof}

\subsection{Remarks on the Main Results}\label{app:theory_remarks}

\paragraph{Why first-order tangent invariance.}
Vanishing first derivatives of the one-step latent error $E$ are sufficient for exact transfer in \Cref{thm:transfer}. Without
first-order tangent invariance, a Schur $\Acl$ places no bound on $\rho(\mat{F}_0)$: partitioning
$\Acl=\left[\begin{smallmatrix}\mat{A}_{xx}&\mat{A}_{x\Psi}\\ \mat{A}_{\Psi x}&\mat{A}_{\Psi\Psi}\end{smallmatrix}\right]$
gives $\mat{F}_0=\mat{A}_{xx}+\mat{A}_{x\Psi}\mat{J}_{\Psi_\theta}$. Whenever $\mat{A}_{x\Psi}\ne\mat{0}$, choosing
$\mat{J}_{\Psi_\theta}=\omega\,\vec{v}_1\vec{u}_1\trans$ along the leading singular vectors of $\mat{A}_{x\Psi}$ makes
$\rho(\mat{F}_0)\ge|\tr\mat{F}_0|/n_x=|\tr\mat{A}_{xx}+\omega\sigma_1(\mat{A}_{x\Psi})|/n_x$ grow arbitrarily large with
$\omega$ while $\Acl$ remains fixed and Schur.

\paragraph{Matching for one gain or all gains.}
\Cref{thm:transfer}(a) requires the closed-loop error Jacobian $\mathcal{E}=\Jtheta\mat{F}-\Acl\Jtheta$ to vanish for the
specific feedback gain $\mat{K}$ in use. Tangent invariance (\Cref{def:ti}) is an open-loop property of the learned model
$(\mat{A},\mat{B})$ alone ($\Jtheta\mat{A}_f=\mat{A}\Jtheta$ and $\Jtheta\mat{B}_f=\mat{B}$), which by~\eqref{eq:ecal}
guarantees $\mathcal{E}=\mat{0}$ simultaneously for every feedback gain $\mat{K}$. Given offline estimates
$(\hat{\mat{A}}_f,\hat{\mat{B}}_f)$, one can also impose tangent invariance after training via the minimum-norm
projection
\begin{equation}\label{eq:anchored}
  \mat{A}' \;=\; \mat{A} + (\Jtheta\hat{\mat{A}}_f - \mat{A}\Jtheta)\Jtheta^{+}, \qquad \mat{B}' \;=\; \Jtheta\hat{\mat{B}}_f,
\end{equation}
which we term \emph{anchored invariance}. However, modifying $(\mat{A},\mat{B})$ after training changes the LQR gain $\mat{K}$; for
uncentered controllers this shifts the closed-loop equilibrium, requiring iterative re-anchoring or locating the equilibrium
from offline data via the Newton--Kantorovich theorem~\citep{ortega1970iterative}.
TIK in \Cref{prop:ibc} avoids this iteration by centering at the target equilibrium and enforcing tangent invariance by
construction throughout training.

\paragraph{Estimating curvature constants and selecting LQR weights.}
In our two-scale offline verification protocol of \Cref{sec:exp_datacert}, fitting a local cubic polynomial on a wider
neighborhood averages out observation noise; at upright pendulum equilibria, odd symmetry implies $D^2f(\vec{0},\vec{0})=\mat{0}$,
making a cubic fit necessary to capture the leading nonlinear remainder. The empirical constants $L_H$ and $L_T$ are computed
from the scaled norms of the fitted second- and third-order polynomial tensors plus their Hoeffding noise bounds, multiplied by
a safety factor of $2$. Because $\hat{\mat{K}}=\mat{K}\Jtheta$ is the LQR gain of $(\hat{\mat{A}}_f,\hat{\mat{B}}_f)$
for any choice of cost weights $(\mat{Q}_x,\mat{R})$ without retraining the TIK model, the weights can be chosen to enlarge the
real stability radius $r_{\mathbb{R}}(\hat{\mat{F}})$ relative to the estimation bound $\eta$, as detailed in \Cref{app:ibc_cert}.

% ══════════════════════════════════════════════════════════════
%  C  Regional Certificates and Dual-Mode Koopman MPC
% ══════════════════════════════════════════════════════════════
\section{Proof of the Certified Terminal Set}\label{app:rmpc}

We first prove \Cref{thm:region}, which measures distances and curvature in the Lyapunov metric defined by $\mat{S}\succ0$.
We then obtain the Euclidean bound in \Cref{thm:region_main}. The constrained swing-up experiments in
\Cref{sec:exp_bench} use the metric formulation.

Let the true nonlinear system be $\vec{x}_{t+1}=f(\vec{x}_t,\vec{u}_t)$ with target equilibrium $f(\vec{x}^\ast,\vec{u}^\ast)=\vec{x}^\ast$,
input box $\mathcal{U}=\{\vec{u}:\vec{u}_{\min}\le\vec{u}\le\vec{u}_{\max}\}$ containing $\vec{u}^\ast$ in its interior, and
state box $\mathcal{X}=\{\vec{x}:\vec{x}_{\mathrm{lo}}\le\vec{x}\le\vec{x}_{\mathrm{hi}}\}$ containing $\vec{x}^\ast$ in its
interior. Write $\bar u_i=\min(u_{\max,i}-u^\ast_i,\,u^\ast_i-u_{\min,i})>0$ and $\vec{e}=\vec{x}-\vec{x}^\ast$. Under the local
linear feedback law
\begin{align}\label{eq:catch}
  \vec{\pi}(\vec{x})=\vec{u}^\ast-\hat{\mat{K}}(\vec{x}-\vec{x}^\ast),
\end{align}
the input satisfies its limits on the set $\mathcal{S}=\{\vec{x}:|\hat{\mat{K}}_i(\vec{x}-\vec{x}^\ast)|\le\bar u_i\ \forall i\}$,
where the deployed closed-loop map is $g_{\mathrm{loc}}(\vec{x})=f(\vec{x},\vec{\pi}(\vec{x}))$ with Jacobian
$\mat{F}=Dg_{\mathrm{loc}}(\vec{x}^\ast)=\mat{A}_f-\mat{B}_f\hat{\mat{K}}$. Given $\mat{S}\succ0$, let $\mat{V}=\mat{S}^{1/2}$,
define the weighted vector norm $\|\vec{a}\|_{\mat{S}}=\|\mat{V}\vec{a}\|_2$, and denote the induced matrix norm by
$\|\mat{M}\|_{\mat{S}}=\|\mat{V}\mat{M}\mat{V}^{-1}\|_2$.
All terms in the metric test must use these transformed coordinates: the error bound controls
$\mat{V}(\hat{\mat{F}}-\mat{F})\mat{V}^{-1}$, and a derivative of order $k$ transforms as
$\mat{V}D^kg_{\mathrm{loc}}[\mat{V}^{-1}\cdot,\ldots,\mat{V}^{-1}\cdot]$.
Euclidean curvature estimates cannot be substituted without the corresponding norm factors.

\begin{assumption}[Closed-loop curvature in the Lyapunov metric]\label{assum:curvP}
Let $g_{\mathrm{loc}}$ be $C^3$ on a convex set $\mathcal{N}\subseteq\mathcal{S}$ containing the target and all regressors.
In the multilinear norms induced by $\|\cdot\|_{\mat{S}}$, assume
$\|D^2g_{\mathrm{loc}}(\vec{x}^\ast)\|_{\mat{S}}\le L_H^{\mat{S}}$ and
$\|D^3g_{\mathrm{loc}}(\vec{x})\|_{\mat{S}}\le L_T^{\mat{S}}$ for all $\vec{x}\in\mathcal{N}$. Offline closed-loop transitions satisfy $\vec{y}_i=g_{\mathrm{loc}}(\vec{x}_i)+\vec{\epsilon}_i$
with $\|\vec{\epsilon}_i\|_2\le\epsilon$.
\end{assumption}

\begin{lemma}[Metric Taylor remainder and least-squares fit bound]\label{lem:taylorP}\label{lem:fitP}
Under \Cref{assum:curvP}:
\begin{enumerate}
  \item[(a)] For every $\vec{x}\in\mathcal{N}$ with $\vec{e}=\vec{x}-\vec{x}^\ast$,
  $\|g_{\mathrm{loc}}(\vec{x})-\vec{x}^\ast-\mat{F}\vec{e}\|_{\mat{S}}\le\tfrac12L_H^{\mat{S}}\|\vec{e}\|_{\mat{S}}^2+\tfrac16L_T^{\mat{S}}\|\vec{e}\|_{\mat{S}}^3$.
  \item[(b)] If $\hat{\mat{F}}$ is fitted by least squares on pairs $(\vec{x}_i,\vec{y}_i)$ in $\mathcal{N}$ with regressors
  $\vec{\zeta}_i=[\vec{e}_i;\,1]$, Gram matrix $\mat{\Sigma}=\sum_i\vec{\zeta}_i\vec{\zeta}_i\trans\succ0$, and
  $\vec{g}_{i,D}$ denoting the first $n_x$ entries of $\mat{\Sigma}^{-1}\vec{\zeta}_i$, then
  \begin{align}\label{eq:etaP}
    \|\hat{\mat{F}}-\mat{F}\|_{\mat{S}} &\;\le\; \eta_{\mat{S}} \;:=\; \min\Bigl\{\sum_i\varrho_i\|\mat{V}^{-1}\vec{g}_{i,D}\|_2,\;
    \sqrt{n_x}\max_j\sum_i\varrho_i|(\mat{V}^{-1}\vec{g}_{i,D})_j|\Bigr\},
  \end{align}
  where $\varrho_i=\tfrac12L_H^{\mat{S}}\|\vec{e}_i\|_{\mat{S}}^2+\tfrac16L_T^{\mat{S}}\|\vec{e}_i\|_{\mat{S}}^3+\|\mat{V}\|_2\,\epsilon$.
\end{enumerate}
\end{lemma}

\begin{proof}
Part~(a) follows from Taylor's theorem with integral remainder along the segment from $\vec{x}^\ast$ to $\vec{x}$ in the norm
$\|\cdot\|_{\mat{S}}$. For part~(b), writing $\vec{y}_i=\vec{x}^\ast+\mat{F}\vec{e}_i+\vec{\ell}_i$ gives
$\mat{V}(\hat{\mat{F}}-\mat{F})\mat{V}^{-1}=\sum_i(\mat{V}\vec{\ell}_i)(\mat{V}^{-1}\vec{g}_{i,D})\trans$ with
$\|\mat{V}\vec{\ell}_i\|_2\le\varrho_i$ by part~(a). For any unit vector $\vec{b}\in\R^{n_x}$, the triangle inequality gives
$\bigl\|\sum_i(\mat{V}\vec{\ell}_i)(\mat{V}^{-1}\vec{g}_{i,D})\trans\vec{b}\bigr\|_2\le\sum_i\varrho_i|(\mat{V}^{-1}\vec{g}_{i,D})\trans\vec{b}|$;
applying Cauchy--Schwarz yields the first bound in~\eqref{eq:etaP}, and applying $\|\vec{b}\|_1\le\sqrt{n_x}$ yields the second.
The weighted Euclidean counterpart is derived in the proof of \Cref{thm:knownfp}.
\end{proof}

\begin{theorem}[Certified invariant terminal set in the Lyapunov metric]\label{thm:region}
Let $\mat{S}\succ0$, suppose \Cref{assum:curvP} holds, and let $\hat{\mat{F}}$ and $\eta_{\mat{S}}$ be given by
\Cref{lem:fitP}. Set $\hat\vartheta_{\mat{S}}=\|\hat{\mat{F}}\|_{\mat{S}}$ and assume
$\delta:=1-\hat\vartheta_{\mat{S}}-\eta_{\mat{S}}>0$. Let $s^\ast$ be the positive root of $\tfrac12L_H^{\mat{S}}s+\tfrac16L_T^{\mat{S}}s^2=\delta$, or $+\infty$
if both curvature bounds are zero. Define
\begin{align}\label{eq:cstar}
  c^\ast=\min\Bigl\{(s^\ast)^2,\ c_{\mathcal{N}},\ \min_i\frac{\bar u_i^2}{\hat{\mat{K}}_i\mat{S}^{-1}\hat{\mat{K}}_i\trans},\
  \min_{j,\pm}\frac{(b_j^{\pm})^2}{(\mat{S}^{-1})_{jj}}\Bigr\},
\end{align}
where $c_{\mathcal{N}}=\sup\{c:\Omega_c\subseteq\mathcal{N}\}$,
$\Omega_c=\{\vec{x}:(\vec{x}-\vec{x}^\ast)\trans\mat{S}(\vec{x}-\vec{x}^\ast)\le c\}$, $b_j^{+}=x_{\mathrm{hi},j}-x^\ast_j$, and
$b_j^-=x^\ast_j-x_{\mathrm{lo},j}$. A zero gain row gives an infinite input bound. For every $0<c\le c^\ast$
with $\Omega_c\subseteq\mathcal{N}$, the ellipsoid satisfies $\Omega_c\subseteq\mathcal{X}$,
the control input $\vec{\pi}(\vec{x})$ lies within $\mathcal{U}$ on $\Omega_c$, and $\Omega_c$ is forward invariant under the
deployed loop; moreover, for $c<(s^\ast)^2$, every trajectory starting in $\Omega_c$ satisfies
$\|\vec{x}_t-\vec{x}^\ast\|_{\mat{S}}\le\lambda(\sqrt{c})^t\|\vec{x}_0-\vec{x}^\ast\|_{\mat{S}}$ with
$\lambda(\sqrt{c})=\hat\vartheta_{\mat{S}}+\eta_{\mat{S}}+\tfrac12L_H^{\mat{S}}\sqrt{c}+\tfrac16L_T^{\mat{S}}c<1$.
\end{theorem}

\begin{proof}[Proof of \Cref{thm:region} and \Cref{thm:region_main}]
In \Cref{thm:region}, maximizing the linear forms $\hat{\mat{K}}_i\vec{e}$ and $\pm e_j$ over $\{\vec{e}\trans\mat{S}\vec{e}\le c\}$
gives $\sqrt{c\,\hat{\mat{K}}_i\mat{S}^{-1}\hat{\mat{K}}_i\trans}\le\bar u_i$ and $\sqrt{c(\mat{S}^{-1})_{jj}}\le b_j^\pm$ by
the definition of $c^\ast$ in~\eqref{eq:cstar}, ensuring input and state constraint satisfaction on $\Omega_c$. For any
$\vec{x}\in\Omega_c\subseteq\mathcal{N}$ with $s=\|\vec{e}\|_{\mat{S}}\le\sqrt{c}\le s^\ast$, the triangle inequality and
\Cref{lem:fitP} yield
\begin{align*}
  \|g_{\mathrm{loc}}(\vec{x})-\vec{x}^\ast\|_{\mat{S}}
  \;\le\; \|\hat{\mat{F}}\|_{\mat{S}}s + \|\mat{F}-\hat{\mat{F}}\|_{\mat{S}}s + \|g_{\mathrm{loc}}(\vec{x})-\vec{x}^\ast-\mat{F}\vec{e}\|_{\mat{S}}
  \;\le\; \lambda(s)\,s \;\le\; s,
\end{align*}
proving forward invariance $g_{\mathrm{loc}}(\Omega_c)\subseteq\Omega_c$ and geometric decay when $c<(s^\ast)^2$.
To obtain \Cref{thm:region_main} (stated in Euclidean norms), note that for $\|\vec{e}\|_2\le r$,
$\|\vec{w}\|_{\mat{S}}\le\sqrt{\lambda_{\max}(\mat{S})}\|\vec{w}\|_2$ and
$\|\vec{e}\|_2\le\|\vec{e}\|_{\mat{S}}/\sqrt{\lambda_{\min}(\mat{S})}$ imply
$\|g_{\mathrm{loc}}(\vec{x})-\vec{x}^\ast\|_{\mat{S}}\le\bigl[\hat\vartheta+\sqrt\kappa(\eta+\tfrac12L_Hr+\tfrac16L_Tr^2)\bigr]\|\vec{e}\|_{\mat{S}}$.
The bound in \Cref{thm:region_main} makes this bracket at most $1$ and ensures forward invariance on
$\Omega_c\subseteq\{\|\vec{e}\|_2\le r\}$. A strict bound gives exponential convergence.
\end{proof}

% ══════════════════════════════════════════════════════════════
%  D  Experimental setup
% ══════════════════════════════════════════════════════════════
\section{Experimental Setup}\label{app:setup}

A single export script generates the reported numerical results from evaluation logs. A separate pipeline reruns the experiments from trained checkpoints. Code, trained models, and simulated offline datasets will
be released upon publication.

\paragraph{Benchmark systems and offline datasets.}
\Cref{tab:systems} summarizes the simulated dynamical systems studied in this paper. For the pendulums and cart-poles, the
offline dataset consists of short trajectories starting from random states under random control inputs; for the
torque-controlled seven-joint arm, a joint PD controller tracks random setpoints with exploration noise. On all benchmarks,
the true system Jacobians $(\mat{A}_f,\mat{B}_f)$ for ground-truth evaluation are computed via central finite differences.

\begin{table}[t]
\caption{\textbf{Summary of benchmark dynamical systems.} $n_x,n_u$: state and control dimensions; $\Delta t$: discrete control
period; $|u|_{\max}$: actuator bound per input channel; Transitions: total one-step transitions in the offline training
dataset; Simulator: physics engine used for dynamics and ground-truth Jacobians.}
\label{tab:systems}
\centering\footnotesize\setlength{\tabcolsep}{3.0pt}
\resizebox{\linewidth}{!}{%
\begin{tabular}{@{}lrrlrrrl@{}}
\toprule
System & $n_x$ & $n_u$ & Physical state coordinates & $\Delta t$\,(s) & $|u|_{\max}$ & Transitions & Simulator \\
\midrule
\multicolumn{8}{@{}l}{\emph{Pendulums}} \\
Damped pendulum & 2 & 1 & $(\theta,\dot\theta)$ & $\pnSysDampingPendulumDt$ & $\pnSysDampingPendulumU$ & $\pnSysDampingPendulumTr$ & exact (ODE) \\
Hanging double pendulum & 4 & 2 & $(\theta_{1,2},\dot\theta_{1,2})$ & $\pnSysDoublePendulumDt$ & $\pnSysDoublePendulumU$ & $\pnSysDoublePendulumTr$ & exact (ODE) \\
Double pendulum (upright) & 4 & 2 & $(\theta_{1,2},\dot\theta_{1,2})$ & $\pnSysDoublePendulumUpDt$ & $\pnSysDoublePendulumUpU$ & $\pnSysDoublePendulumUpTr$ & exact (ODE) \\
\midrule
\multicolumn{8}{@{}l}{\emph{Cart-poles}} \\
Cart-pole (ODE) & 4 & 1 & $(x,\theta,\dot x,\dot\theta)$ & $\pnSysCartPoleDt$ & $\pnSysCartPoleU$ & $\pnSysCartPoleTr$ & exact (ODE) \\
Cart-pole (MuJoCo) & 4 & 1 & $(x,\theta,\dot x,\dot\theta)$ & $\pnSysMujocoCartPoleDt$ & $\pnSysMujocoCartPoleU$ & $\pnSysMujocoCartPoleTr$ & exact (MuJoCo) \\
Cart-pole swing-up (MuJoCo) & 5 & 1 & $(x,\sin\theta,\cos\theta,\dot x,\dot\theta)$ & $\pnSysMujocoCartPoleSwingDt$ & $\pnSysMujocoCartPoleSwingU$ & $\pnSysMujocoCartPoleSwingTr$ & exact (MuJoCo) \\
Double cart-pole (MuJoCo) & 6 & 1 & $(x,\theta_{1,2},\dot x,\dot\theta_{1,2})$ & $\pnSysMujocoDoubleCartPoleDt$ & $\pnSysMujocoDoubleCartPoleU$ & $\pnSysMujocoDoubleCartPoleTr$ & exact (MuJoCo) \\
\midrule
\multicolumn{8}{@{}l}{\emph{Robot manipulators and legged systems}} \\
Seven-joint arm (analytic) & 14 & 7 & joint positions, velocities & $\pnSysFrankaCleanDt$ & $\pnSysFrankaCleanU$ & $\pnSysFrankaCleanTr$ & exact (ODE) \\
Kinova Gen3 arm (SPARK) & 14 & 7 & joint positions, velocities & $0.02$ & $10$ & $50{,}000$ & exact (SPARK) \\
Unitree Go2 quadruped & 35 & 12 & joints, velocities, base pose & $0.01$ & $1.0$ & $\pnSysGoTwoTr$ & MuJoCo \\
\bottomrule
\end{tabular}%
}
\end{table}

\paragraph{Network architectures and training regimes.}
Unless specified otherwise, the neural encoder $\Psi_\theta$ is a multilayer perceptron with two hidden layers of width $256$
and ReLU activations. Standard Koopman and Koopman + invariance penalty each use $10$ random seeds per system. The
extended evaluation in \Cref{app:fixedpoint_more} includes additional regularization regimes with $5$ seeds each: Gaussian state
noise augmentation, a $1$-Lipschitz encoder parameterized with orthogonal layers~\citep{wang2023direct}, and a penalty on the
product of the latent feedback gain norm and encoder Lipschitz constant $\|\mat{K}_\Psi\|_2\Lpsi$~\citep{abuduweili2024lipschitz},
evaluated with and without a variance floor to prevent feature collapse.

\subsection{Training Objectives and Baseline Details}\label{app:training}

Unless stated otherwise, Koopman models are trained on $H=15$-step windows from the offline trajectories. Each rollout starts from
$\hat{\vec{z}}_0=\Phi_\theta(\vec{x}_0)$ and advances according to $\hat{\vec{z}}_{k+1}=\mat{A}\hat{\vec{z}}_k+\mat{B}\vec{u}_k$. The
discounted multi-step state prediction loss and the normalized latent linearity loss~\citep{lusch2018deep} are
\begin{align}\label{eq:loss}
  \mathcal{L}_{\mathrm{pred}}=\frac{1}{\sum_k\gamma^k}\sum_{k=0}^{H-1}\gamma^k
  \bigl\|\mat{C}\hat{\vec{z}}_{k+1}-\vec{x}_{k+1}\bigr\|_2^2 ,\quad
  \mathcal{L}_{\mathrm{lat}}=\frac{1}{\bar\sigma_\Psi^2\sum_k\gamma^k}\sum_{k=0}^{H-1}\gamma^k
  \bigl\|[\hat{\vec{z}}_{k+1}-\Phi_\theta(\vec{x}_{k+1})]_\Psi\bigr\|_2^2 ,
\end{align}
where $[\cdot]_\Psi$ extracts the encoder coordinates (the last $N-n_x$ entries) of a latent vector and $\bar\sigma_\Psi^2$ is the batch variance of
$\Psi_\theta(\vec{x})$, used to normalize the loss against feature scaling. The auxiliary inverse control
loss~\citep{abuduweili2025scaling} is
\begin{align}\label{eq:lossid}
  \mathcal{L}_{\mathrm{ctrl}}=\frac{1}{\sum_k\gamma^k}\sum_{k=0}^{H-1}\gamma^k\frac{1}{n_u}
  \bigl\|\mat{B}^{+}\bigl(\Phi_\theta(\vec{x}_{k+1})-\mat{A}\Phi_\theta(\vec{x}_k)\bigr)-\vec{u}_k\bigr\|_2^2 .
\end{align}
Standard Koopman minimizes $\mathcal{L}_{\mathrm{pred}}+\mathcal{L}_{\mathrm{ctrl}}$; Koopman + invariance penalty minimizes
$\mathcal{L}_{\mathrm{pred}}+\mathcal{L}_{\mathrm{ctrl}}+w\,\mathcal{L}_{\mathrm{lat}}$; and TIK fine-tuning in \Cref{app:ibc}
minimizes $\mathcal{L}_{\mathrm{pred}}$ plus the complement-radius hinge penalty under~\eqref{eq:ibc}. Standard training uses Adam with initial learning rate $10^{-3}$
decayed by $0.95$ every $1000$ steps, batch size $64$, gradient norm clipping at $1.0$, and $60{,}000$ gradient steps with
$\gamma=0.8$. Swing-up models use $H=40$ and $\gamma=0.95$. TIK fine-tuning uses the schedule in \Cref{app:ibc}.

\paragraph{Neural dynamics baselines.}
The neural dynamics baseline models $\vec{x}_{t+1}=\vec{x}_t+h_\phi(\vec{x}_t,\vec{u}_t)$, where $h_\phi$ is an MLP with two hidden layers of width $256$.
It uses the same architecture, offline data, rollout horizon, and optimizer as the Koopman encoder.
Neural model + NMPC linearizes $\vec{x}+h_\phi(\vec{x},\vec{u})$ at the current state at every control
step, centers the linearization at the target equilibrium, and solves the resulting convex QP with the same horizon, cost
weights, and input bounds as Koopman MPC. Neural model + MPPI~\citep{williams2017information} uses the same stage and terminal
cost, with sample count, exploration noise, and temperature tuned on the true linearization.

\textbf{Dual-mode MPPI settings.} In the swing-up study, each budget fixes the number of samples $K$ and horizon $H$ from measured
per-step runtimes. For each budget, the terminal weight is selected on design starts disjoint from the evaluation starts. Only TIK's exploration scale
$\sigma\in\{4,6,8,12\}$\,N and temperature $\lambda\in\{1,3,10,30\}$ are tuned: they are selected on $64$ design starts ($24$ hanging and
$40$ recovery starts across $4$ models) by the highest success rate on the design starts. The baselines keep the default MPPI settings ($\sigma=4$\,N, $\lambda=1$),
so the comparison in \Cref{tab:swrmpc} combines predictor quality with this tuning.

\subsection{Local Linearization Estimator Validation}\label{app:estimator}

At any query point $(\vec{x},\vec{u})$, we fit $\vec{x}_{i+1}\approx\vec{c}+\hat{\mat{A}}_f\vec{x}_i+\hat{\mat{B}}_f\vec{u}_i$
via distance-weighted least squares~\eqref{eq:localfit} on the $k=\max(200,8(n_x+n_u+1))$ nearest transitions in the offline
dataset. Across the $\pnRhoRouteN$ trained checkpoints on the benchmark systems in \Cref{tab:systems}, the estimated closed-loop
spectral radius $\rho(\hat{\mat{F}})$ agrees with the finite-difference simulator reference $\rho(\mat{F})$ to within a median
error of $\pnRhoRouteMed$ ($95$th percentile $\pnRhoRouteAgree$, worst-case error at most $\pnRouteOtherErrMax$).

\subsection{Evaluation Protocol and Statistical Testing}\label{app:protocol}

\begin{figure}[tbp]
\centering
\includegraphics[width=0.96\linewidth]{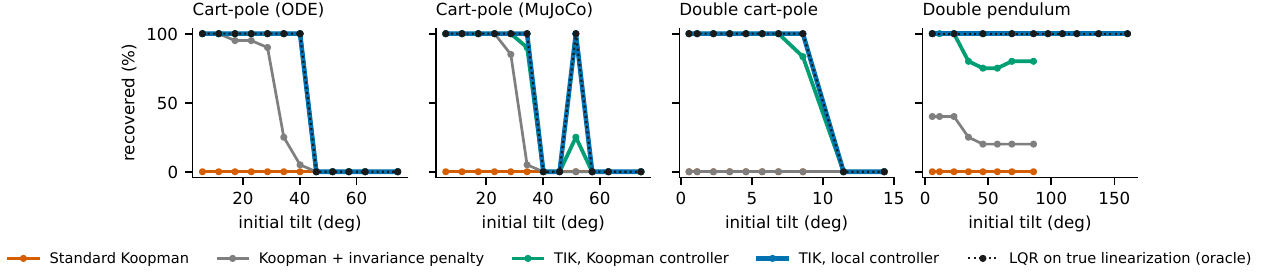}
\caption{\textbf{Empirical recovery from initial tilts.} Fraction of runs returning to the upright with all poles tilted at
rest (two tilt signs, up to $10$ seeds per learned curve, one oracle). TIK local feedback uses the fitted physical LQR gain
$\hat{\mat{K}}$. The lifted TIK controller uses the checkpoint after additional closed-loop training (\Cref{app:dagger}).}

\label{fig:roa}
\end{figure}

For each controller, we evaluate $\rho(\Acl)$, $\rho(\mat{F})$, and $\rho(\mat{\Pi}\Acl)$ at its closed-loop equilibrium. For
uncentered controllers (\Cref{app:fixedpoint_more}), equilibria of $g(\vec{x})=f(\vec{x},-\mat{K}\Phi_\theta(\vec{x}))$ are
found via damped Newton iteration; an equilibrium is admissible if the control input is strictly unsaturated and the state lies
within the $95$th percentile of the training data distribution. Every admissible equilibrium lies at least $\pnFpReluMin$ away
from any ReLU activation boundary, ensuring that $\Phi_\theta$ is continuously differentiable in a neighborhood of the
equilibrium. Nonlinear verification rollouts perturb the equilibrium along the leading eigenvector of $\mat{F}$ and along four
random directions and simulate the saturated closed loop for $3000$ steps: a rollout is classified as stable if perturbations
decay tenfold, unstable if they grow hundredfold, and undecided if $\rho(\mat{F})$ is too close to $1$ to resolve within
$3000$ steps. For the recovery range in \Cref{fig:roa}, all pole angles of the initial state equal $\pm\theta$ and all other coordinates are zero. The
saturated nonlinear loop runs for $1500$ steps ($2000$ on the double cart-pole), and a run counts as recovered if the final state
lies within $10^{-2}$ of the target. The tilts are $0.1$ to $1.3$\,rad on the cart-poles, $0.01$ to $0.25$\,rad on the double
cart-pole, and $0.1$ to $2.8$\,rad on the double pendulum. For each model, we average recovery over the two tilt signs. Each curve then averages these values across models. Where statistical significance is reported in the appendix, paired comparisons use the Wilcoxon signed-rank test
for continuous metrics or the exact McNemar test for binary outcomes, with Holm step-down corrections across multiple
comparisons.

% ══════════════════════════════════════════════════════════════
%  E  Supporting results
% ══════════════════════════════════════════════════════════════
\section{Supporting Experimental Results}\label{app:results}

\subsection{Extended Analysis of the Nominal Certificate}\label{app:fixedpoint_more}\label{app:related_more}

\paragraph{Extended model classes.}
\Cref{tab:classes_app} extends \Cref{tab:classes} to a broader set of Koopman architectures pooled across the four open-loop
unstable benchmark systems. Unlike in the main text, the controllers in \Cref{tab:classes_app}, apart from the centered row,
are evaluated without centering at the target equilibrium: each model is tested at the equilibrium $\vec{x}^\ast$ of its own
deployed loop $g(\vec{x}^\ast)=\vec{x}^\ast$, provided an equilibrium exists within the actuator limits and training domain. A
model is counted as \emph{tested} if it has such an admissible equilibrium and its latent loop is nominally certified
with $\rho(\Acl)<1$.

\begin{table}[tbp]
\caption{\textbf{Nominal stability certificate failure rate across extended model classes} (four open-loop unstable systems
pooled). Tested / trained: models that are nominally certified ($\rho(\Acl)<1$) and admit an unsaturated equilibrium within the
training domain, out of all trained models. Unstable on true system (\%): percentage of tested models whose deployed loop is
physically unstable ($\rho(\mat{F}(\vec{x}^\ast))>1$).}
\label{tab:classes_app}
\centering\small\setlength{\tabcolsep}{3.5pt}
\renewcommand{\arraystretch}{0.94}
\resizebox{\linewidth}{!}{%
\begin{tabular}{@{}llrr@{}}
\toprule
Class & Model architecture & \shortstack{Tested /\\trained} & \shortstack{Plant unstable (\%)\\$\rho(\mat{F}(\vec{x}^\ast))>1$} \\
\midrule
Linear & Standard Koopman & $\pnFpUncOLUN\,/\,\pnFpUncOLUTrained$ & $\pnClRecUnstPct$ \\
state & \quad without inverse control loss & $\pnNcNoFp\,/\,\pnNcNoN$ & $\pnClNoIdUnstPct$ \\
recovery & \quad centered at target equilibrium & $\pnCenUncN\,/\,\pnFpUncOLUTrained$ & $\pnClCenUnstPct$ \\
 & EDMD, quadratic polynomial dictionary & $\pnGenPolyTwoCertFp\,/\,\pnGenPolyTwoN$ & $\pnClPolyTwoUnstPct$ \\
 & EDMD, cubic polynomial dictionary & $\pnGenPolyThreeCertFp\,/\,\pnGenPolyThreeN$ & $\pnClPolyThreeUnstPct$ \\
 & EDMD, thin-plate radial basis functions & $\pnEDMDTotFp\,/\,\pnEDMDTotN$ & $\pnClRbfUnstPct$ \\
 & EDMD, random Fourier features & $\pnGenRffCertFp\,/\,\pnGenRffN$ & no stabilizing LQR \\
 & Autoencoder, linear state decoder & $\pnAETotFp\,/\,\pnAETotN$ & $\pnClAeLinUnstPct$ \\
\midrule
Published & EDMD for Koopman MPC~\citep{korda2018linear} & $\pnExKmFp\,/\,\pnExKmN$ & $\pnClKmUnstPct$ \\
control & Deep Koopman with control~\citep{shi2022deep} & $\pnExDkucFp\,/\,\pnExDkucN$ & $\pnClDkucUnstPct$ \\
methods & Stable operator~\citep{mamakoukas2023learning} & $\pnExSocFp\,/\,\pnExSocN$ & $\pnClSocUnstPct$ \\
 & Stable embedding~\citep{fan2024stablekoopman} & $\pnExSkelFp\,/\,\pnExSkelN$ & $\pnClSkelUnstPct$ \\
\midrule
Nonlinear & Autoencoder, nonlinear MLP decoder & $\pnGenAeNlCertFp\,/\,\pnGenAeNlN$ & $\pnClAeNlUnstPct$ \\
decoder & Autoencoder, latent linearity loss only & $\pnGenAeClCertFp\,/\,\pnGenAeClN$ & $\pnClAeClUnstPct$ \\
\midrule
Ours & \textbf{TIK (ours)} & $\pnTikPairTested\,/\,\pnTikPairTrained$ & $\pnClTikUnstPct$ \\
\bottomrule
\end{tabular}%
}
\end{table}

In \Cref{tab:classes_app}, $100\%$ of tested Standard Koopman models and $100\%$ of centered Standard Koopman models
are unstable on the true system. Removing the auxiliary inverse control loss reduces the failure rate only slightly to
$\pnClNoIdUnstPct\%$, with $\pnNcNoUnst$ of $\pnNcNoFp$ models unstable and McNemar $p=\pnNcP$, showing that failures persist
without the inverse control loss. We observe failures across all tested architectures that recover the physical state linearly
from the latent state. These models have eigenvalues near the unit circle and LQR gains with norms up to
$\|\mat{K}\|_2=\pnGenLinKMax$. Large gains can amplify errors in the input matching condition in \eqref{eq:ecal}. Autoencoders with a nonlinear MLP decoder have fewer local instabilities, but
because the physical state $\vec{x}=g_\phi(\vec{z})$ is a nonlinear function of $\vec{z}$, state constraints and quadratic state
costs become nonconvex in $\vec{z}$, so these models lose the convex QP structure of Koopman MPC. \Cref{tab:fixedpoint} breaks down the
results by system across all pooled regularization regimes, including noise augmentation and encoder Lipschitz penalties:
$\pnFpOLUunstK$ of $\pnFpOLUtrained$ trained controllers, or $\pnFpOLUpctFp\%$ of those admitting an unsaturated equilibrium,
are unstable on the true system, whereas controllers on open-loop stable systems with $\rho(\mat{A}_f)<1$ rarely fail.

\begin{table}[t]
\caption{\textbf{Breakdown of nominal certificate failures by benchmark system} (Standard Koopman and regularized training
regimes pooled, excluding the invariance penalty; uncentered controllers). The first four systems are open-loop unstable
($\rho(\mat{A}_f)>1$). Models with $\vec{x}^\ast$: models admitting an unsaturated equilibrium out of all trained models (all
are nominally certified). Spectral radii and the physical linearization error ratio $\|\mat{\Delta}\|_2/d_1(\mat{F}_0)$
(\Cref{prop:decomp}) are medians across seeds.}
\label{tab:fixedpoint}
\centering\small\setlength{\tabcolsep}{3.6pt}
\resizebox{\linewidth}{!}{% Auto-generated by paper_export.emit_table_fixedpoint.
\begin{tabular}{@{}lrrrrrrr@{}}
\toprule
 & open-loop & models & certified & \multicolumn{2}{c}{deployed, at $\vec{x}^\ast$} & simulation & closure \\
\cmidrule(lr){5-6}
System & $\rho(\mat{A}_f)$ & with $\vec{x}^\ast$ & $\rho(\Acl)$ & $\rho(\mat{F}(\vec{x}^\ast))$ & unstable & confirms & $\|\mat{\Delta}\|/d_1$ \\
\midrule
Double cart-pole (MuJoCo) & 1.093 & 22/25 & 0.9935 & 1.0672 & 19 (76\%) & 21/21 & 4780 \\
Cart-pole (ODE) & 1.083 & 28/30 & 0.9863 & 1.0015 & 15 (50\%) & 27/27 & 18 \\
Cart-pole (MuJoCo) & 1.082 & 30/30 & 0.9840 & 0.9979 & 12 (40\%) & 30/30 & 27 \\
Upright double pendulum & 1.060 & 30/30 & 0.9769 & 1.0242 & 27 (90\%) & 30/30 & 19 \\
Damped pendulum & 0.990 & 20/20 & 0.9853 & 0.9889 & 5 (25\%) & 20/20 & 16 \\
Seven-joint arm & 0.980 & 18/18 & 0.8486 & 0.9930 & 3 (17\%) & 18/18 & 46 \\
Hanging double pendulum & -- & 0/20 & \multicolumn{5}{c}{no admissible $\vec{x}^\ast$} \\
\bottomrule
\end{tabular}
}
\end{table}

\begin{figure}[tbp]
\centering
\includegraphics[width=0.72\linewidth]{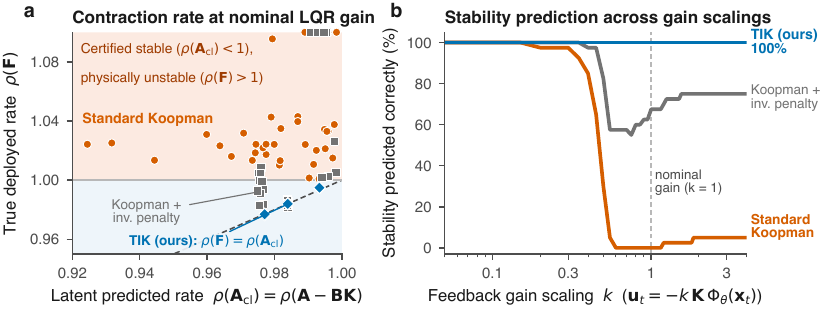}
\caption{\textbf{Predicted and deployed spectral radii and stability as the feedback gain is scaled.}
(a)~Spectral radius predicted by the latent model, $\rho(\Acl)=\rho(\mat{A}-\mat{B}\mat{K})$, plotted against the deployed spectral radius $\rho(\mat{F})$ at the nominal LQR gain for Standard Koopman (orange circles), Koopman + invariance penalty (gray
squares), and TIK (blue diamonds). (b)~Percentage of models ($40$ seeds per method) whose latent stability prediction matches
the true deployed stability as the feedback gain is scaled by $k\in[\pnPrKmin,\pnPrKmax]$.}
\label{fig:predict}
\end{figure}

\paragraph{Predicted versus deployed stability.}
\Cref{fig:predict} shows the data behind the gain-scaling result of \Cref{sec:exp_fixedpoint}.

\paragraph{Insensitivity to LQR cost weights and centering.}\label{app:gainsweep}\label{app:repair_more}
Sweeping the LQR control penalty $\mat{R}$ across three orders of magnitude over $[\pnGsRMin,\pnGsRMax]$ leaves $\pnGsUncAll$ of
$\pnGsUncN$ Standard Koopman controllers unstable at every weight. This occurs because the closed-loop error~\eqref{eq:ecal} is
dominated by the open-loop state-transition error $\|\Jtheta\mat{A}_f-\mat{A}\Jtheta\|_2$ (median $\pnGsUncEstate$), which is
independent of the feedback gain $\mat{K}$. By contrast, every TIK model remains stable across all tested LQR weights.

\paragraph{Soft invariance penalties.}\label{sec:exp_invariance}
Adding the latent linearity loss $\mathcal{L}_{\mathrm{lat}}$ with weight $w$ in \Cref{tab:invariance} eliminates the failure
on the single cart-poles, where the learned encoder becomes nearly linear, but leaves $\pnInvMujocoDoubleCartPoleOneUnst$ of
$\pnInvMujocoDoubleCartPoleOneN$ double cart-pole models unstable at $w=1$. Minimizing the average latent prediction error
on the training data does not enforce the derivative matching conditions sufficient for \Cref{thm:transfer}. Directly penalizing the error Jacobian $\|\mathcal{E}\|_2$ at the target during training still
leaves $\pnItUnst$ of $\pnItFp$ runs unstable, and post-hoc anchored invariance~\eqref{eq:anchored} converges on
$\pnAIMujocoDoubleCartPoleZeroConv$ of $\pnAIMujocoDoubleCartPoleZeroNAnch$ uncentered double cart-pole models.

\begin{table}[t]
\caption{\textbf{Effect of the soft invariance penalty weight $w$.} Values are medians over $n$ seeds. Setting $w=0$ corresponds to
Standard Koopman. Latent and state errors are one-step prediction errors on held-out validation data; nonlin.\ measures the
relative variation of the encoder Jacobian across the dataset ($0$ for an affine encoder). The final column is the number
unstable among models with an admissible equilibrium. The sweep includes runs evaluated at their own equilibria, so these
counts differ from the centered comparison in \Cref{tab:classes}.}
\label{tab:invariance}
\centering\small\setlength{\tabcolsep}{3.1pt}
\resizebox{\linewidth}{!}{% Auto-generated by paper_export.emit_table_invariance.
\begin{tabular}{@{}lrrrrrrrrrr@{}}
\toprule
System & $w$ & $n$ & latent err. & state err. & 14-step err. & nonlin. & $\rho(\Acl)$ & $\rho(\mat{F}(\vec{x}^\ast))$ & $\|\mathcal{E}\|_2$ & unstable \\
\midrule
Cart-pole (ODE) & 0 & 10 & 0.526 & 0.203 & 0.013 & 0.006 & 0.9873 & 1.0199 & 134 & 10/10 \\
Cart-pole (ODE) & 0.1 & 10 & 0.002 & 0.061 & 0.013 & 0.006 & 0.9839 & 0.9842 & 0.34 & 0/10 \\
Cart-pole (ODE) & 1 & 10 & 0.001 & 0.035 & 0.012 & 0.003 & 0.9839 & 0.9846 & 0.21 & 0/10 \\
Cart-pole (ODE) & 10 & 10 & 0.000 & 0.027 & 0.012 & 0.001 & 0.9839 & 0.9843 & 0.09 & 0/10 \\
\addlinespace
Cart-pole (MuJoCo) & 0 & 10 & 0.528 & 0.217 & 0.012 & 0.006 & 0.9886 & 1.0210 & 131 & 10/10 \\
Cart-pole (MuJoCo) & 0.1 & 10 & 0.002 & 0.057 & 0.012 & 0.006 & 0.9840 & 0.9850 & 0.33 & 0/10 \\
Cart-pole (MuJoCo) & 1 & 10 & 0.001 & 0.033 & 0.011 & 0.003 & 0.9839 & 0.9833 & 0.18 & 0/10 \\
Cart-pole (MuJoCo) & 10 & 10 & 0.000 & 0.028 & 0.011 & 0.001 & 0.9839 & 0.9838 & 0.076 & 0/10 \\
\addlinespace
Double cart-pole & 0 & 10 & 0.525 & 0.457 & 0.391 & 0.091 & 0.9918 & 1.1480 & 265 & 9/9 \\
Double cart-pole & 1 & 10 & 0.007 & 0.285 & 0.377 & 0.069 & 0.9944 & 1.1311 & 34 & 5/6 \\
Double cart-pole & 10 & 5 & 0.004 & 0.261 & 0.379 & 0.043 & 0.9929 & 1.0059 & 11 & 5/5 \\
\addlinespace
Seven-joint arm & 0 & 5 & 0.620 & 0.050 & 0.062 & 0.079 & 0.9969 & 0.9992 & 14 & 1/5 \\
Seven-joint arm & 1 & 5 & 0.003 & 0.047 & 0.064 & 0.116 & 0.9816 & 0.9777 & 0.056 & 0/5 \\
Seven-joint arm & 10 & 5 & 0.002 & 0.056 & 0.068 & 0.079 & 0.9825 & 0.9779 & 0.082 & 0/5 \\
\bottomrule
\end{tabular}
}
\end{table}

\paragraph{Predictive safety filters and unknown equilibria.}\label{sec:exp_mujoco}\label{sec:exp_certs}\label{app:certs_more}\label{app:dcrobust}
Wrapping a Standard Koopman controller in a predictive safety filter~\citep{wabersich2018linear} that predicts trajectories
using the learned latent model still violates state constraints in $\pnSMviolPsfKoop\%$ of MuJoCo cart-pole episodes, compared
to $\pnSMviolPsfData\%$ when the safety filter predicts using the local linear fit~\eqref{eq:localfit}, with Holm
$p=\pnSMpPsfData$. When the equilibrium of an uncentered controller is unknown, estimating it from $400$ offline
transitions and applying the offline stability test yields correct verdicts for $\pnDCLocTotConc$ of $\pnDCLocTotN$ Standard Koopman models with zero false conclusions across safety factors in
$[1,3]$ and neighbor counts $k$.

\subsection{Training and Certifying TIK}\label{app:ibc}

\Cref{alg:tik} summarizes the complete TIK procedure. Each TIK model is fine-tuned from the trained Standard Koopman checkpoint
of the same seed using Adam with learning rate $10^{-4}$, initializing $\mat{W}$ to the trained $\mat{A}$ and $\mat{N}_j=\mat{0}$,
and using a soft hinge penalty $100\max(0,\rho_c-\bar\rho)^2$ on the spectral radius $\rho_c$ of the orthogonal complement
block $(\mat{I}-\Jtheta\Jtheta^{+})\mat{W}(\mat{I}-\Jtheta\Jtheta^{+})$ to encourage the dynamics on the complement to be Schur. \Cref{prop:ibc}(b) requires a separate Schur check.
Regulation models are fine-tuned for $6000$ steps ($\bar\rho=0.98$), and bilinear swing-up models for $20{,}000$ steps
($\bar\rho=0.999$). On the double cart-pole, where the deployed spectral radius of Standard Koopman ranges from
$\pnIbcSrcRhoMin$ to $\pnIbcSrcRhoMax$, fine-tuning with TIK brings every model below $\rho(\mat{F})\le\pnIbcRhoMax<1$.
Independent replications using freshly drawn offline datasets yield $\pnRepDcpStab/\pnRepDcpN$ stable models on the double
cart-pole, $\pnRepDcpOffStab/\pnRepDcpOffN$ at an off-origin equilibrium, and $\pnRepUpStab/\pnRepUpN$ on the upright double
pendulum.

\paragraph{Closed-loop training extends the recovery range.}\label{app:dagger}
Because the offline training trajectories do not visit all large-tilt closed-loop states, we fine-tune TIK in closed-loop
rounds ($4$ rounds with $6{,}000$ updates on the cart-poles, and $20$ rounds with $40{,}000$ updates and $70\%$ replay-pool
sampling on the double cart-pole and double pendulum). In each round, we simulate the current Koopman controller from $300$
initial states (all pole angles tilted by up to $\theta_{\max}$ with $10\%$ actuator exploration noise) for $250$ steps, and
add $15$-step windows with pole angles below $3\theta_{\max}$ to a replay pool ($\theta_{\max}=0.7$\,rad on the cart-poles,
$0.15$\,rad on the double cart-pole, and $1.5$\,rad on the double pendulum). \Cref{tab:dagger} reports the result: tangent
invariance is preserved by construction, and all $\pnDagTotN$ models that complete training are stable on the true system (one
double cart-pole seed diverged to NaN and is excluded). With the same closed-loop data but no tangent matching, only $\pnDagCtrlStable$ of $\pnDagCtrlN$ controllers are stable on the plant.

\begin{table}[tbp]
\caption{\textbf{Recovery range of the TIK controller before and after training on closed-loop rollouts.} Largest tilt (rad) with at least
half of the runs recovered, median over models; ``not worse'': models whose range did not shrink; ``control'': (model, tilt) cells
recovered by the model trained with the same data but no invariance constraint.}
\label{tab:dagger}
\centering\small\setlength{\tabcolsep}{5pt}
\begin{tabular}{@{}lccccc@{}}
\toprule
System & Oracle & Before $\uparrow$ & After $\uparrow$ & Not worse $\uparrow$ & Control cells $\downarrow$ \\
\midrule
Cart-pole (ODE) & $\pnDagCpOracle$ & $\pnDagCpBase$ & $\pnDagCpNew$ & $\pnDagCpBetter/\pnDagCpN$ & $\pnDagCpCtrlCells$ \\
Cart-pole (MuJoCo) & $\pnDagMcpOracle$ & $\pnDagMcpBase$ & $\pnDagMcpNew$ & $\pnDagMcpBetter/\pnDagMcpN$ & $\pnDagMcpCtrlCells$ \\
Double cart-pole & $\pnDagDcpOracle$ & $\pnDagDcpBase$ & $\pnDagDcpNew$ & $\pnDagDcpBetter/\pnDagDcpN$ & $\pnDagDcpCtrlCells$ \\
Double pendulum & $>\!1.5$ & $\pnDagDpuBase$ & $\pnDagDpuNew$ & $\pnDagDpuBetter/\pnDagDpuN$ & -- \\
\bottomrule
\end{tabular}
\end{table}

\paragraph{Checking the offline error bounds.}\label{app:ibc_cert}\label{app:kmpc}
In \Cref{tab:ibccert}, the two-scale protocol fits a local cubic polynomial (\Cref{cor:cubic}) on $\pnTSWideN$ transitions
within radius $5\times10^{-2}$ of the target and fits the linear model~\eqref{eq:localfit} on $\pnTSNarrowN$ transitions within
radius $5\times10^{-3}$. \Cref{tab:ibccert_checks} compares the estimated bounds with simulator derivatives across the fits
in \Cref{tab:ibccert} and additional noise regimes. The error bound~\eqref{eq:fitbound} covers the simulator Jacobian error in every listed fit.
However, the curvature estimate underestimates simulator curvature in some fits, so the checks do not establish all assumptions of \Cref{thm:knownfp}.

\paragraph{Local MPC implementation.} In condensed form, the Koopman MPC quadratic
program~\eqref{eq:kmpc_qp} optimizes over the $N_h=20$ control inputs and is solved via DAQP in a median of
$\pnKmSolveMed\,\mu$s; near the target equilibrium, its effective feedback gain agrees with the infinite-horizon Koopman LQR
gain $\mat{K}$ to within $\pnKmKErr$ on every model (\Cref{prop:kmpc}).

\begin{table}[t]
\caption{\textbf{Offline error bounds checked against simulator derivatives.} The study includes the regimes in
\Cref{tab:ibccert} and additional noise levels ($100$ fits per cart-pole row and $40$ per double cart-pole row).}
\label{tab:ibccert_checks}
\centering\small\setlength{\tabcolsep}{4pt}
\begin{tabular}{@{}lllcc@{}}
\toprule
System & Offline data regime & Noise $\epsilon$ & \shortstack{Bound $\eta$ contains\\true error $\uparrow$} & \shortstack{Curvature bound\\holds $\uparrow$} \\
\midrule
Cart-pole swing-up & Near the target & $0$ & $\pnICDefZCov/\pnICDefZCovN$ & $\pnICDefZLCov/\pnICDefZCovN$ \\
 & Two scales & $0$ & $\pnICDefZTCov/\pnICDefZTCovN$ & $\pnICDefZTLCov/\pnICDefZTCovN$ \\
 & Two scales & $10^{-5}$ & $\pnICDsgFiveTCov/\pnICDsgFiveTCovN$ & $\pnICDsgFiveTLCov/\pnICDsgFiveTCovN$ \\
 & Two scales, noisy states & $10^{-7}$ & $\pnICDsgSevenTVCov/\pnICDsgSevenTVCovN$ & $\pnICDsgSevenTVLCov/\pnICDsgSevenTVCovN$ \\
 & Closed-loop trajectories & $3\times10^{-8}$ & $\pnICDsgThEightTJCov/\pnICDsgThEightTJCovN$ & $\pnICDsgThEightTJLCov/\pnICDsgThEightTJCovN$ \\
\midrule
Double cart-pole & Two scales & $0$ & $\pnIDDefZCov/\pnIDDefZCovN$ & $\pnIDDefZLCov/\pnIDDefZCovN$ \\
 & Two scales & $10^{-6}$ & $\pnIDDsgSixCov/\pnIDDsgSixCovN$ & $\pnIDDsgSixLCov/\pnIDDsgSixCovN$ \\
 & Two scales, noisy states & $10^{-9}$ & $\pnIDDsgNineVCov/\pnIDDsgNineVCovN$ & $\pnIDDsgNineVLCov/\pnIDDsgNineVCovN$ \\
 & Closed-loop trajectories & $10^{-10}$ & $\pnIDDsgTenJCov/\pnIDDsgTenJCovN$ & $\pnIDDsgTenJLCov/\pnIDDsgTenJCovN$ \\
\bottomrule
\end{tabular}
\end{table}

\subsection{Swing-Up Ablation, Safe-Control-Gym, and Manipulator Details}\label{sec:exp_swing}\label{app:recover}\label{app:latency}\label{app:robust}\label{app:bench}\label{sec:exp_arm}\label{app:arm}

\paragraph{Swing-up ablation and equal-effort comparison.}
On the MuJoCo cart-pole swing-up task with continuous angle embedding $(x,\sin\theta,\cos\theta,\dot x,\dot\theta)$, an episode
succeeds if the pole is held upright ($\cos\theta>0.98$) over the final $2$\,s of a $10$\,s rollout without touching the rail
stops. A single MPPI planning step with $512$ trajectory samples requires a median of $\pnLatKoopMs$\,ms on the bilinear Koopman
model versus $\pnLatMlpMs$\,ms on the neural network model; under tighter time budgets, each planner selects the sample count
and horizon that fits the budget and maximizes validation success. \Cref{tab:ibc} presents the component ablation discussed in
\Cref{sec:exp_bench}, illustrating the separate effects of tangent matching and bilinear actuation. When both TIK and
Neural model + MPPI are trained from scratch for $3\times$ as many gradient steps ($180{,}000$ steps), the neural network
recovers from aggressive kicks slightly more often when compute is unconstrained ($\pnRtMlpRec\%$ vs.\ $\pnRtInvRecOwn\%$),
whereas TIK matches its $100\%$ swing-up rate, achieves $99\%$ combined success at $1$\,ms (vs.\ $\pnRtLatOneMlp\%$ for Neural
MPPI), and remains the only controller capable of executing within the $0.25$\,ms budget ($80\%$ success).

\paragraph{Why the two LQR controllers can differ.}
In centered coordinates, their unsaturated actions satisfy
\[
 \vec{u}_{\mathrm{lift}}-\vec{u}_{\mathrm{fit}}
 =-\mat{K}\bigl(\Phi_\theta(\vec{x})-\Jtheta\vec{x}\bigr)
 =o(\|\vec{x}\|_2).
\]
Thus \Cref{prop:ibc} gives equal feedback derivatives at the target, not equal actions throughout the upright neighborhood.
The $88\%$ versus $99\%$ success rates in \Cref{tab:ibc} are compatible with this local result. They do not isolate the cause of the gap:
that would require matched switching states, thresholds, and saturation rules. The budget sweep in \Cref{tab:swrmpc}
is a separate protocol, even where a reported percentage is the same.

\begin{table}[t]
\caption{\textbf{Component ablation on cart-pole swing-up} ($20$ models per row). Entry is the fraction of MPPI-only runs reaching the upright neighborhood. Success is the fraction
completing the task with MPPI alone.
The two MPPI + LQR columns report overall success after switching to lifted LQR or LQR on the fitted physical linearization.
The final column counts models with an unstable deployed lifted LQR loop among those with an admissible upright fixed point.}
\label{tab:ibc}
\centering\footnotesize\setlength{\tabcolsep}{3.0pt}
\renewcommand{\arraystretch}{0.92}
\resizebox{\linewidth}{!}{%
\begin{tabular}{@{}lccrrrrr@{}}
\toprule
 & & & \multicolumn{2}{c}{MPPI only (\%)} & \multicolumn{2}{c}{MPPI + LQR (\%)} & \\
\cmidrule(lr){4-5}\cmidrule(lr){6-7}
Model variant & Matched & Bilinear & Entry & Success & Lifted & Fitted & Unstable \\
\midrule
Standard Koopman & -- & -- & $\pnSwISrcPlanReach$ & $\pnSwISrcPlan$ & $\pnSwISrcLifted$ & $\pnSwISrcData$ & $\pnSwISrcUpUnst/\pnSwISrcUpFp$ \\
Fine-tuned without TIK & -- & -- & $\pnSwIFtPlanReach$ & $\pnSwIFtPlan$ & $\pnSwIFtLifted$ & $\pnSwIFtData$ & $\pnSwIFtUpUnst/\pnSwIFtUpFp$ \\
Fine-tuned without TIK & -- & \checkmark & $\pnSwIBilFreePlanReach$ & $\pnSwIBilFreePlan$ & $\pnSwIBilFreeLifted$ & $\pnSwIBilFreeData$ & $\pnSwIBilFreeUpUnst/\pnSwIBilFreeUpFp$ \\
TIK, linear input & \checkmark & -- & $\pnSwILinPlanReach$ & $\pnSwILinPlan$ & $\pnSwILinLifted$ & $\pnSwILinData$ & $\pnSwILinUpUnst/\pnSwILinUpFp$ \\
\midrule
\textbf{TIK (ours)} & \checkmark & \checkmark & $\pnSwIBilPlanReach$ & $\pnSwIBilPlan$ & $\pnSwIBilLifted$ & $\pnSwIBilData$ & $\pnSwIBilUpUnst/\pnSwIBilUpFp$ \\
\bottomrule
\end{tabular}%
}
\end{table}

\begin{table}[tbp]
\caption{\textbf{Real-time MPC on simulated seven-joint manipulators under per-step computation budgets.} Success rate (\%) on Kinova Gen3
reaching around an obstacle ($\pnArmModels\times\pnArmEpisodes$ episodes) and on a torque-controlled arm reaching six goals
($\pnMpkSeeds\times\pnMpkEp$ episodes). The linear MPC reference uses a fit from logged transitions on Kinova and a
single workspace-center linearization on the six-goal arm. The archived protocol gives the reference configuration for
this latter baseline and the Go2 baseline, but leaves their Jacobian source unspecified.}
\label{tab:robots}
\centering\small\setlength{\tabcolsep}{4pt}
\renewcommand{\arraystretch}{0.88}
\begin{tabular}{@{}llrrr@{}}
\toprule
 & & \multicolumn{2}{c}{Kinova reach (\%) $\uparrow$} & Six-goal arm (\%) $\uparrow$ \\
\cmidrule(lr){3-4}\cmidrule(l){5-5}
Class & Method & $5$\,ms & $20$\,ms & $1$\,ms \\
\midrule
Linear & Linear MPC & $\pnArmFiveLinSucc$ & $\pnArmTwentyLinSucc$ & $\pnMrLinOneSucc$ \\
Neural & Neural model + MPPI & $\pnArmFiveTunedSucc$ & $\pnArmTwentyTunedSucc$ & $\pnMrMppiOneSucc$ \\
 & Neural model + NMPC & -- & -- & $\pnMrSlOneSucc$ \\
Koopman & Standard Koopman & $\pnArmFiveStdSucc$ & $\pnArmTwentyStdSucc$ & $\pnMrStdOneSucc$ \\
\midrule
Ours & \textbf{TIK (ours)} & $\pnArmFiveIbcSucc$ & $\pnArmTwentyIbcSucc$ & $\pnMrOursOneSucc$ \\
\bottomrule
\end{tabular}
\end{table}

\paragraph{Safe-control-gym and Kinova Gen3 details.}
\Cref{tab:bench} reports the complete results on the safe-control-gym cart-pole stabilization benchmark~\citep{yuan2022safecontrolgym}.
TIK reduces state constraint violations from $\pnScgCpStdViol\%$ for Standard Koopman to $\pnScgCpIbcViol\%$, matching LQR on
the oracle linearization ($\pnScgCpLqrViol\%$). On the Kinova Gen3 arm ($50$\,Hz joint-acceleration control reaching $\pnArmGoals$
goal poses around a spherical obstacle, solved via OSQP), TIK solves each QP step in a median of $\pnArmFiveIbcMsMed$\,ms,
reaches the goal in $100\%$ of episodes at $5$\,ms (Holm $p=\pnArmPFiveSucc$ vs.\ Neural model + MPPI), and achieves lower closed-loop
trajectory cost at $20$\,ms ($\pnArmTwentyIbcCost$ vs.\ $\pnArmTwentyStdCost$ for Standard Koopman, $p=\pnArmPStdCost$).

\begin{table}[t]
\caption{\textbf{Safe-control-gym cart-pole stabilization benchmark.} Percentage of episodes that violate state constraints
($\downarrow$) and that successfully stabilize to the target ($\uparrow$) across $\pnScgCpN$ models $\times$ $\pnScgEpisodes$
initial states.}
\label{tab:bench}
\centering\small\setlength{\tabcolsep}{4.5pt}
\renewcommand{\arraystretch}{0.9}
\resizebox{\linewidth}{!}{%
\begin{tabular}{@{}llrr@{}}
\toprule
Class & Controller & Constraint violations (\%) $\downarrow$ & Stabilized (\%) $\uparrow$ \\
\midrule
Linear & Oracle LQR (simulator Jacobians) & $\pnScgCpLqrViol$ & $\pnScgCpLqrSucc$ \\
 & Fitted linear LQR (offline logs) & $\pnScgCpLqrEstViol$ & $\pnScgCpLqrEstSucc$ \\
 & Oracle linear MPC (benchmark) & $\pnScgCpMpcViol$ & $\pnScgCpMpcSucc$ \\
\midrule
Neural & Neural model + NMPC & $\pnScgNnViol$ & $\pnScgNnSucc$ \\
 & Neural model + MPPI & $\pnScgNnMppiViol$ & $\pnScgNnMppiSucc$ \\
\midrule
Koopman & Standard Koopman & $\pnScgCpStdViol$ & $\pnScgCpStdSucc$ \\
 & \quad + safety filter on latent model & $\pnScgCpPsfLatViol$ & $\pnScgCpPsfLatSucc$ \\
 & \quad + safety filter on linear fit & $\pnScgCpPsfEstViol$ & $\pnScgCpPsfEstSucc$ \\
\midrule
Ours & \textbf{TIK (ours)} & $\pnScgCpIbcViol$ & $\pnScgCpIbcSucc$ \\
\bottomrule
\end{tabular}%
}
\end{table}

\subsection{Multiple Equilibria and Higher-Dimensional Robotic Systems}\label{app:multipoint}

\begin{proposition}[Obstruction to multi-equilibrium invariance with a linear input]\label{prop:multipoint}
Consider a linear latent model $\hat{\vec{z}}_{t+1}=\mat{A}\vec{z}_t+\mat{B}\vec{u}_t$ with state recovery $\mat{C}\mat{J}_i=\mat{I}$
at equilibria $i=1,\dots,k$. If $\mat{A}\mat{J}_i=\mat{J}_i\hat{\mat{A}}_i$ and $\mat{B}=\mat{J}_i\hat{\mat{B}}_i$ for all
$i=1,\dots,k$, then: (a)~the Markov parameters satisfy $\mat{C}\mat{A}^m\mat{B}=\hat{\mat{A}}_i^m\hat{\mat{B}}_i$ for every
$m\ge0$ and every $i$; and (b)~if each pair $(\hat{\mat{A}}_i,\hat{\mat{B}}_i)$ is controllable, then
$\hat{\mat{A}}_i=\hat{\mat{A}}_1$ and $\hat{\mat{B}}_i=\hat{\mat{B}}_1$ for all $i$.
\end{proposition}

\begin{proof}
(a)~By induction on $m$, $\mat{A}^m\mat{J}_i=\mat{J}_i\hat{\mat{A}}_i^m$, so
$\mat{C}\mat{A}^m\mat{B}=\mat{C}\mat{A}^m\mat{J}_i\hat{\mat{B}}_i=\mat{C}\mat{J}_i\hat{\mat{A}}_i^m\hat{\mat{B}}_i=\hat{\mat{A}}_i^m\hat{\mat{B}}_i$.
(b)~By part~(a), the matrix $\mathcal{C}=\mat{C}[\mat{B},\,\mat{A}\mat{B},\,\dots,\,\mat{A}^{n_x-1}\mat{B}]$ equals the
controllability matrix of every pair $(\hat{\mat{A}}_i,\hat{\mat{B}}_i)$ and therefore has full row rank $n_x$. Since
$\hat{\mat{A}}_i\mathcal{C}=\mat{C}[\mat{A}\mat{B},\,\dots,\,\mat{A}^{n_x}\mat{B}]$ is independent of $i$, right-multiplying by
$\mathcal{C}^{+}$ gives $\hat{\mat{A}}_i=\hat{\mat{A}}_1$, and $\hat{\mat{B}}_i=\mat{C}\mat{B}=\hat{\mat{B}}_1$.
\end{proof}

\begin{corollary}[Non-existence of exact linear-input liftings across distinct equilibria]\label{cor:noexact}
Let $\Phi$ be a finite-dimensional differentiable lifting with linear state recovery $\mat{C}\Phi(\vec{x})=\vec{x}$, and let $\mat{A},\mat{B},\mat{C}$ be constant matrices. If $\Phi(f(\vec{x},\vec{u}))=\mat{A}\Phi(\vec{x})+\mat{B}\vec{u}$
holds for all $(\vec{x},\vec{u})$, then all equilibria of $f$ with controllable linearizations share the same
linearization $(\mat{A}_f,\mat{B}_f)$. In particular, no such exact lifting exists for a pendulum, whose downward and upright
equilibria have distinct controllable linearizations.
\end{corollary}

\begin{proof}
Differentiating $\Phi(f(\vec{x},\vec{u}))=\mat{A}\Phi(\vec{x})+\mat{B}\vec{u}$ at any equilibrium $(\vec{x}_e,\vec{u}_e)$ yields
the relations $\mat{J}_\Phi(\vec{x}_e)\mat{A}_f(\vec{x}_e)=\mat{A}\mat{J}_\Phi(\vec{x}_e)$ and
$\mat{J}_\Phi(\vec{x}_e)\mat{B}_f(\vec{x}_e)=\mat{B}$ with $\mat{C}\mat{J}_\Phi(\vec{x}_e)=\mat{I}$, so \Cref{prop:multipoint}(b)
applies directly.
\end{proof}

Bilinear input allows the effective matrices to vary across goals. The restriction in \Cref{prop:multipoint} therefore
no longer follows. Joint feasibility still depends on the goal embeddings and fitted Jacobians.
See also \citet{liu2025immersions,shang2026existence,ge2026koopman} for related restrictions across operating points.
At goal $i$, write $\vec{z}_i=\Phi_\theta(\vec{x}_i)$, $\mat{A}_i^z=\mat{A}+\sum_j u^\ast_{ij}\mat{N}_j$, and
$\mat{B}_i^z=\mat{B}+[\mat{N}_1\vec{z}_i\ \cdots\ \mat{N}_{n_u}\vec{z}_i]$. Training projects onto the local
matching equations $\mat{A}_i^z\mat{J}_i=\mat{J}_i\hat{\mat{A}}_i$ and $\mat{B}_i^z=\mat{J}_i\hat{\mat{B}}_i$ and
constrains each $\vec{z}_i$ to be fixed under $\vec{u}_i^\ast$. At fixed encoder weights these constraints are linear in
$\mat{A},\mat{B},\{\mat{N}_j\}$, and we check feasibility numerically. For a local model in goal deviations, exact matching and the Schur, stabilizability, and detectability conditions of
\Cref{prop:ibc}(b) give the fitted physical LQR gain on the tangent space.

\paragraph{Seven-joint arm across compute budgets.}\label{app:mparm}
On the torque-controlled seven-joint manipulator with a $14$-dimensional state (\Cref{tab:systems}), the $\pnMpkGoals$ goal poses have true linearizations
differing by up to $\pnMpkASpread$. With the bilinear TIK model, multi-point tangent invariance holds to within $\pnMpkResid$
across all $\pnMpkSeeds$ seeds, stabilizing all $\pnMpkStableOurs/\pnMpkStableOurs$ goal--model pairs, compared with $\pnMpkStableStd/\pnMpkStableOurs$ for Standard Koopman. The offline test returned
stable verdicts for $\pnMrCert/\pnMrCertN$ TIK pairs and $\pnMpkCertStd/\pnMpkCertStdN$ Standard Koopman pairs, with no false
stable verdicts in simulator checks. In \Cref{tab:multipoint}, Standard Koopman fails at all
budgets ($0\%$) and Neural MPPI fails at $1$--$2$\,ms ($\pnMrMppiOneSucc\%$--$\pnMrMppiTwoSucc\%$), whereas multi-point TIK
achieves $100\%$ at $1$--$5$\,ms and outperforms Neural NMPC at $1$\,ms ($\pnMrSlOneSucc\%$, Holm $p=\pnMrPSlOne$).
Linear MPC at the workspace center also reaches $\pnMrLinOneSucc\%$ at $1$\,ms, so this task shows no success-rate advantage for TIK over that baseline.

\begin{table}[t]
\caption{\textbf{Torque-controlled seven-joint manipulator reaching six goals across computation budgets.} Success rate (\%)
over $\pnMpkSeeds$ models and $\pnMpkEp$ episodes per model per budget. TIK's $10$\,ms run had deadline misses
under machine load. A repeat at lower load reached $\pnMrOursTenRep\%$. The table reports the original budget sweep.}
\label{tab:multipoint}
\centering\small\setlength{\tabcolsep}{4.5pt}
\renewcommand{\arraystretch}{0.88}
\resizebox{\linewidth}{!}{%
\begin{tabular}{@{}llrrrrr@{}}
\toprule
 & & \multicolumn{5}{c}{Success rate (\%) at per-step compute budget $\uparrow$} \\
\cmidrule(l){3-7}
Class & Controller & $1$\,ms & $2$\,ms & $5$\,ms & $10$\,ms & $20$\,ms \\
\midrule
Linear & Linear MPC (workspace center) & $\pnMrLinOneSucc$ & $\pnMrLinTwoSucc$ & $\pnMrLinFiveSucc$ & $\pnMrLinTenSucc$ & $\pnMrLinTwentySucc$ \\
Neural & Neural model + MPPI & $\pnMrMppiOneSucc$ & $\pnMrMppiTwoSucc$ & $\pnMrMppiFiveSucc$ & $\pnMrMppiTenSucc$ & $\pnMrMppiTwentySucc$ \\
 & Neural model + NMPC (relinearized) & $\pnMrSlOneSucc$ & $\pnMrSlTwoSucc$ & $\pnMrSlFiveSucc$ & $\pnMrSlTenSucc$ & $\pnMrSlTwentySucc$ \\
Koopman & Standard Koopman MPC & $\pnMrStdOneSucc$ & $\pnMrStdTwoSucc$ & $\pnMrStdFiveSucc$ & $\pnMrStdTenSucc$ & $\pnMrStdTwentySucc$ \\
\midrule
Ours & \textbf{TIK (ours, multi-point QP)} & $\pnMrOursOneSucc$ & $\pnMrOursTwoSucc$ & $\pnMrOursFiveSucc$ & $\pnMrOursTenSucc$ & $\pnMrOursTwentySucc$ \\
\bottomrule
\end{tabular}%
}
\end{table}

\paragraph{Unitree Go2 quadruped posture recovery and contact boundaries.}\label{app:go2}
We evaluate a Unitree Go2 quadruped with a $35$-dimensional state in MuJoCo Menagerie under $100$\,Hz joint-target control over a $500$\,Hz PD loop,
holding $\pnGoAnchors$ standing postures and recovering from impulse pushes without foot slip in \Cref{tab:go2}. Across the
$\pnGoAnchors$ postures, the true linearized state matrices differ by $\pnGoDAMin$--$\pnGoDAMax\%$. Linear MPC designed at a single pose
succeeds in $\pnGrLinSucc\%$ of episodes and the robot falls in $\pnGrLinFall\%$. Standard Koopman succeeds in
$\pnGrOldSucc\%$. Multi-point TIK with a smooth encoder, goal offsets consistent with the model, and tightened constraints reaches $\pnGrOursSucc\%$ under nominal conditions and $\pnGrOursNoise\%$--$\pnGrOursMassHi\%$
under sensor noise, friction changes, and $+20\%$ payload mass, outperforming Neural MPPI at $\pnGrSmppiSucc\%$ ($p=\pnGrPSmppi$)
and Neural NMPC at $\pnGrSlSucc\%$ ($p=\pnGrPSl$) while running $6.4$--$8.1\times$ faster at $\pnGrOursCompMed$\,ms per step. A Koopman model without tangent matching, using the same smooth encoder and constraint tightening, reaches $\pnGrFreeSucc\%$.
The offline test is inconclusive in all $\pnGoCertN$ cases, with $r_{\mathbb{R}}(\hat{\mat{F}})\le\pnGoRrMax<\eta$.
The deployed spectral radius is near one (median $\pnGoRhoTrueMed$), and nonsmooth contact transitions can violate the
local differentiability assumptions.

\begin{table}[t]
\caption{\textbf{Unitree Go2 quadruped ($35$D state, $12$D input) posture recovery under pushes and domain shifts}
($\pnGoEpisodes$ episodes per entry).}
\label{tab:go2}
\centering\footnotesize\setlength{\tabcolsep}{2.2pt}
\renewcommand{\arraystretch}{0.92}
\resizebox{\linewidth}{!}{%
\begin{tabular}{@{}llrrrrrrr@{}}
\toprule
 & & \multicolumn{5}{c}{Success rate (\%) $\uparrow$} & & \\
\cmidrule(lr){3-7}
Class & Method & Nominal & Noise & Friction & $0.8\times$ mass & $1.2\times$ mass & Falls (\%) $\downarrow$ & ms/step $\downarrow$ \\
\midrule
Linear & Linear MPC (single standing pose) & $\pnGrLinSucc$ & -- & -- & -- & -- & $\pnGrLinFall$ & $\pnGrLinCompMed$ \\
Neural & Neural model + MPPI + filter & $\pnGrSmppiSucc$ & -- & -- & -- & -- & $\pnGrSmppiFall$ & $\pnGrSmppiCompMed$ \\
 & Neural model + NMPC & $\pnGrSlSucc$ & -- & -- & -- & -- & $\pnGrSlFall$ & $\pnGrSlCompMed$ \\
Koopman & Standard Koopman QP & $\pnGrOldSucc$ & -- & -- & -- & -- & $\pnGrOldFall$ & $\pnGrOldCompMed$ \\
\midrule
Ours & \textbf{TIK (ours, multi-point QP)} & $\pnGrOursSucc$ & $\pnGrOursNoise$ & $\pnGrOursFric$ & $\pnGrOursMassLo$ & $\pnGrOursMassHi$ & $\pnGrOursFall$ & $\pnGrOursCompMed$ \\
\bottomrule
\end{tabular}%
}
\end{table}

\subsection{Dual-Mode QP-MPC on the Double Cart-Pole}\label{app:dcpdual}

We evaluate the dual-mode QP-MPC of \Cref{sec:kmpc} on the double cart-pole from tilted starts:
$|\theta_1|\in[18^\circ,28^\circ]$, $\theta_2$ of opposite sign, and cart offsets up to $0.3$\,m
($\pnDmEps$ starts per model). The local controller is $\vec{u}=-\hat{\mat{K}}\vec{x}$.
With only the original offline data, TIK has one-step error $0.08$ on local-controller trajectories and
$\pnDmOld\%$ MPC success. We then fine-tune for $80{,}000$ updates, drawing $60\%$ of each batch from
$15$-step windows under the local controller from tilted starts with input noise on disjoint seeds, and adding near-target windows.
We save checkpoints every $5{,}000$ updates and select by MPC success on disjoint validation starts.

The selected models have one-step error $0.01$ and succeed in $\pnDmTikAll\%$ of rollouts across
$\pnDmModels$ models ($\pnDmTikLat\%$ and $\pnDmTikUnc\%$ in the two regimes; up to $95\%$ for one seed).
Of all rollouts, $\pnDmEnt\%$ enter $\Omega_c$; every model succeeds at least once.
Fitted linear MPC with the same terminal set reaches $\pnDmLin\%$, and fitted LQR applied from the start reaches
$\pnDmCatch\%$. Standard Koopman dual-mode MPC has no successful rollout ($10/10$ models fail).
Training on the same closed-loop data without tangent matching gives $\pnDmCtrl\%$ success
($\pnDmCtrlZero/\pnDmCtrlN$ models never succeed).

\end{document}